%% file: main.tex
\documentclass[lettersize,journal]{IEEEtran}
\usepackage{amsmath,amsfonts}
\usepackage{algorithmic}
\usepackage{algorithm}
\usepackage{array}
\usepackage{textcomp}
\usepackage{stfloats}
\usepackage{url}
\usepackage{hyperref}
\usepackage{verbatim}
\usepackage{graphicx}
\usepackage{cite}
\usepackage[table]{xcolor}
\usepackage{xcolor}
\usepackage{arydshln}
\usepackage{tabularx} 
\usepackage{caption}
\usepackage{float}
\usepackage{booktabs}
\usepackage{multirow}
\usepackage{adjustbox}
\usepackage{makecell}
\usepackage{subcaption}
\usepackage{rotating}  
\usepackage[normalem]{ulem} 

\newif\ifshowrevisions
\showrevisionstrue  
\definecolor{myblue}{RGB}{0,0,0}

\ifshowrevisions
  \newcommand{\new}[1]{{\color{myblue}#1}}
  \newcommand{\del}[1]{{\color{red}\sout{#1}}} 
\else
  \newcommand{\new}[1]{#1}
  \newcommand{\del}[1]{}
  
\fi

\input{sections/_preamble}

\definecolor{somegray}{rgb}{0.5, 0.5, 0.5}
\newcommand{\darkgrayed}[1]{\textcolor{somegray}{#1}}
\makeatletter
\newcommand*\titleheader[1]{\gdef\@titleheader{#1}}
\AtBeginDocument{%
  \let\st@red@title\@title
  \def\@title{%
    \vskip-1.5em
    \bgroup\normalfont\large\centering\@titleheader\par\egroup

    \vskip0.5em
    \st@red@title}
}
\makeatother

\titleheader{
\darkgrayed{
\fontsize{11.9}{11}\selectfont
This paper has been accepted for publication in the IEEE Transactions on Robotics (T-RO), 2026.\ \textcopyright\ IEEE
}
}

\title{Event-Aided Sharp Radiance Field\\ Reconstruction for Fast-Flying Drones}
\begin{document}

\input{sections/_title}

\input{sections/0_abstract}
\input{sections/1_introduction}
\input{sections/2_relatedworks}
\input{sections/3_method}

\input{sections/4_experiments}

\input{sections/5_discussions}
\input{sections/6_conclusions}

\input{sections/7_acknowledgements}

\bibliographystyle{IEEEtran}
\bibliography{main}

\input{sections/n_biography}

\end{document}

%% file: sections/_preamble.tex
\usepackage{multirow}
\usepackage{xspace}

\usepackage{pifont}

\usepackage[normalem]{ulem}
\useunder{\uline}{\ul}{}

\newcommand{\myparagraph}[1]{\par\vspace{0.3em}\noindent\textbf{#1}\hspace{0.2em}}

\usepackage{tikz}
\usepackage{pgfplots}
\usepackage{pgfplotstable}

\pgfplotsset{compat=1.18}

\pgfplotsset{select coords between index/.style 2 args={
    x filter/.code={
        \ifnum\coordindex<#1\fi
        \ifnum\coordindex>#2\fi
    }
}}

\pgfplotsset{
    standard/.style={
        enlargelimits=true,
        enlarge x limits=false,
        ymajorgrids=true,
        grid style=dashed,
        xtick pos=left,
        ytick pos=left,
        max space between ticks=25,
        set layers=tick labels on top
    },
    layers/tick labels on top/.define layer set=
        {axis background,axis grid,axis ticks,axis lines,main,%
          axis tick labels,
          axis descriptions,axis foreground}
        {/pgfplots/layers/standard}
}


%% file: sections/_title.tex



\author{Rong Zou\textsuperscript{*}, Marco Cannici\textsuperscript{*}, and Davide Scaramuzza\textsuperscript{} \\
Robotics and Perception Group, University of Zurich, Switzerland \\
}

\twocolumn[{%
    \renewcommand\twocolumn[1][]{#1}%
    \maketitle
    \thispagestyle{empty}
    \vspace{-5ex}
    \input{floaters/figures/teaser_wide}
}]

\makeatletter
\renewcommand{\@makefntext}[1]{%
  \noindent\makebox[0.8em][l]{\@makefnmark}#1}
\makeatother

\renewcommand{\thefootnote}{\fnsymbol{footnote}}
\footnotetext[1]{R. Zou and M. Cannici equally contributed to this work.}

\renewcommand{\thefootnote}{\arabic{footnote}}

\renewcommand{\thefootnote}{}
\footnotetext{This work was supported by the National Centre of Competence in Research (NCCR) Robotics (grant agreement No. 51NF40-185543) through the Swiss National Science Foundation (SNSF), the European Union’s Horizon Europe Research and Innovation Programme under grant agreement No. 101120732 (AUTOASSESS), the European Research Council (ERC) under grant agreement No. 864042 (AGILEFLIGHT), and the Swiss AI Initiative by a grant from the Swiss National Supercomputing Centre (CSCS) under project ID a03 on Alps.}

\renewcommand{\thefootnote}{\arabic{footnote}}

%% file: floaters/figures/teaser_wide.tex
\begin{center}
    \includegraphics[width=\linewidth]{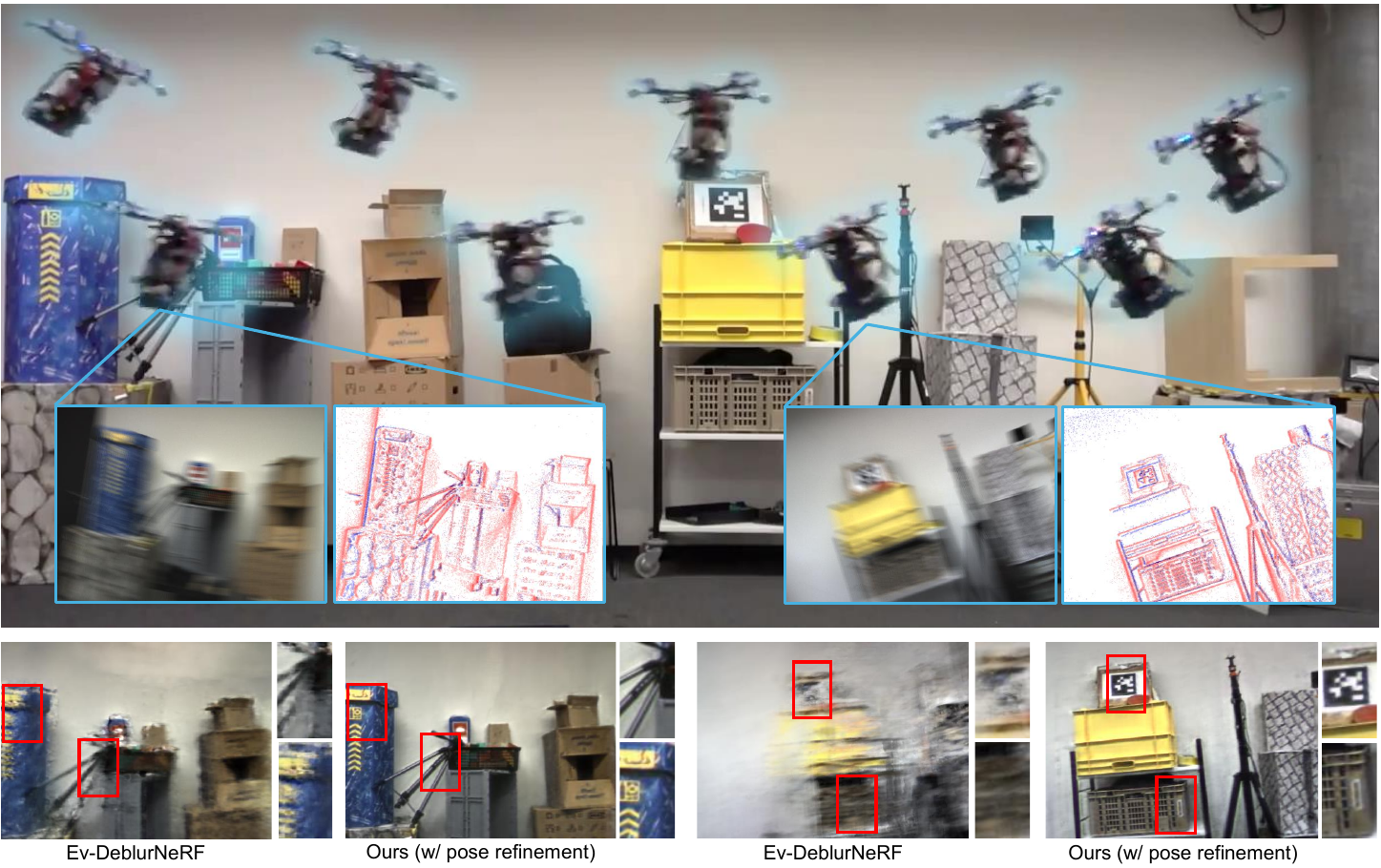}
    
    \captionof{figure}{
    Our system recovers sharp geometry and texture from footage captured by a drone flying at 2 m/s, learning a neural radiance field directly from motion-blurred images and event data collected during flight (\textbf{top}). Without relying on ground-truth poses or external motion capture, it refines the drone’s trajectory during training, starting from a rough visual-inertial odometry prior. Once optimized, the model can render photorealistic views from novel perspectives (\textbf{bottom}), paving the way for high-speed, vision-based inspection tasks in agile robotics.
    }
    \label{fig:teaser_wide}
\end{center}

%% file: sections/0_abstract.tex
\begin{abstract}
Fast-flying aerial robots promise rapid inspection under limited battery constraints, with direct applications in infrastructure inspection, terrain exploration, and search and rescue. However, high speeds lead to severe motion blur in images and induce significant drift and noise in pose estimates, making dense 3D reconstruction with Neural Radiance Fields (NeRFs) particularly challenging due to their high sensitivity to such degradations. In this work, we present a unified framework that leverages asynchronous event streams alongside motion‑blurred frames to reconstruct high‑fidelity radiance fields from agile drone flights. By embedding event‑image fusion into NeRF optimization and jointly refining event-based visual‑inertial odometry priors using both event and frame 
modalities, our method recovers sharp radiance fields and accurate camera trajectories without ground‑truth supervision. We validate our approach on both synthetic data and real‑world sequences captured by a fast‑flying drone. 
Despite highly dynamic drone flights, where RGB frames are severely degraded by motion blur and pose priors become unreliable, our method reconstructs high-fidelity radiance fields and preserves fine scene details,
delivering a performance gain of over 50\% on real-world data compared to state-of-the-art methods.

\end{abstract}


%% file: sections/1_introduction.tex
\begin{center}
\textsc{Multimedia Materials} 
\end{center}

Video of experiments: \url{https://youtu.be/dVaH0VVXhQc}

Code: \url{https://github.com/uzh-rpg/event-sharp-nerf-drones}

\section{Introduction} \label{sec:introduction}

\IEEEPARstart{R}{obotic} systems operating in the real world often face a fundamental trade-off between sensing quality and operational efficiency. 
\new{In particular, fast-flying aerial robots are essential for
tasks such as 
large-scale infrastructure inspection (e.g., powerlines, pipelines, or railways), time-constrained terrain exploration, large-area agricultural and forestry surveys, or search and rescue. In these applications, flying faster enables a larger area to be covered within narrow time windows and under a limited battery budget, which is crucial for maximizing mission efficiency \cite{bauersfeld2022range}.}
At low speeds, energy is wasted in maintaining lift without making significant progress, while high-speed flight introduces aerodynamic drag. Striking the right balance can lead to shorter mission durations, fewer battery replacements, and more effective deployment in time-sensitive settings.

However, increasing flight speed makes high-fidelity perception significantly more challenging.  Motion blur corrupts visual data, while fast dynamics degrade the accuracy of visual-inertial odometry, limiting the performance of downstream tasks such as 3D reconstruction and scene understanding. These limitations pose a challenge for generating dense, photorealistic 3D models, crucial for map-based planning, semantic reasoning, and safe navigation.

Recent advances in learning-based scene representations, such as Neural Radiance Fields (NeRFs) \cite{mildenhall2020nerf} and 3D Gaussian Splatting \cite{kerbl20233d}, have shown great promise for view-consistent reconstruction from sparse camera views. These methods have rapidly found 
applications in robotics for localization, mapping, and visual simulation. 
However, they rely on two core assumptions: that input images are free of motion blur, and that accurate camera poses are available. In the context of fast-flying drones, neither of these assumptions holds. As a result, radiance field learning under fast motion remains largely an unsolved problem.

Prior works \cite{qi20243,deguchi2024e2gs,weng2024eadeblur,zhang2024elite} have attempted to bridge this gap by leveraging event cameras, which offer high temporal resolution and robustness to motion blur, making them well-suited for high-speed scenarios. These methods typically aim to recover sharp images and accurate poses from events, via model-based \cite{pan2019bringing} or learning-based \cite{sun2022event} deblurring approaches. Once deblurred, images are passed to off-the-shelf structure-from-motion tools like COLMAP \cite{schonberger2016structure} to recover camera poses for NeRF training. However, under fast motion, these pipelines degrade significantly. Model-based methods often rely on simplified sensor dynamics and can fail under fast motion dynamics or sensor noise. Meanwhile, learning-based approaches may hallucinate inconsistent image content across views, leading to poor pose estimates. Furthermore, by relying solely on reconstructed images for pose estimation, these approaches miss the opportunity to directly incorporate the continuous, high-frequency information available in the event stream. 

In this work, we address these limitations by proposing a unified framework for sharp radiance field reconstruction 
tailored to 
fast-flying drones operating under noisy pose conditions. Our method integrates motion-blurred images and asynchronous event data using a shared, learnable camera trajectory module, drawing on a continuous-time formulation \cite{ma2024continuous}. We show that this representation enables joint modeling of event-based trajectories and approximation of the motion blur formation process, allowing both events and frames to supervise each other during training. Unlike previous approaches that treat event and image poses separately \cite{cannici2024mitigating}, our formulation refines a single, shared trajectory initialized using event-based visual-inertial odometry. This allows our system to recover accurate scene structure and motion without requiring ground-truth trajectories or slow, offline preprocessing.

We validate our method on synthetic and real-world drone sequences featuring fast flight and challenging motion profiles. To support reproducibility and benchmarking, we introduce the first drone dataset for radiance field reconstruction under fast motion, featuring synchronized motion-blurred RGB images and event data captured with a beamsplitter-based setup. Our results show that the proposed method achieves high-fidelity reconstructions even when initialized with noisy pose priors, significantly outperforming existing NeRF-based deblurring baselines on real drone data.

\textbf{Our contributions are:}
\begin{itemize}
    \item A NeRF-based framework for radiance field reconstruction in high-speed robotic settings, capable of recovering sharp, photorealistic scene representations from motion-blurred images and asynchronous event data. By exploiting the complementary strengths of both modalities, 
    the proposed framework enables high-fidelity mapping even under agile maneuvers. Our method delivers a performance gain of more than 50\% on real-world data captured by a fast-flying drone, significantly outperforming state-of-the-art approaches.
    \item A continuous-time, shared trajectory formulation that unifies pose refinement and motion blur modeling within a single optimization process. Our approach allows mutual supervision between events and frames during training, and improves initial event-based visual-inertial odometry priors, leading to more robust and consistent reconstructions under fast motion.
    \item A real-world demonstration of our approach on \new{data captured by} a custom aerial platform equipped with a beamsplitter-based sensor rig featuring hardware-synchronized RGB and event cameras. 
    \new{The collected data are post-processed offline to demonstrate the method’s performance under high-speed flight conditions up to 2 m/s. Accurate ground-truth poses provided by a motion capture system are used for evaluation.}
    We release the dataset and code to support reproducible research in high-speed, event-driven scene reconstruction for robotics.
\end{itemize}

%% file: sections/2_relatedworks.tex
\section{Related Work} \label{sec:related_works}

\myparagraph{Learning 3D Reconstruction under Motion Blur.}
Neural 3D scene representations, such as Neural Radiance Fields (NeRFs) \cite{mildenhall2020nerf} and 3D Gaussian Splatting (3DGS) \cite{kerbl20233d}, 
have become powerful tools for novel view synthesis and dense scene reconstruction. Their success in recovering geometry and appearance from sparse multi-view imagery has led to growing interest in robotics applications \cite{rosinol2022nerf,zheng2024gaussiangrasper,chen2025splat}. While early methods assumed clean, well-calibrated inputs and operated offline, 
later work introduced faster optimization \cite{muller2022instant,chen2022tensorf,hanson2025speedy,fang2024mini}, support for dynamic scenes \cite{park2021nerfies,li2022neural}, and robustness to pose noise \cite{lin2021barf,bian2023nope,fu2024colmap}.

In line with these efforts, 
a particularly active area of research focuses on extending these methods to recover accurate 3D geometry from motion-blurred images. 
Deblur-NeRF \cite{ma2022deblur} addresses this by jointly estimating a latent sharp radiance field and a learned view-dependent blur model. PDRF \cite{peng2022pdrf} introduces coarse-to-fine estimation using proxy geometry, while DP-NeRF \cite{lee2023dp} constrains motion to lie in rigid subspaces. BAD-NeRF \cite{wang2022bad} takes a more geometric approach, recovering camera motion during exposure via photometric bundle adjustment. 
This direction has also been explored in the context of 3D Gaussian Splatting. 
Deblur-GS \cite{chen2024deblur} and BAD-Gaussians \cite{zhao2024bad} adapt NeRF-style joint optimization to 3DGS by modeling image formation under motion blur and refining both camera trajectories and Gaussian parameters. BAGS \cite{peng2024bags} extends this by introducing a Blur Proposal Network that estimates dense per-pixel blur kernels and a mask identifying sharp regions, enabling robust reconstruction under spatially-varying blur. Gaussian Splatting on the Move \cite{seiskari2024gaussian} further extends this line of work to rolling shutter settings, leveraging visual-inertial odometry within a differentiable 3DGS pipeline.
Despite these improvements, these methods remain sensitive to the quality of input poses and struggle when all training views are similarly affected by fast motion. In mobile robotics scenarios—particularly onboard drones—such motion patterns are common and unavoidable. 
Moreover, these methods rely exclusively on RGB images, which are inherently limited by blur and low temporal resolution.

\myparagraph{Event-Based Neural 3D Reconstruction.}
Event cameras offer an alternative sensing modality for robotics applications
such as visual odometry, SLAM, and scene reconstruction \cite{vidal2018ultimate,tulyakov2022time}, particularly in fast and challenging environments. Unlike conventional cameras that capture full intensity images at fixed rates, event sensors asynchronously report per-pixel brightness changes at microsecond resolution. This enables low-latency perception, naturally avoids motion blur, and performs reliably under challenging lighting conditions.
These advantages have led to growing interest in incorporating events into neural 3D reconstruction frameworks, including both NeRF and Gaussian Splatting approaches, with the potential to enable accurate scene reconstruction under fast motion. Ev-NeRF \cite{hwang2023ev} and EventNeRF \cite{rudnev2022eventnerf} demonstrate that events alone can supervise the reconstruction of a static scene, using the event generation model \cite{gallego2022survey} as a supervisory signal. Robust e-NeRF \cite{low2023robust} adds supervision through a normalized gradient loss to improve robustness to pose variation. Event-based 3DGS methods have recently emerged as an alternative. EV-GS \cite{wu2024ev} adopts a fully event-based pipeline, initializing Gaussians from randomly sampled scene points. Event3DGS \cite{xiong2024event3dgs} accumulates events based on scene entropy and uses them to iteratively initialize and refine a 3D Gaussian point cloud. Finally, a different approach is proposed in EvGGS \cite{wang2024evggs}, which introduces a generalizable, feedforward pipeline for event-based Gaussian Splatting by combining learned depth estimation, intensity reconstruction, and Gaussian regression for direct geometry prediction from events.
Building on these approaches, our method introduces motion-blurred images as an additional source of supervision, allowing for more precise geometry reconstruction and the recovery of colored textures.

\myparagraph{Sharp 3D from Events and Blurred Frames.}
Given the success of event cameras in 3D reconstruction under fast motion, recent work has explored combining asynchronous events with motion-blurred RGB frames to enable sharper and more robust neural scene reconstruction. By fusing the high temporal precision of events with the rich appearance information in blurred images, these methods aim to recover accurate geometry and texture even in challenging dynamic conditions. E-NeRF \cite{klenk2022nerf} demonstrates that combining events with blurry images can already reduce motion blur in NeRF reconstructions, while E2NeRF \cite{qi2023e2nerf} takes a step further by explicitly modeling the blur formation process during training, leading to sharper textures and more accurate color reconstruction. Ev-DeblurNeRF \cite{cannici2024mitigating} further refines this setup by incorporating network enhancements from PDRF~\cite{peng2022pdrf} and DP-NeRF~\cite{lee2023dp}, event-wise supervision, a learnable response function, and model-based priors.
Inspired by these NeRF-based approaches, recent methods extend similar ideas to 3D Gaussian Splatting. E2GS \cite{deguchi2024e2gs} adopts a Deblur-GS-style \cite{chen2024deblur}  pipeline, using event-based deblurred images and COLMAP~\cite{schonberger2016structure} poses for Gaussian initialization. In contrast, Event3DGS \cite{xiong2024event3dgs} uses an iterative strategy that first reconstructs geometry using events alone and then refines appearance from blurry images in a second stage, optimizing only color.

Despite these advances, most existing methods rely on pre-deblurred images or external structure-from-motion pipelines such as COLMAP~\cite{schonberger2016structure}, to estimate camera poses. These approaches are not only slow but also brittle under fast motion—event-based deblurring can fail with unmodeled dynamics, and pose estimates may drift due to inconsistent image content. To overcome this, we initialize the trajectory using a real-time event-based visual-inertial odometry system \cite{vidal2018ultimate}, avoiding expensive camera estimation. We then refine this trajectory during training through joint optimization, where both blurry images and events supervise a shared, learnable pose. This enables consistent alignment across modalities and accurate radiance field reconstruction from noisy priors, while fully exploiting the high temporal resolution of events, which is crucial for robust mapping in high-speed robotics scenarios.

%% file: sections/3_method.tex
\input{floaters/figures/architecture}

\section{Method} \label{sec:method}

We aim to reconstruct a latent, sharp radiance field of a static scene from visual data captured during fast flight. In particular, given input observations $\{\mathcal{I}$, $\mathcal{E}$, $\mathcal{M}\}$ corresponding to a sequence of motion-blurred RGB frames  $\mathcal{I} =   \{\mathbf{C}^\text{blur}_i\}_{i=1}^{N_I}$, asynchronous events $\mathcal{E} = \{\mathbf{e}_j\}_{j=1}^{N_E}$, and inertial measurements $\mathcal{M} = \{(\mathbf{a}_k, \boldsymbol{\omega}_k, t_k)\}_{k=1}^{N_M}$ captured from a moving camera, we seek to recover the latent continuous volumetric function
\begin{equation} \label{eq:nerf_mapping}
    \mathcal{F} : (\mathbf{x}, \mathbf{d}) \rightarrow (\mathbf{c}, \sigma),    
\end{equation}
which maps a 3D location $\mathbf{x} \in \mathbb{R}^3$ and viewing direction $\mathbf{d} \in \mathbb{R}^3$ to its emitted color $\mathbf{c}  \in \mathbb{R}^3$ and volume density $\sigma  \in \mathbb{R}$. Each input image $\mathbf{C}^\text{blur}_i \in \mathbb{R}^{H \times W \times 3}$ captures a motion-blurred RGB frame at timestamp $t_i$, while the event stream provides asynchronous measurements $\mathbf{e}_j = (\mathbf{u}_j, t_j, p_j)$, where $\mathbf{u}_j = (u_j, v_j)$ denotes the pixel location, $t_j$ the timestamp, and $p_j \in \{-1, 1\}$ the polarity of the detected brightness change.

Our method builds upon recent event-aided deblur NeRF models \cite{qi2023e2nerf,klenk2022nerf,cannici2024mitigating}, which optimize a NeRF using both image- and event-based supervision. Inspired by Ev-DeblurNeRF \cite{cannici2024mitigating}, we adopt a tri-branch architecture that jointly leverages information from different sensing modalities to guide the learning of a sharp radiance field. The first branch operates on blurred RGB frames and supervises the radiance field via a photometric blur model. 
A second, event-based branch supervises the radiance field using microsecond-level brightness changes captured by the event sensor, which are interpreted through a learnable camera response function (CRF) that adapts to real sensor characteristics. Finally, the third branch introduces prior knowledge to constrain training further and resolve ambiguities arising from severe blur or sparse event information. 
An overview of the proposed method is presented in Figure~\ref{fig:network}.

Unlike previous approaches, our method does not rely on accurate, high-frequency camera poses during image exposure or at event timestamps. Instead, we propose a unified framework that estimates a temporally continuous trajectory by combining event-based odometry with a learned residual correction module based on \cite{ma2024continuous}. This module refines the coarse odometry at arbitrary time resolutions, enabling recovery of sub-millisecond-accurate camera poses for each triggered event as well as the continuous trajectory traced during the exposure of each image. Crucially, in our proposed architecture the image- and event-based branches work together to cooperatively refine both the radiance field and the camera trajectory, enabling sharper reconstructions through improved pose estimation.

In the following sections, we describe each component in detail, starting with the trajectory representation and pose refinement in Sec.~\ref{sec:pose_refinement}, followed by the radiance field supervision via events and blur modeling in Secs.~\ref{sec:image_supervision} and \ref{sec:event_supervision}.

\subsection{Trajectory Representation and Pose Refinement}
\label{sec:pose_refinement}

Accurate training of event- and image-based NeRF models requires temporally precise camera poses at the exact locations where each observation was captured. This challenge becomes even more pronounced in event-aided deblur settings, where one must estimate not only the camera trajectory during each image exposure, but also the precise pose at which every event was triggered. Prior approaches \cite{cannici2024mitigating,klenk2022nerf,qi2023e2nerf} either assume access to densely sampled ground-truth poses, rarely feasible in real-world scenarios, or rely on two-stage pipelines that deblur frames using event integration \cite{pan2019bringing} and then estimate poses with structure-from-motion tools like COLMAP \cite{schonberger2016structure}. These methods not only depend heavily on the quality of the deblurred images, which degrade under fast motion, but also estimate poses only at discrete timestamps, rather than modeling the continuous camera motion. In contrast, we propose a unified, learning-based trajectory model that generalizes to fast, high-dynamic scenarios without requiring accurate intermediate reconstructions or pose annotations.

We begin by computing a coarse estimate of the camera trajectory using an off-the-shelf visual-inertial SLAM or VIO system. In our implementation, we adopt UltimateSLAM~\cite{vidal2018ultimate} for its ability to fuse both image and event features for robust pose tracking. Despite the presence of motion blur in the RGB frames, image gradients still contribute to feature matching, while the high temporal resolution of the event stream ensures resilience to fast motion. This provides discrete pose predictions $\mathcal{P}_\text{SLAM} = \left\{ \mathbf{T}_\text{SLAM}(\tau) \right\}_{\tau=1}^{T}$ where each $\mathbf{T}_\text{SLAM}(\tau) \in \text{SE(3)}$ represents the estimated camera pose at time $\tau$. In our implementation, UltimateSLAM~\cite{vidal2018ultimate} outputs pose estimates at the RGB camera rate, yielding a relatively sparse set of trajectory samples $\tau \in \{t_1, t_2, \dots, t_{N_I}\}$, where $N_I$ is the number of captured RGB frames. These sparse pose estimates serve as the initial trajectory prior, which we refine via a continuous-time pose model. 

To enable refinement at arbitrary times, we first construct a continuous initial trajectory $\mathbf{T}_\text{init}(t) \in \mathrm{SE}(3)$ by interpolating the discrete SLAM pose estimates. Specifically, we apply a spherical linear interpolation ($\operatorname{SLERP}$) that interpolates rotations on the $\mathrm{SO}(3)$ manifold and translations linearly in $\mathbb{R}^3$: \looseness=-1
\begin{equation}
\mathbf{T}_\text{init}(t) = \operatorname{SLERP}_{\mathrm{SE(3)}} \big( \mathcal{P}_\text{SLAM}, t \big),
\end{equation}
where $t$ denotes any arbitrary timestamp during the camera motion.
While this continuous prior provides a smooth estimate of the trajectory, it inevitably suffers from drift and local inaccuracies. To address this, we learn to estimate a residual refinement transformation $\Delta \mathbf{T}(t) \in \mathrm{SE}(3)$ that corrects the initial pose at each timestamp:
\begin{equation}
    \mathbf{T}_\text{ref}(t) = \Delta \mathbf{T}(t)\cdot \mathbf{T}_\text{init}(t).
\end{equation}

Following \cite{ma2024continuous}, we parametrize this refinement as a multi-layer perceptron (MLP) network $\Delta \mathbf{T}_\text{refine}(t) = f_\phi(\gamma_{L_t}(t))$:
\begin{equation}
    f_\phi : \mathbb{R}^{2L_t} \to \mathrm{SE}(3),
\end{equation}
where $\gamma_{L_t}(\cdot)$ denotes a frequency-encoded representation of time, following the sinusoidal positional encoding introduced in NeRF~\cite{mildenhall2020nerf}. The network $f_\phi$ maps this embedding to the residual pose $\Delta \mathbf{T}_\text{refine}(t)$, computed from the predicted translation vector $\mathsf{v}(t) \in \mathbb{R}^3$ and quaternion $\mathsf{q}(t) \in \mathbb{R}^4$.

At any timestamp $t$, the refined camera pose $\mathbf{T}_\text{ref}(t)$ defines the origin and orientation of the virtual camera for that instant. 
\new{The scene is rendered using this pose, and the resulting rendered view is supervised by both event and RGB camera observations (except for the frame-only case in Sec.~\ref{sec:Experimental} where supervision is obtained solely from RGB images). In both cases, the blurry images together with the motion-blur modeling techniques introduced in Sec.~\ref{sec:image_supervision} are incorporated during training. 
}
\new{During each training iteration, the photometric and event-based losses computed on these rendered virtual views are backpropagated through the differentiable rendering process to both the scene representation and the pose refiner. This means that the scene parameters and the camera poses are jointly optimized within the same training step, rather than in separate stages.}
The initial coarse trajectory is crucial, as it provides a plausible starting point that guides the system toward a consistent structure and appearance. 
In the next sections, we describe how these refined poses are used in practice to render motion-blurred images and event representations for supervision.

\subsection{Image Supervision via Motion Blur Rendering}
\label{sec:image_supervision}

Following previous literature \cite{ma2022deblur,cannici2024mitigating,lee2023dp,peng2022pdrf}, to enable radiance field learning from motion-blurred RGB frames, we adopt a physically grounded blur formation model that integrates sharp color predictions over the exposure interval. This allows us to supervise the radiance field using blurred observations, without access to ground-truth sharp frames, provided the camera motion during exposure is accurately modeled.

Given a pixel $\mathbf{u}$ in a frame captured at time $t_i$, where $t_i$ denotes the center of the exposure interval for image $i$, we define $\mathbf{r}(\mathbf{u}, t_i)$ as the ray that originates from the camera center of the refined camera pose $\mathbf{T}_\text{ref}(t_i)$ and passes through the pixel $\mathbf{u}$. We estimate the color of the blurred pixel $\mathbf{u}$ observed in $\mathbf{C}^\text{blur}_i$ by averaging the colors observed along the set of rays $\mathbf{r}(\mathbf{u}, t_i)$ as the camera moves during the exposure interval $[t_i - \tau/2,\ t_i + \tau/2]$. This simulates the physical image formation process, where the observed pixel color corresponds to the integral of the latent sharp radiance accumulated over the exposure duration \cite{ma2022deblur}.

To perform this integration, we sample a set of $M$ timestamps $\{t^{(1)}, \ldots, t^{(M)}\} = \mathfrak{S}_\tau(t_i)$ within the exposure interval, centered at the image timestamp $t_i$. For each sampled time $t^{(m)}$, we retrieve the corresponding camera pose from the refined trajectory $\mathbf{T}_\text{ref}(t^{(m)})$ and cast a ray $\mathbf{r}(\mathbf{u}, t^{(m)})$ from the resulting viewpoint. In our experiments, we found that implementing $\mathfrak{S}$ as uniform sampling within the exposure interval provides accurate and stable results. However, our formulation is general and could accommodate learned sampling strategies--for example, modeling $\mathfrak{S}(\cdot)$ as an MLP conditioned on $t_i$, similar to $f_\phi$, or adopting adaptive schemes based on event rates, as explored in prior work \cite{qi20243}.

To render individual views, we follow standard NeRF volume rendering \cite{mildenhall2020nerf}.
Along each ray, we sample a set of 3D points and query the continuous volumetric function $\mathcal{F}$ defined in~\eqref{eq:nerf_mapping} to obtain colors and densities at each location. In practice, $\mathcal{F}$ 
is implemented as 
two MLPs, coarse and fine, denoted by $F_\Omega^c$ and $F_\Omega^f$. Inspired by \cite{peng2022pdrf,cannici2024mitigating}, we augment both networks with explicit feature volumes \cite{chen2022tensorf} $\mathcal{V}_s$ and $\mathcal{V}_l$. These volumes are evaluated at the sampled ray positions along each ray to produce features $f^c$, $f^f$, which we concatenate to the inputs of the coarse and fine MLPs, respectively. These features accelerate convergence and enhance rendering quality, as they introduce spatially anchored features that are easier to optimize.

The colors and densities collected along each ray are then aggregated through a volumetric rendering operator \cite{mildenhall2020nerf}, 
which we denote
as $\mathcal{R}$.  The final blurred pixel color is obtained by averaging these latent sharp values over the sampled timestamps:
\begin{equation} \label{eq:blur_color}
\hat{\mathbf{C}}^\text{blur}(\mathbf{u}, t_i) = g\left( \sum_{m=1}^{M} w_m \, \mathcal{R}\big( \mathbf{r}(\mathbf{u}, t^{(m)}) \big) \right),
\end{equation}
where $w_m$ denotes the blending weight for the $m$-th timestamp and satisfies $\sum_{m=1}^M w_m = 1$, and $g(\cdot)$ is a gamma correction function. Similar to \cite{qi2023e2nerf}, we use uniform weights to blend the rendered colors over time. 

For brevity, we denote the rendered pixel in Eq. \ref{eq:blur_color} as $\hat{\mathbf{C}}^\text{blur}_{\mathbf{r}}$, we finally compare the synthetic and observed blurry pixels over a batch $\mathcal{B}_b$ of pixel-timestamp pairs $(\mathbf{u}_k,t_k)$:
\begin{equation}
\label{eq:loss_blur}
\mathcal{L}_\text{blur} = \frac{1}{|\mathcal{B}_b|} \sum_{\mathbf{r}_k \in \mathcal{B}_b} \Big[
\left\|\hat{\mathbf{C}}^\text{blur}_{\mathbf{r}_{k,c}} - \mathbf{C}^\text{blur}_{\mathbf{r}_k} \right\|_2^2
+ \left\| \hat{\mathbf{C}}^\text{blur}_{\mathbf{r}_{k,f}} - \mathbf{C}^\text{blur}_{\mathbf{r}_k} \right\|_2^2 \Big]
\end{equation}
where subscripts $c$ and $f$ indicate outputs from the coarse and fine networks, respectively, and $\mathbf{C}^\text{blur}_{\mathbf{r}_k}$ the ground-truth blurred color at the pixel corresponding to $\textbf{r}_k$. Note that, for compactness, we refer to rays $\mathbf{r}_k$ in the summation although the batch is formally defined over pixel-timestamp pairs $(\mathbf{u}_k, t_k)$ from which the corresponding rays are derived (i.e., the rays cast from the refined poses $\mathbf{T}_\text{ref}(t_k)$ and passing through $\mathbf{u}_k$).

While the blur rendering in Eq. \ref{eq:loss_blur} provides a strong supervisory signal, it still leaves the learning problem under-constrained, especially under high-dynamic motion where motion blur severely degrades textured areas. To further guide the radiance field toward plausible sharp appearances, we follow~\cite{cannici2024mitigating} and introduce an additional supervision signal based on prior deblurring estimates of the training images. 

Specifically, for each ray $\mathbf{r}_k \in \mathcal{B}_b$, sampled at mid-exposure $t_k$, 
we associate a pseudo ground-truth sharp target 
$\mathbf{C}_{\mathbf{r}_k}^\text{prior}$ obtained by deblurring the blurry input using the event-based double integral (EDI)~\cite{pan2019bringing}. We then supervise the NeRF-rendered color at the mid-exposure pose to match this deblurred estimate:
\begin{equation}
\label{eq:loss-prior}
\mathcal{L}_\text{prior} = \frac{1}{|\mathcal{B}_b|} \!\sum_{\mathbf{r}_k \in \mathcal{B}_b}\!\! \Big[
\left\| \mathcal{R}_c(\mathbf{r}_k) \!-\! \mathbf{C}_{\mathbf{r}_k}^\text{prior} \right\|_2^2 +
\left\| \mathcal{R}_f(\mathbf{r}_k) \!-\! \mathbf{C}_{\mathbf{r}_k}^\text{prior} \right\|_2^2
\Big],
\end{equation}
where $\mathcal{R}_c(\mathbf{r}_k)$ and $\mathcal{R}_f(\mathbf{r}_k)$ are the coarse and fine volumetric renderings of the ray $\mathbf{r}_k$ at mid-exposure.

Since EDI is a model-based deblurring process that assumes accurate and noise-free event measurements, its outputs can contain artifacts when the event data is noisy or the scene violates model assumptions. To address this, we employ the prior-based supervision primarily during the early stages of training to guide the radiance field toward a plausible, sharp initialization. As training progresses, we gradually reduce the weight of this loss, allowing the NeRF to refine its reconstruction beyond the model-based estimate, driven by the direct event supervision discussed in the next section.

\subsection{Event-Based Supervision}
\label{sec:event_supervision}

While motion-blurred RGB frames provide strong supervision for radiance field optimization, they are inherently limited by their low temporal resolution and the lack of texture under fast motion. To complement the image and prior-based supervision, we leverage the high temporal resolution of events to directly guide both the radiance field and the camera trajectory refinement. 

Each event $\mathbf{e}_j = (\mathbf{u}_j, t_j, p_j)$ encodes a brightness change of polarity $p_j$ at pixel $\mathbf{u}_j$ and time $t_j$. Unlike RGB frames, these measurements are significantly less affected by motion blur and can thus provide precise, high-frequency supervision. Following prior works \cite{rudnev2022eventnerf,hwang2023ev}, we train the model to predict log-brightness changes that simulate those observed \new{by} the event camera during motion. To this end, we synthesize the log-brightness at each event timestamp via volumetric rendering of the radiance field. 

Specifically, for each event $\new{\mathbf{e}_j} = (\mathbf{u}_j, t_j, p_j)$, we identify its most recent preceding event at the same pixel, denoted $\mathbf{e}_{j^-} = (\mathbf{u}_j, t_{j^-}, p_{j^-})$ with $t_{j^-} < t_j$. We estimate the camera poses at which these events were triggered using the refined trajectory, $\mathbf{T}_\text{ref}(t_j)$ and $\mathbf{T}_\text{ref}(t_{j^-})$ (Sec.~\ref{sec:pose_refinement}), and render the corresponding RGB colors $\mathcal{R}(\mathbf{u}_j, t_j)$ and $\mathcal{R}(\mathbf{u}_j, t_{j^-})$ via volumetric rendering at these timestamps.

\input{floaters/figures/drone_and_camera}

We estimate the corresponding brightness change by converting the rendered colors to log-brightness values using a learnable event camera response function (eCRF) and computing their difference. In particular, we define the log-brightness perceived by the event camera's pixel $\mathbf{u}_j$ at time $t_j$ as:
\begin{equation}
\label{eq:log_brightness}
    \Delta \hat{L}_j = \hat{L}_j \!-\! \hat{L}_{j^-} \,\, \text{with} \,\,
    \hat{L}_j = \log\left(h(\mathrm{eCRF}_\Psi(\mathcal{R}(\mathbf{u}_j, t_j), p_j))\right)
\end{equation}
where $h(\cdot)$ is a luma conversion function, and $\mathrm{eCRF}_\Psi$ is an MLP that models the pixel-wise, polarity-dependent response of the event sensor. This learnable mapping accounts for real-world deviations from the ideal event generation model, such as pixel-level threshold variability and sensor non-linearities.

Finally, we supervise the predicted brightness change to match the expected change recorded by the camera over a batch $\mathcal{B}_e$:
\begin{equation}
\label{eq:loss_event}
    \mathcal{L}_\text{event} = \frac{1}{|\mathcal{B}_e|} \sum_{\mathbf{e}_j \in \mathcal{B}_e} \Big[
\big\| \Delta \hat{L}_{j,c} - \Delta L_j^\text{cam} \big\|_2^2 +
\big\| \Delta \hat{L}_{j,f} - \Delta L_j^\text{cam} \big\|_2^2
\Big],
\end{equation}
where $\Delta L_j^\text{cam} = p_j \Theta_{p_j}$ is the expected change recorded by the camera, $\Theta_{p_j}$ is the contrast threshold corresponding to the event polarity (positive or negative), and the subscripts $c$ and $f$ denote quantities computed using the coarse and fine radiance field renderings, respectively.

This event loss not only promotes sharper reconstructions, particularly in regions with limited RGB support, but also directly supervises the refined trajectory at sub-frame resolution. By querying $\mathbf{T}_\text{ref}(\cdot)$ at individual event timestamps, the model can propagate gradient signals back to the pose refiner module $f_\phi$, contributing to adjusting the trajectory even within frame exposures. This tight coupling of radiance and motion learning enhances overall reconstruction accuracy and consistency, enabling the recovery of sharp geometry even under severe motion blur.

%% file: floaters/figures/architecture.tex
\begin{figure*}
    \centering
    \includegraphics[width=\textwidth]{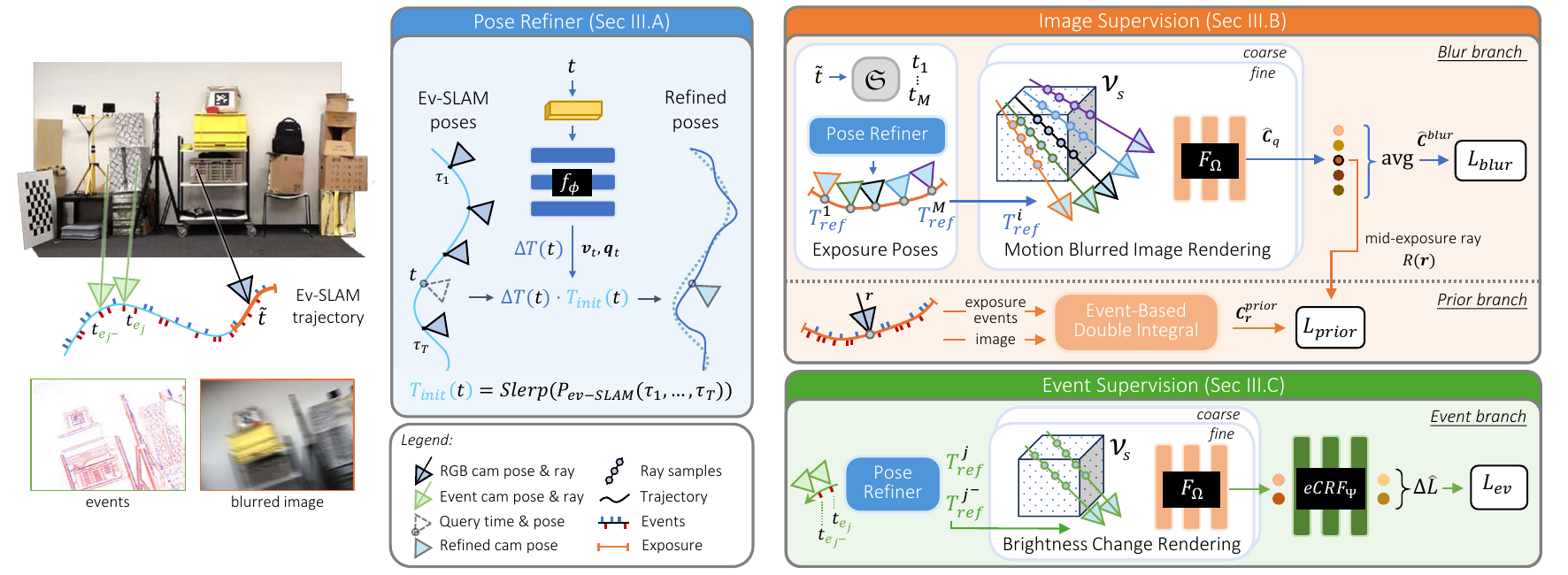}
    \caption{\new{Overview of our proposed architecture. We reconstruct a sharp radiance field from a set of motion-blurred RGB frames and events. 
    The estimated trajectory from an event-based SLAM system, $T_{init}$, is refined through a learned \textit{Pose Refiner} that takes a query timestamp $t$ as input and predicts residual corrections at arbitrary time resolutions. 
    This refined trajectory is used to supervise the radiance field through three complementary branches: 
    a \textit{blur branch} models image formation across the exposure time; an \textit{event branch} supervises high-temporal-resolution brightness changes via a learned event camera response function; and a \textit{prior branch} introduces model-based deblur constraints. All branches share the same learned camera trajectory, allowing events and images to jointly refine the scene representation and motion estimates.}}

    \label{fig:network}
\end{figure*}

%% file: floaters/figures/drone_and_camera.tex
\begin{figure}
  \centering
  \begin{subfigure}{\columnwidth}
    \centering
    \includegraphics[height=0.15\textheight]{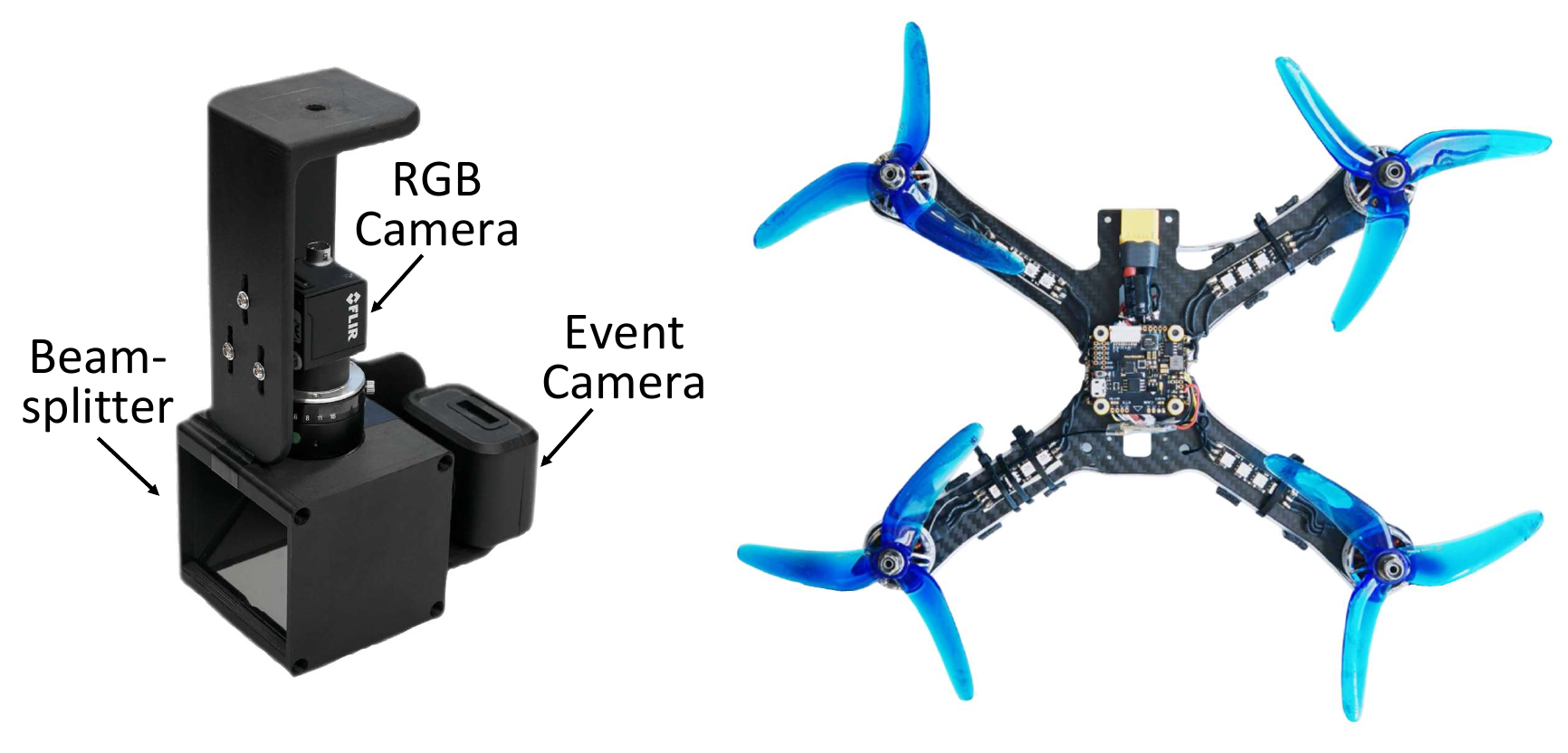}
    \label{fig:drone_view}
  \end{subfigure}%

  \caption{Data collection setup. A beamsplitter with an RGB and an event camera is attached to a quadrotor platform.}
  \label{fig:drone_camera}
\end{figure}

%% file: sections/4_experiments.tex
\input{floaters/tables/evdeblur_blender}

\input{floaters/tables/evdeblur_cdavis}
\section{Experiments}
\label{sec:experiments}

We validate our method on both synthetic and real-world datasets, comparing it against recent image-based and event-based baselines. We begin with synthetic scenarios using Ev-DeblurBlender \cite{cannici2024mitigating}, which provides \new{ground-truth} poses and allows us to ablate components of our network in a controlled setting. We then evaluate performance on real-world data with Ev-DeblurCDAVIS \cite{cannici2024mitigating}, before introducing two new challenging datasets collected under fast motion: Gen3-HandHeld, which tests robustness to varying motion blur, and Gen3-DroneFlight, which represents the most difficult setting with high-speed drone flight data.

\subsection{Implementation Details}

\myparagraph{Training.}
We implement our code using PyTorch.
We train using a batch size of 1024 rays for the blur loss ($\mathcal{B}_b$) and 2048 rays for the event loss ($\mathcal{B}_e$), and sample 64 coarse and 64 fine points along each ray. The number of motion samples $M$ per exposure is scene-dependent and typically ranges from 7 to 11 based on motion complexity. The pose refiner $f_\phi$ is implemented as an 8-layer MLP with 256-dimensional hidden units and ReLU activations, following~\cite{ma2024continuous}, while we follow \cite{cannici2024mitigating} for implementing the fine and coarse MLPs defining the NeRF. We use the Adam optimizer~\cite{kingma2014adam} to minimize the total loss $\mathcal{L} = \lambda_b \mathcal{L}_{\text{blur}} + \lambda_e \mathcal{L}_{\text{event}} + \lambda_{p} \mathcal{L}_{\text{prior}}$ where we set $\lambda_b= 1.0$ and $\lambda_e = 0.1$, while we decay $\lambda_p$ from an initial value of $0.1$ to zero by iteration 20{,}000 with a cosine scheduling to let the NeRF move beyond the model-based prior. We train for 30{,}000 iterations with a learning rate exponentially decaying from $5 \cdot 10^{-3}$ to $5 \cdot 10^{-6}$.
\new{
The total runtime for training a single scene is around 3 hours on one NVIDIA A100 GPU. 
}

\subsection{Datasets}

\myparagraph{Ev-DeblurBlender.}
A synthetic dataset introduced in \cite{cannici2024mitigating}, rendered using high-frame-rate Blender simulations and comprising four scenes: \emph{factory}, \emph{pool}, \emph{tanabata}, and \emph{trolley}. Blurry images are generated by integrating high-FPS frames over 40 ms exposures, while events are simulated using ESIM~\cite{rebecq2018esim} with thresholds $\Theta_{-1}=\Theta_{+1}=0.2$.

\myparagraph{NoisyPose-EvDeblurBlender.}
To evaluate robustness to trajectory errors, we introduce a noisy variant of the Ev-DeblurBlender dataset. 
Starting from ground‑truth poses, we inject synthetic drift by first selecting six uniformly spaced points along each trajectory and sampling at each point a random 6 DoF perturbation. We then linearly interpolate the sampled perturbations to all intermediate poses. The perturbations increase progressively with traveled distance to simulate the behavior of a VIO system affected by drift. We use a global \textit{noise level} $l$ to control the overall magnitude of the perturbations.
Letting $\epsilon^{(l)} = (\epsilon^{(l)}_t, \epsilon^{(l)}_R)$ denote the accumulated error per meter of traveled distance for noise level $l$, 
where $\epsilon^{(l)}_t$ is the translation error (in centimeters) and $\epsilon^{(l)}_R$ the rotation error (in degrees), we test 4 levels to quantify performance degradation across methods, defined as follows for noise levels $l=1$ to $l=4$: $(2\, \text{cm}, 0.2^\circ),\,(4\, \text{cm}, 0.4^\circ),\,(8\, \text{cm}, 0.8^\circ),\,(12\, \text{cm}, 1.2^\circ)$.

\input{floaters/tables/noisy_poses_blender}

\myparagraph{Ev-DeblurCDAVIS.}
A real-world dataset introduced in~\cite{cannici2024mitigating}, which includes ground-truth sharp reference images as well as precise poses for training and evaluation. It uses a Color DAVIS346 sensor to record events and RGB frames (346×260 resolution) with a $100$ ms exposure time. \new{Ground-truth} poses are obtained via a linear slider’s motor encoder, enabling evaluation under controlled 
motion at speeds on the order of 0.1 m/s.
We follow the same train-test split as in \cite{cannici2024mitigating}, using 11 to 18 blurry training views and 5 sharp test views. \looseness=-1

\myparagraph{Gen3-HandHeld and Gen3-DroneFlight.}
We collect two fast-motion datasets using a beamsplitter setup (Figure~\ref{fig:drone_camera}) consisting of a Prophesee Gen3 event camera (640×480 resolution) and a color FLIR Blackfly S camera, both viewing the same scene through the beamsplitter and hardware-synchronized. The Prophesee Gen3 features an IMU that is synchronized with the event stream and used for pose estimation with UltimateSLAM~\cite{vidal2018ultimate}. We run UltimateSLAM~\cite{vidal2018ultimate} on the full trajectory to obtain an initial pose prior, and then subsample the training images to replicate the setup of Ev-DeblurNeRF~\cite{cannici2024mitigating}, using approximately 34 blurred training images and 5 sharp test images per sequence. Gen3-HandHeld features fast motion performed with the rig handheld, moving at variable speeds and with different exposure times for the RGB camera. We use this dataset to evaluate robustness to varying levels of blur severity. Gen3-HandHeld includes three exposure settings (10ms, 30ms and 50ms) and three speed profiles (0.8$\sim$1 m/s, 1.2$\sim$1.4 m/s, 1.4$\sim$1.8 m/s) for a total of 9 sequences. To demonstrate the effectiveness of our approach in real-world scenarios, we record another dataset, named Gen3-DroneFlight, featuring the same beamsplitter setup but mounted on a real quadrotor (shown in Figure~\ref{fig:drone_camera}). We use this setup to perform flight with left-to-right, top-to-down motions in front of target scenes, reaching speeds of up to 2 m/s. We use an external motion capture system to maneuver the drone, as well as to record \new{ground-truth} camera poses for evaluation. 
\new{Crucially, the motion capture poses are never used during training. These poses are used solely for evaluation, as our system only relies on approximate poses from UltimateSLAM~\cite{vidal2018ultimate}.}
\looseness=-1

\subsection{Baselines}
We follow the evaluation setup established in previous work~\cite{cannici2024mitigating}, ensuring consistent comparisons across both frame-only and event-driven baselines. For NeRF-based frame-only deblurring, we consider DeblurNeRF~\cite{ma2022deblur}, BAD-NeRF~\cite{wang2022bad}, DP-NeRF~\cite{lee2023dp}, and PDRF~\cite{peng2022pdrf}. We also include video deblurring approaches (MPRNet~\cite{zamir2021multi}, PVDNet~\cite{son2021recurrent} and EFNet~\cite{sun2022event}), followed by scene reconstruction using NeRF~\cite{mildenhall2020nerf}. 
For EFNet~\cite{sun2022event}, we also evaluate a variant, denoted as EFNet*, where the pretrained model is applied to a given scene after finetuning on the other scenes in the dataset.
Among event-based methods, we evaluate E-NeRF~\cite{klenk2022nerf}, E$^2$-NeRF~\cite{qi2023e2nerf}, and Ev-DeblurNeRF~\cite{cannici2024mitigating}. All baselines are run with official code and default hyperparameters, using the same input data and poses for fairness. Similar to~\cite{cannici2024mitigating}, we perform extensive evaluation in simulation and then select the best-performing baselines for a more in-depth analysis on real-world scenarios.

\subsection{Experimental Validation}
\label{sec:Experimental}

\myparagraph{State-of-the-Art Comparison.}
We begin our evaluation by benchmarking the proposed method on the Ev-DeblurBlender and Ev-DeblurCDAVIS datasets. Both datasets provide ground-truth camera poses, allowing us to fairly compare against baselines that do not perform pose refinement during training. This setting also serves to validate that our method, under ideal conditions, can accurately infer intermediate poses during exposure and at event timestamps and achieve sharp reconstruction quality on par with the best existing approaches.

Results for the synthetic Ev-DeblurBlender dataset are reported in Table~\ref{tab:evdeblur_blender}. Our method consistently outperforms all baselines. It achieves a relative PSNR improvement of 10.7\% over the best-performing image-only NeRF-based method 
and surpasses two-stage approaches that first deblur images and then train a NeRF
by 5.7\% in PSNR. Compared to event-based NeRF methods such as E-NeRF~\cite{klenk2022nerf} and E$^2$-NeRF~\cite{qi2023e2nerf}, our method provides a 17.9\% PSNR gain.

We observe analogous trends on the real-world Ev-DeblurCDAVIS dataset, as shown in Table~\ref{tab:evdeblur_cdavis}. Our method again surpasses frame-only baselines, as well as all event-driven approaches. In particular, we outperform E$^2$-NeRF~\cite{qi2023e2nerf} by {8.4\%} PSNR and perform on par with Ev-DeblurNeRF~\cite{cannici2024mitigating}. Although our method performs on par with Ev-DeblurNeRF~\cite{cannici2024mitigating}, as will be shown in the next section, its advantage becomes more evident as pose noise increases. We provide qualitative comparisons in Figure \ref{fig:cdavis}.

\input{floaters/tables/trajectory_table}
\input{floaters/figures/traj_results_factory_AND_trolley_1.2}

\myparagraph{Robustness to Pose Noise.}
We next evaluate the robustness of our method to noisy camera trajectories using the NoisyPose-EvDeblurBlender dataset, a modified version of Ev-DeblurBlender where training poses are perturbed by synthetic drift. This perturbation simulates the gradual degradation commonly observed in visual-inertial odometry systems during prolonged motion. Results are reported in Table~\ref{tab:denoise_results} and illustrated qualitatively in Figure~\ref{fig:noisy_poses}.

For this study, we compare against Ev-DeblurNeRF~\cite{cannici2024mitigating}, 
the best-performing baseline, which also 
shares the most architectural similarities with our method and thus serves as the most meaningful baseline to isolate the effect of trajectory refinement. Our pose refinement strategy is fully learned and unconstrained by fixed coordinate frames. As such, it can result in globally shifted or rotated trajectories that are photometrically valid but misaligned with the \new{ground-truth} frame. To account for this, we first align \new{ground-truth} poses to the learned trajectory via Procrustes alignment on the training views. We then further register the test poses to the learned NeRF by minimizing a photometric loss between the reference and rendered views, following BAD-NeRF protocol~\cite{wang2022bad}.

Ev-DeblurNeRF~\cite{cannici2024mitigating} struggles to recover meaningful geometry, with PSNR dropping to 14.82 dB on average in the mildest noise case (see Table~\ref{tab:denoise_results}), compared to the 30.42 dB achieved when using \new{ground-truth} poses (see Table~\ref{tab:evdeblur_blender}). In contrast, our method consistently compensates for the drift, achieving an average of 29.88 dB PSNR across scenes. As the noise level increases, performance degrades but remains robust even in the most challenging setting, yielding an average of 26.49 dB despite substantial initial trajectory drift. Notably, in the longest sequences, \emph{trolley} and \emph{tanabata}, where drift affects the perturbed trajectory the most (108.12 ATE and 99.50 ATE, respectively; see Table~\ref{tab:pose_results}), our method still successfully recovers geometry, achieving 25.93 dB and 23.97 dB PSNR.

The impact of our joint pose refiner module is further demonstrated in Table~\ref{tab:pose_results}, where we analyze the trajectory error of the poses recovered by our method after the full training procedure has converged. We denote the noisy input trajectory as \textit{perturbed}, and report performance for \textit{frame only} and \textit{frame + events} configurations, where the event-based branch of our network is disabled or enabled, respectively. We measure Absolute Trajectory Error RMSE (ATE RMSE) in centimeters, Relative Pose Error in translation (RPE$_\text{trans}$) as a percentage of the traveled distance, and Relative Pose Error in rotation (RPE$_\text{rot}$) in degrees per meter, using the toolbox proposed in \cite{zhang18tutorial}. For RPE, we divide the ground‑truth trajectory into six equal‑length segments, use the five segment midpoints as reference poses, compute both translational and rotational errors over the segment length at each reference pose, and average the five resulting errors. 

\input{floaters/figures/qualitative_cdavis}

\input{floaters/tables/bslarge_handheld}

The results in Table~\ref{tab:pose_results} show that combining both frames and events leads to significantly improved performance, as the pose refiner benefits from supervision not only from the frames but also from the high-temporal-resolution information provided by events. On average, using frames alone improves the initial perturbed trajectory from 67.69\,cm to 18.22\,cm ATE in the most challenging setting, while incorporating events yields a further improvement, lowering ATE to only 7.47\,cm. When the initial poses are within 30\,cm (noise levels 1 and 2), 
our full method with events
achieves sub-centimeter ATE, highlighting the contribution of the event-based supervision. 

While precise trajectory recovery is not the primary objective of our method, these results demonstrate its ability to achieve accurate poses by exploiting dense 3D reconstruction as a way to jointly refine poses during training. 
Additional results are provided in Figure~\ref{fig:traj_combined}, where we qualitatively visualize the refined trajectories and report the changes in trajectory error with the distance traveled. 

\input{floaters/figures/qualitative_noisy_blender}
\input{floaters/figures/qualitative_bsgen3_handheld}

\myparagraph{Gen3-HandHeld Results.}
We continue our evaluation on real-world sequences captured via a \new{beamsplitter} setup. Compared to the CDAVIS dataset, this setup presents greater challenges, as it consists of two independent sensors with distinct responses to light. This necessitates modeling these differences explicitly, which we achieve through the learnable camera response function (CRF).  We first focus on the Gen3-HandHeld dataset, which comprises handheld scenes recorded with varying camera motion speeds and RGB exposure times. 
To ensure comparable brightness across settings, we manually adjust the lens aperture when increasing exposure durations. Camera poses are initially estimated using UltimateSLAM~\cite{vidal2018ultimate} and subsequently refined during training. For comparison, we include the best-performing event-based baselines, namely Ev-DeblurNeRF~\cite{cannici2024mitigating} and E$^2$NeRF~\cite{qi2023e2nerf}. To ensure a fair comparison, we provide these baselines with the refined poses obtained from our method at convergence, representing a best-case scenario where they benefit from more accurate trajectories, although such supervision would not be available in practice. To further highlight the contribution of our joint pose refinement, we also report results for Ev-DeblurNeRF~\cite{cannici2024mitigating} trained directly on the original UltimateSLAM~\cite{vidal2018ultimate} poses.

The results are shown in Table~\ref{tab:bslarge_handheld}. 
Despite using noisy UltimateSLAM poses~\cite{vidal2018ultimate} as initialization, our method \new{maintains} high performance across different exposure times and speed configurations, achieving 21.31 dB PSNR in the most challenging setting. 
In contrast, Ev-DeblurNeRF~\cite{cannici2024mitigating} using the same poses fails to produce competitive results in all settings due to the lack of a pose refinement stage. 
The performance gap between Ev-DeblurNeRF~\cite{cannici2024mitigating} and our method is consistently greater than 7 dB in PSNR.
Crucially, even when provided with our refined poses, both Ev-DeblurNeRF~\cite{cannici2024mitigating} and E$^2$NeRF~\cite{qi2023e2nerf} baselines consistently underperform relative to our approach. The first is limited to linearly interpolating poses at event timestamps, while the second only considers events during image exposures, thus underutilizing high-frequency supervision. Qualitative results illustrating these findings are in Figure~\ref{fig:bsgen3_handheld}.

\myparagraph{Gen3-DroneFlight Results.} We present the results for drone flight sequences in Table~\ref{tab:gen3_eval} and Figure~\ref{fig:bsgen3_drone}.
These results conclude our analysis by demonstrating the successful deployment of our algorithm on data collected onboard a real aerial robotic platform. The Gen3-DroneFlight dataset presents the most challenging scenarios in our evaluation: aerial sequences recorded at 30 ms exposure, where the drone follows a zig-zag trajectory in front of three different scenes (a visual representation is provided in Figure~\ref{fig:teaser_wide}). This motion induces substantial lateral blur across the entire image, unlike circular trajectories where regions near the rotation center remain relatively sharp. The challenge is further amplified by high-frequency vibrations from the drone’s motors, which introduce additional disturbances to both images and events.

From Table~\ref{tab:gen3_eval} we can see that our method maintains the best performance across varying scenes and speed profiles.
We continue to outperform the Ev-DeblurNeRF~\cite{cannici2024mitigating} and E$^2$NeRF~\cite{qi2023e2nerf} baselines, even when they are provided with our refined poses as input. While in this case, the baselines benefit from starting with accurate trajectories, our method begins from noisy, inaccurate poses and must learn to refine them during training, thus experiencing not only motion blur but also visual distortions caused by inaccurate camera trajectories. 
Despite this additional challenge, our method achieves average improvements of 4.3\% in PSNR, 7.1\% in LPIPS, and 6\% in SSIM, highlighting the strength of our joint trajectory and radiance field optimization. 
When provided with our refined poses as initialization, our method achieves even better reconstruction quality.
Qualitative results are provided in Figure~\ref{fig:bsgen3_drone}, where our method clearly recovers fine textures and structural details that are lost in the baseline reconstructions.

\input{floaters/figures/qualitative_bsgen3_drone_flight}
\new{
\myparagraph{Robustness to Trajectory Length.} 
We evaluate the robustness of our approach to varying trajectory lengths using the Gen3-DroneFlight dataset. The trajectories in scenes \textsc{BoxStack}, \textsc{Equipment}, and \textsc{PlayCorner} measure approximately 7.5–8 meters, while those in scenes \textsc{Models}, \textsc{ReadingCorner}, and \textsc{Lounge} are roughly twice as long, at 15–16 meters.
The results in Table~\ref{tab:gen3_eval} indicate that our method reliably outperforms all baselines across trajectories of both lengths.

Results from different scenes are not directly comparable due to scene-specific factors such as different textures and object layout. To mitigate these scene-dependent variations and enable a more reliable scalability evaluation, we captured a larger-scale scene covering the combined area of the \textsc{Models} and \textsc{ReadingCorner} scenes. The drone was flown along a 30-meter trajectory at the fastest speed profile (maximum speed of 2 m/s). The results are reported in Table~\ref{tab:traj_length}. As shown, our method maintains stable reconstruction quality as the trajectory length increases, demonstrating robustness to larger-scale trajectories.
}

\new{
\myparagraph{Component Validation and Ablation Study.} 
We validate the technical components of our method on the \textsc{playCorner} scene from the Gen3-DroneFlight dataset. Specifically, we ablate the loss terms in Eq.~\ref{eq:loss_blur} (blur loss $\mathcal{L}_\text{blur}$), Eq.~\ref{eq:loss-prior} (prior loss $\mathcal{L}_\text{prior}$) and Eq.~\ref{eq:loss_event} (event loss $\mathcal{L}_\text{event}$), and we experimentally evaluate the eCRF module of Eq.~\ref{eq:log_brightness}. Note that we do not include an event-only variant, since events lack absolute intensity and color information and a radiance field trained solely with event supervision is not comparable to RGB-supervised variants. Therefore, we demonstrate the contribution of event supervision by adding or removing the event loss to models trained with at least one RGB loss. Results are presented in Table~\ref{tab:ablations}.

Adding the event loss on top of the blur loss yields large improvements (+20\% PSNR, +21\% LPIPS, +31\% SSIM on average), confirming the strong benefit of event supervision.
When using only the prior loss, compared to using only the blur loss, the performance gap increases significantly from +0.21dB at 1 m/s to +5.51dB in the 2 m/s case, with an average gap of +2.53dB PSNR, demonstrating the usefulness of the pseudo ground-truth target obtained from the model-based deblurring process, particularly in high-speed scenarios.
Similarly, after adding the event loss on top of the prior loss, the average performance increases considerably (+7\% PSNR, +21\% LPIPS and +17\% SSIM), which again demonstrates the significance of event supervision.
Combining the blur loss and the prior loss yields better results than using either loss alone. Nonetheless, configurations that include the event loss (e.g., blur+event or prior+event) consistently outperform the blur+prior combination.
The highest performance is achieved when all three loss terms are used jointly.
Finally, removing the eCRF module leads to noticeable performance degradation (-2\% PSNR, -11\% LPIPS, -3\% SSIM), demonstrating the effectiveness of modeling the event camera response function.
}
\input{floaters/tables/bslarge_drone}

\input{floaters/tables/traj_length}
\input{floaters/tables/ablations}

%% file: floaters/tables/evdeblur_blender.tex
\begin{table*}
\caption{Quantitative comparison on the synthetic Ev-DeblurBlender dataset. Best results are reported in bold.}
\label{tab:evdeblur_blender}

\centering
\resizebox{\textwidth}{!}{
\setlength{\tabcolsep}{4pt}
\begin{tabular}{l ccc ccc ccc ccc ccc}
\toprule
\multirow{2}{*}{Method}& \multicolumn{3}{c}{\textsc{Factory}} & \multicolumn{3}{c}{\textsc{Pool}} & \multicolumn{3}{c}{\textsc{Tanabata}} & \multicolumn{3}{c}{\textsc{Trolley}} & \multicolumn{3}{c}{\textsc{Average}} \\
    \cmidrule(lr){2-4} \cmidrule(lr){5-7} \cmidrule(lr){8-10} \cmidrule(lr){11-13} \cmidrule(lr){14-16}

 & PSNR$\uparrow$ & LPIPS$\downarrow$ & SSIM$\uparrow$ & PSNR$\uparrow$ & LPIPS$\downarrow$ & SSIM$\uparrow$ & PSNR$\uparrow$ & LPIPS$\downarrow$ & SSIM$\uparrow$ & PSNR$\uparrow$ & LPIPS$\downarrow$ & SSIM$\uparrow$ & PSNR$\uparrow$ & LPIPS$\downarrow$ & SSIM$\uparrow$ \\
\midrule
DeblurNeRF~\cite{ma2022deblur} & 24.52 & 0.25 & 0.79 & 26.02 & 0.34 & 0.69 & 21.38 & 0.28 & 0.71 & 23.58 & 0.22 & 0.79 & 23.87 & 0.27 & 0.74 \\
BAD-NeRF~\cite{wang2022bad} & 21.20 & 0.22 & 0.64 & 27.13 & 0.23 & 0.70 & 20.89 & 0.25 & 0.65 & 22.76 & 0.18 & 0.73 & 22.99 & 0.22 & 0.68 \\
PDRF~\cite{peng2022pdrf} & 27.34 & 0.17 & 0.87 & 27.46 & 0.32 & 0.72 & 24.27 & 0.20 & 0.81 & 26.09 & 0.15 & 0.86 & 26.29 & 0.21 & 0.81 \\
DP-NeRF~\cite{lee2023dp} & 26.77 & 0.20 & 0.85 & 29.58 & 0.24 & 0.79 & 27.32 & 0.11 & 0.85 & 27.04 & 0.14 & 0.87 & 27.68 & 0.17 & 0.84 \\
\midrule
MPRNet~\cite{zamir2021multi} + NeRF & 19.09 & 0.37 & 0.56 & 25.49 & 0.39 & 0.64 & 17.79 & 0.42 & 0.51 & 19.82 & 0.31 & 0.62 & 20.55 & 0.37 & 0.58 \\
PVDNet~\cite{son2021recurrent} + NeRF & 22.50 & 0.29 & 0.71 & 23.89 & 0.43 & 0.52 & 20.26 & 0.33 & 0.64 & 22.49 & 0.25 & 0.74 & 22.28 & 0.32 & 0.65 \\
EFNet~\cite{sun2022event} + NeRF & 20.91 & 0.32 & 0.63 & 27.03 & 0.31 & 0.73 & 20.68 & 0.31 & 0.64 & 21.69 & 0.25 & 0.69 & 22.58 & 0.30 & 0.67 \\
EFNet*~\cite{sun2022event} + NeRF & 29.01 & 0.14 & 0.87 & 29.77 & 0.18 & 0.80 & 27.76 & 0.11 & 0.87 & 29.40 & 0.09 & 0.89 & 28.99 & 0.13 & 0.86 \\
\midrule
ENeRF~\cite{klenk2022nerf} & 22.46 & 0.19 & 0.79 & 25.51 & 0.28 & 0.72 & 22.97 & 0.16 & 0.83 & 21.07 & 0.20 & 0.80 & 23.00 & 0.21 & 0.79 \\
E$^2$NeRF~\cite{qi2023e2nerf} & 24.90 & 0.17 & 0.78 & 29.57 & 0.18 & 0.78 & 23.06 & 0.19 & 0.74 & 26.49 & 0.10 & 0.85 & 26.00 & 0.16 & 0.78 \\
Ev-DeblurNeRF~\cite{cannici2024mitigating} & 31.79 & 0.06 & 0.93 & \textbf{31.51} & 0.14 & 0.84 & 28.67 & 0.08 & 0.90 & 29.72 & 0.07 & 0.92 & 30.42 & 0.08 & 0.90 \\
\midrule
Ours & \textbf{32.37} & \textbf{0.06} & \textbf{0.94} & 31.13 & \textbf{0.14} & \textbf{0.84} & \textbf{29.06} & \textbf{0.07} & \textbf{0.90} & \textbf{30.04} & \textbf{0.06} & \textbf{0.92} & \textbf{30.65} & \textbf{0.08} & \textbf{0.90} \\
\bottomrule
\end{tabular}

}
\end{table*}

%% file: floaters/tables/evdeblur_cdavis.tex
\begin{table*}[!t]
\caption{Quantitative comparison on the real-world Ev-DeblurCDAVIS dataset. Best results are reported in bold.}
\label{tab:evdeblur_cdavis}

\centering
\resizebox{\textwidth}{!}{
\setlength{\tabcolsep}{1pt}
\begin{tabular}{l ccc ccc ccc ccc ccc ccc}
\toprule
\multirow{2}{*}{Method} & \multicolumn{3}{c}{\textsc{Batteries}} & \multicolumn{3}{c}{\textsc{Power supplies}} & \multicolumn{3}{c}{\textsc{Lab Equipment}} & \multicolumn{3}{c}{\textsc{Drones}} & \multicolumn{3}{c}{\textsc{Figures}} & \multicolumn{3}{c}{\textsc{Average}} \\
    \cmidrule(lr){2-4} \cmidrule(lr){5-7} \cmidrule(lr){8-10} \cmidrule(lr){11-13} \cmidrule(lr){14-16} \cmidrule(lr){17-19}

 & PSNR$\uparrow$ & LPIPS$\downarrow$ & SSIM$\uparrow$ & PSNR$\uparrow$ & LPIPS$\downarrow$ & SSIM$\uparrow$ & PSNR$\uparrow$ & LPIPS$\downarrow$ & SSIM$\uparrow$ & PSNR$\uparrow$ & LPIPS$\downarrow$ & SSIM$\uparrow$ & PSNR$\uparrow$ & LPIPS$\downarrow$ & SSIM$\uparrow$ & PSNR$\uparrow$ & LPIPS$\downarrow$ & SSIM$\uparrow$ \\
\midrule
DP-NeRF+TensoRF~\cite{chen2022tensorf} & 26.64 & 0.27 & 0.81 & 25.74 & 0.32 & 0.77 & 27.49 & 0.31 & 0.80 & 26.52 & 0.30 & 0.81 & 27.76 & 0.34 & 0.77 & 26.83 & 0.31 & 0.79 \\
EDI~\cite{pan2019bringing} + NeRF & 28.66 & 0.12 & 0.87 & 28.16 & 0.09 & 0.88 & 31.45 & 0.13 & 0.89 & 29.37 & 0.10 & 0.88 & 31.44 & 0.12 & 0.88 & 29.82 & 0.11 & 0.88 \\
\midrule
E$^2$NeRF~\cite{qi2023e2nerf} & 30.57 & 0.12 & 0.88 & 29.98 & 0.11 & 0.87 & 30.41 & 0.16 & 0.86 & 30.41 & 0.14 & 0.87 & 31.03 & 0.14 & 0.85 & 30.48 & 0.13 & 0.87 \\
Ev-DeblurNeRF~\cite{cannici2024mitigating} & 33.17 & \textbf{0.05} & 0.92 & 32.35 & \textbf{0.06} & 0.91 & \textbf{33.01} & \textbf{0.08} & \textbf{0.91} & 32.89 & \textbf{0.05} & 0.92 & 33.39 & \textbf{0.07} & 0.90 & 32.96 & \textbf{0.06} & 0.91 \\
\midrule
Ours & \textbf{33.20} & 0.08 & \textbf{0.92} & \textbf{32.49} & 0.07 & \textbf{0.91} & 32.83 & 0.11 & 0.91 & \textbf{33.00} & 0.07 & \textbf{0.92} & \textbf{33.63} & 0.11 & \textbf{0.90} & \textbf{33.03} & 0.09 & \textbf{0.91} \\
\bottomrule
\end{tabular}
}
\end{table*}

%% file: floaters/tables/noisy_poses_blender.tex
\begin{table*}
  \centering
    \caption{Quantitative comparison on the NoisyPose-EvDeblurBlender dataset under different pose noise levels. Best results are reported in bold.}
  \label{tab:denoise_results}
  \setlength{\tabcolsep}{3pt}
  \begin{adjustbox}{max width=\textwidth}
  \begin{tabular}{@{} c c ccc ccc ccc ccc ccc @{}}
    \toprule
    \multirow{2}{*}{\shortstack{Noise \\Level}} & \multirow{2}{*}{Method} & \multicolumn{3}{c}{\textsc{Factory}} & \multicolumn{3}{c}{\textsc{Pool}} & \multicolumn{3}{c}{\textsc{Tanabata}} & \multicolumn{3}{c}{\textsc{Trolley}} & \multicolumn{3}{c}{\textsc{Average}} \\
    \cmidrule(lr){3-5} \cmidrule(lr){6-8} \cmidrule(lr){9-11} \cmidrule(lr){12-14} \cmidrule(lr){15-17}
    & & PSNR$\uparrow$ & LPIPS$\downarrow$ & SSIM$\uparrow$ & PSNR$\uparrow$ & LPIPS$\downarrow$ & SSIM$\uparrow$ & PSNR$\uparrow$ & LPIPS$\downarrow$ & SSIM$\uparrow$ & PSNR$\uparrow$ & LPIPS$\downarrow$ & SSIM$\uparrow$ & PSNR$\uparrow$ & LPIPS$\downarrow$ & SSIM$\uparrow$ \\
    \midrule
    \multirow{2}{*}{1} 
    & Ev-DeblurNeRF~\cite{cannici2024mitigating} 
    & 12.67 & 0.40 & 0.19 & 21.07 & 0.38 & 0.36 & 13.40 & 0.39 & 0.23 & 12.15 & 0.52 & 0.21 & 14.82 & 0.42 & 0.25 \\

    & Ours (Frame Only)
    & 28.57 & 0.15 & 0.89 
    & 29.73 & 0.23 & 0.79 
    & 25.28 & 0.17 & 0.83 
    & 27.65 & 0.11 & 0.89 
    & 27.81 & 0.17 & 0.85 \\
    
    & Ours (Frame + Event)         
    & \textbf{31.11} & \textbf{0.07} & \textbf{0.92} 
    & \textbf{30.89} & \textbf{0.15} & \textbf{0.83} 
    & \textbf{28.01} & \textbf{0.09} & \textbf{0.89} 
    & \textbf{29.51} & \textbf{0.07} & \textbf{0.91} 
    & \textbf{29.88} & \textbf{0.09} & \textbf{0.89} \\
    \midrule
    \multirow{2}{*}{2} 
    & Ev-DeblurNeRF~\cite{cannici2024mitigating} 
    & 11.30 & 0.55 & 0.12 & 17.66 & 0.57 & 0.18 & 12.73 & 0.64 & 0.17 & 10.61 & 0.62 & 0.12 & 13.08 & 0.60 & 0.15 \\

    & Ours (Frame Only)
    & 28.15 & 0.16 & 0.88 
    & 29.53 & 0.25 & 0.78 
    & 25.01 & 0.17 & 0.82 
    & 27.18 & 0.12 & 0.88 
    & 27.47 & 0.17 & 0.84 \\
    
    & Ours (Frame + Event)          
    & \textbf{31.23} & \textbf{0.06} & \textbf{0.92} 
    & \textbf{30.68} & \textbf{0.16} & \textbf{0.83} 
    & \textbf{27.95} & \textbf{0.10} & \textbf{0.87} 
    & \textbf{29.25} & \textbf{0.07} & \textbf{0.91} 
    & \textbf{29.78} & \textbf{0.10} & \textbf{0.88} \\
    \midrule
    \multirow{2}{*}{3}
    & Ev-DeblurNeRF~\cite{cannici2024mitigating} 
    & 10.33 & 0.63 & 0.09 & 15.80 & 0.71 & 0.13 & 11.29 & 0.71 & 0.11 &  9.13 & 0.66 & 0.07 & 11.64 & 0.68 & 0.10 \\

    & Ours (Frame Only)
    & 27.84 & 0.16 & 0.88 
    & 29.11 & 0.26 & 0.77 
    & 21.49 & 0.66 & 0.27 
    & 25.94 & 0.15 & 0.84 
    & 26.10 & 0.31 & 0.69 \\
    
    & Ours (Frame + Event)            
    & \textbf{31.31} & \textbf{0.06} & \textbf{0.93} 
    & \textbf{30.25} & \textbf{0.18} & \textbf{0.81} 
    & \textbf{27.08} & \textbf{0.12} & \textbf{0.85} 
    & \textbf{27.57} & \textbf{0.10} & \textbf{0.87} 
    & \textbf{29.05} & \textbf{0.12} & \textbf{0.86} \\
    \midrule
    \multirow{2}{*}{4}
    & Ev-DeblurNeRF~\cite{cannici2024mitigating} 

    & 9.99 & 0.67 & 0.08 & 15.81 & 0.72 & 0.12 & 10.81 & 0.75 & 0.08 &  8.80 & 0.71 & 0.05 & 11.35 & 0.71 & 0.08 \\

    & Ours (Frame Only)
    & 18.34 & 0.33 & 0.46 
    & 20.43 & 0.56 & 0.35 
    & 20.05 & 0.59 & 0.31 
    & 18.84 & 0.33 & 0.52 
    & 19.42 & 0.45 & 0.41 \\
    
    & Ours (Frame + Event)           
    & \textbf{30.79} & \textbf{0.07} & \textbf{0.92} 
    & \textbf{25.25} & \textbf{0.35} & \textbf{0.58} 
    & \textbf{23.97} & \textbf{0.21} & \textbf{0.74} 
    & \textbf{25.93} & \textbf{0.13} & \textbf{0.83} 
    & \textbf{26.49} & \textbf{0.19} & \textbf{0.77} \\
    \bottomrule
  \end{tabular}
  \end{adjustbox}
\end{table*}

%% file: floaters/tables/trajectory_table.tex
\begin{table*}[t]
  \centering
  \caption{Pose estimation error metrics under different pose noise levels on NoisyPose-EvDeblurBlender. (ATE RMSE [cm] $\downarrow$, RPE translation [\%] $\downarrow$, RPE rotation [deg/m] $\downarrow$)}
  \label{tab:pose_results}
  \setlength{\tabcolsep}{3pt}
  \begin{adjustbox}{max width=\textwidth}
  \begin{tabular}{@{} c c
      ccc
      ccc
      ccc
      ccc
      ccc
    @{}}
    \toprule
    \multirow{2}{*}{\shortstack{Noise \\Level}} & \multirow{2}{*}{Method} & \multicolumn{3}{c}{\textsc{Factory}} & \multicolumn{3}{c}{\textsc{Pool}} & \multicolumn{3}{c}{\textsc{Tanabata}} & \multicolumn{3}{c}{\textsc{Trolley}} & \multicolumn{3}{c}{\textsc{Average}} \\
    \cmidrule(lr){3-5} \cmidrule(lr){6-8} \cmidrule(lr){9-11} \cmidrule(lr){12-14} \cmidrule(lr){15-17}
    &
    
     & ATE & RPE\textsubscript{trans} & RPE\textsubscript{rot} 
     & ATE & RPE\textsubscript{trans} & RPE\textsubscript{rot}
     & ATE & RPE\textsubscript{trans} & RPE\textsubscript{rot}
     & ATE & RPE\textsubscript{trans} & RPE\textsubscript{rot}
     & ATE & RPE\textsubscript{trans} & RPE\textsubscript{rot}\\
    \midrule
\multirow{3}{*}{1}
& Perturbed
  & 9.96  & 5.41  & 0.23
  & 4.73  & 6.49  & 1.62
  & 21.55 & 6.61  & 0.19
  & 23.19 & 10.08 & 0.32
  & 14.86 & 7.15  & 0.59 \\

& Frame Only
  & 11.13 & 0.83  & 0.12
  & 0.99  & 0.76  & 0.26
  & 2.57  & 0.74  & 0.18
  & 3.11  & 0.98  & 0.14
  & 4.45  & 0.83  & 0.18 \\

& Frame + Event
  & \textbf{0.21}  & \textbf{0.12}  & \textbf{0.03}
& \textbf{0.14}  & \textbf{0.16}  & \textbf{0.13}
& \textbf{0.58}  & \textbf{0.21}  & \textbf{0.03}
& \textbf{0.87}  & \textbf{0.24}  & \textbf{0.08}
& \textbf{0.45}  & \textbf{0.18}  & \textbf{0.07} \\

\midrule
\multirow{3}{*}{2}
& Perturbed
  & 18.54 & 10.46 & 0.46
  & 8.88  & 12.21 & 3.23
  & 41.42 & 12.78 & 0.39
  & 45.51 & 20.04 & 0.63
  & 28.59 & 13.87 & 1.18 \\

& Frame Only
  & 1.14  & 0.79  & 0.12
  & 0.81  & 0.79  & 0.26
  & 4.46  & 0.85  & 0.14
  & 3.11  & 1.03  & 0.29
  & 2.38  & 0.86  & 0.20 \\

& Frame + Event
  & \textbf{0.32}  & \textbf{0.14}  & \textbf{0.05}
& \textbf{0.14}  & \textbf{0.17}  & \textbf{0.10}
& \textbf{1.02}  & \textbf{0.34}  & \textbf{0.11}
& \textbf{1.25}  & \textbf{0.33}  & \textbf{0.13}
& \textbf{0.68}  & \textbf{0.25}  & \textbf{0.10} \\

\midrule
\multirow{3}{*}{3}
& Perturbed
  & 32.20 & 19.03 & 0.91
  & 15.55 & 21.23 & 6.46
  & 74.71 & 23.01 & 0.77
  & 83.10 & 36.16 & 1.27
  & 51.39 & 24.86 & 2.35 \\

& Frame Only
  & 6.25  & 0.97  & 0.23
  & 4.24  & 1.73  & 0.27
  & 10.13 & 3.68  & 1.17
  & 4.76  & 1.48  & 0.53
  & 6.35  & 1.96  & 0.55 \\

& Frame + Event
& \textbf{0.52}  & \textbf{0.26}  & \textbf{0.10}
& \textbf{1.36}  & \textbf{1.10}  & \textbf{0.10}
& \textbf{3.02}  & \textbf{0.79}  & \textbf{0.27}
& \textbf{3.23}  & \textbf{0.87}  & \textbf{0.42}
& \textbf{2.03}  & \textbf{0.75}  & \textbf{0.22} \\

\midrule
\multirow{3}{*}{4}
& Perturbed
  & 42.69 & 25.63 & 1.37
  & 20.43 & 27.58 & 9.70
  & 99.50 & 30.31 & 1.16
  & 108.12& 45.55 & 1.90
  & 67.69 & 32.27 & 3.53 \\

& Frame Only
  & 14.26 & 5.80  & 2.21
  & 20.11 & 25.52 & 0.90
  & 15.52 & 4.83  & 1.40
  & 22.98 & 9.40  & 1.40
  & 18.22 & 11.39 & 1.48 \\

& Frame + Event
  & \textbf{1.00}  & \textbf{0.50}  & \textbf{0.20}
& \textbf{15.21} & \textbf{16.99} & \textbf{0.73}
& \textbf{8.70}  & \textbf{2.50}  & \textbf{0.81}
& \textbf{4.96}  & \textbf{1.60}  & \textbf{0.82}
& \textbf{7.47}  & \textbf{5.40}  & \textbf{0.64} \\
    \bottomrule
  \end{tabular}
  \end{adjustbox}
\end{table*}

%% file: floaters/figures/traj_results_factory_AND_trolley_1.2.tex
\begin{figure*}
  \centering

  \begin{minipage}[t]{0.02\linewidth}
  \centering
  \raisebox{-2cm}{\rotatebox[origin=t]{90}{{\scriptsize Factory}}}
  \end{minipage}
  \begin{minipage}[t]{0.97\linewidth}
    \centering
    \begin{minipage}[t]{0.62\linewidth}
    \vspace{0pt}
    \centering
    \includegraphics[width=0.48\linewidth, trim=7 8 10 10, clip]{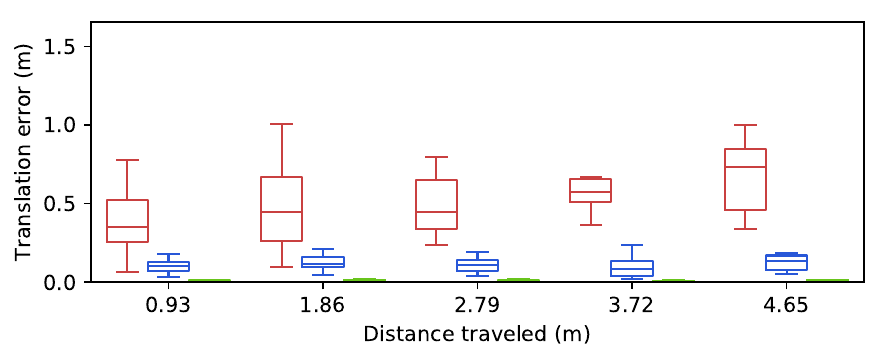}%
    \hfill
    \includegraphics[width=0.48\linewidth, trim=7 8 10 10, clip]{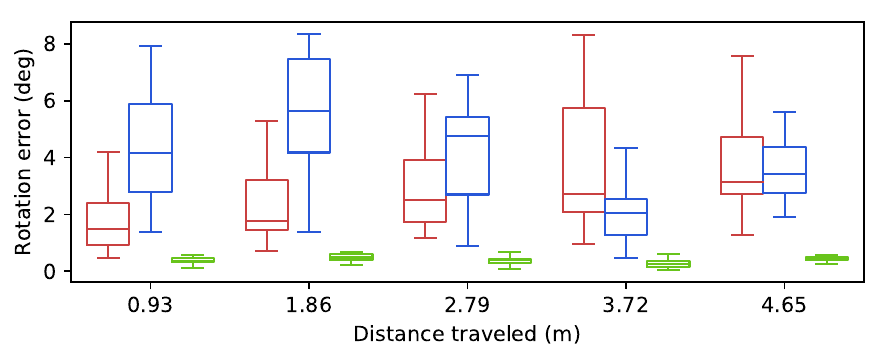}\\[2pt]
    \includegraphics[width=0.48\linewidth, trim=7 8 10 10, clip]{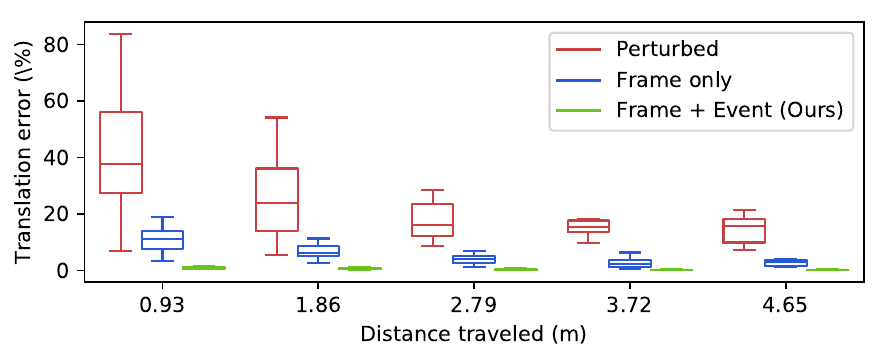}%
    \hfill
    \includegraphics[width=0.48\linewidth, trim=7 8 10 10, clip]{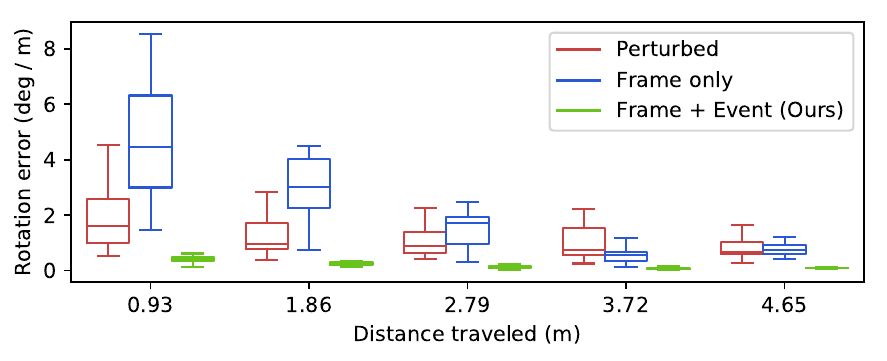}
  \end{minipage}%
  \hfill
  \begin{minipage}[t]{0.36\linewidth}
  \vspace{0pt}
    \centering
    \includegraphics[height=0.65\linewidth, trim=9 8 10 10, clip]{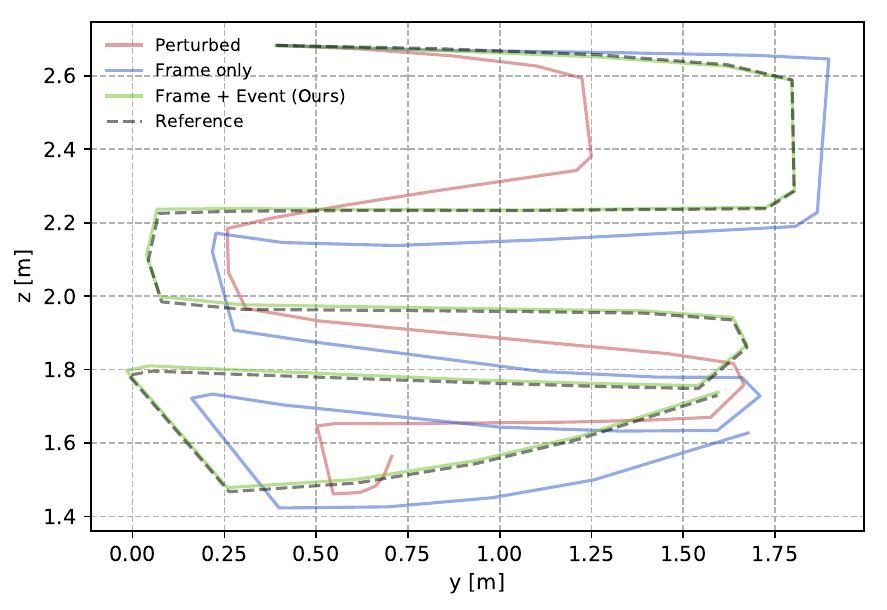}
  \end{minipage}
  \end{minipage}

  \vspace{0.5em}
  \noindent\makebox[\linewidth]{\rule{\linewidth}{0.4pt}}

\begin{minipage}[t]{0.02\linewidth}
  \centering
  \raisebox{-2cm}{\rotatebox[origin=t]{90}{{\scriptsize Trolley}}}
  \end{minipage}
  \begin{minipage}[t]{0.97\linewidth}
    \centering
    \begin{minipage}[t]{0.62\linewidth}
    \vspace{0pt}
    \centering
    \includegraphics[width=0.48\linewidth, trim=7 8 10 10, clip]{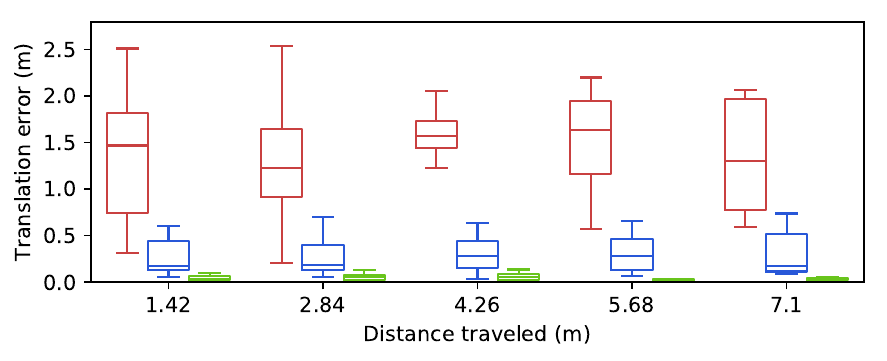}%
    \hfill
    \includegraphics[width=0.48\linewidth, trim=7 8 10 10, clip]{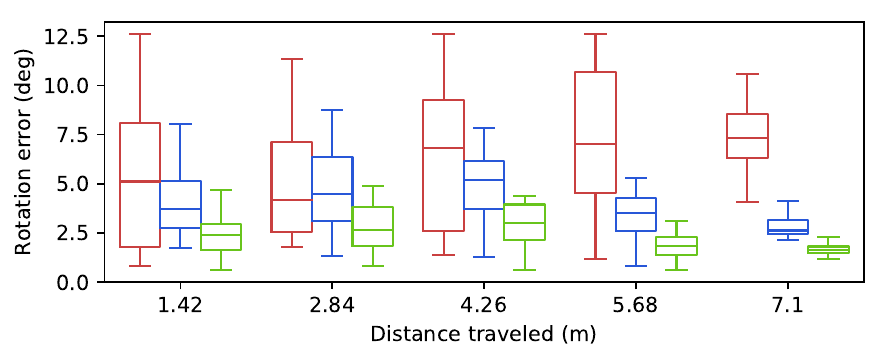}\\[2pt]
    \includegraphics[width=0.48\linewidth, trim=7 8 10 10, clip]{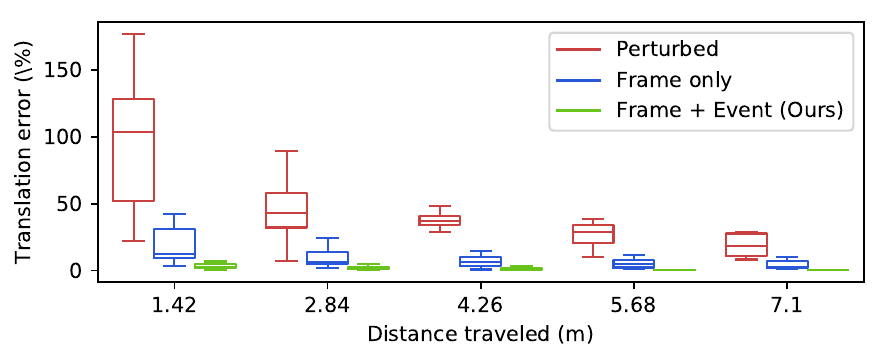}%
    \hfill
    \includegraphics[width=0.48\linewidth, trim=7 8 10 10, clip]{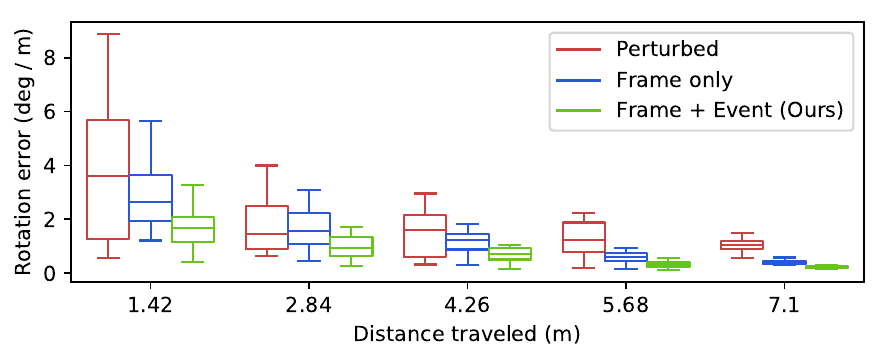}
  \end{minipage}%
  \hfill
  \begin{minipage}[t]{0.36\linewidth}
  \vspace{0pt}
    \centering
    \includegraphics[height=0.62\linewidth, trim=9 8 10 10, clip]{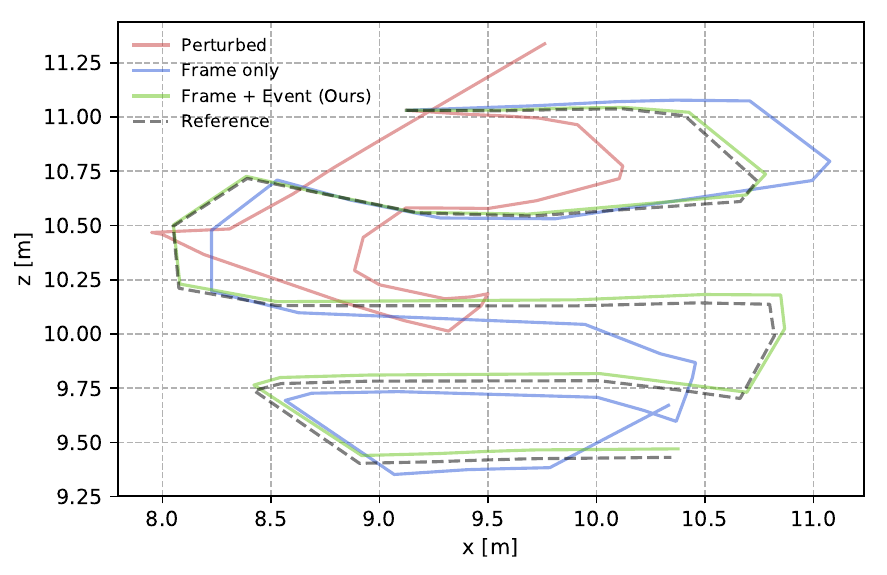}
  \end{minipage}
  \end{minipage}

  \caption{Factory (top) and Trolley (bottom) trajectory analysis on NoisyPose-EvDeblurBlender at noise level 4. We present the change of trajectory errors with traveled distance (left), for both translation (meters and percentage) and rotation (degrees and degrees per meter), along with a visual comparison of the trajectories (right).}
  \label{fig:traj_combined}
\end{figure*}

%% file: floaters/figures/qualitative_cdavis.tex
\begin{figure*}[htbp]
  \centering
  \renewcommand{\arraystretch}{0.75}
  \setlength{\tabcolsep}{1pt}

\begin{tabular}[t]{@{}m{0.015\textwidth} m{0.192\textwidth} m{0.192\textwidth} m{0.192\textwidth} m{0.192\textwidth} m{0.192\textwidth} @{}}
  & 
  \makebox[0.192\textwidth]{\small Train View} &
  \makebox[0.192\textwidth]{\small E$^2$NeRF~\cite{qi2023e2nerf}} &
  \makebox[0.192\textwidth]{\small EvDeblurNeRF~\cite{cannici2024mitigating}} &
  \makebox[0.192\textwidth]{\small \textbf{Ours}}&
  \makebox[0.192\textwidth]{\small Reference View}\\

  \raisebox{0.3cm}{\rotatebox[origin=t]{90}{\small Drones}} &
  \includegraphics[width=\linewidth]{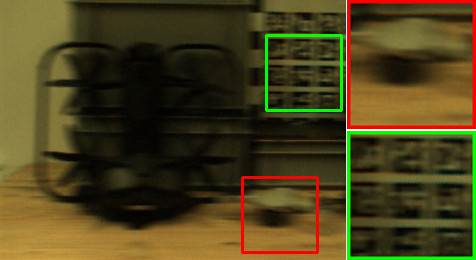} &
  \includegraphics[width=\linewidth]{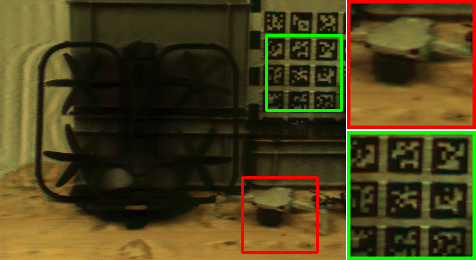}  &
    \includegraphics[width=\linewidth]{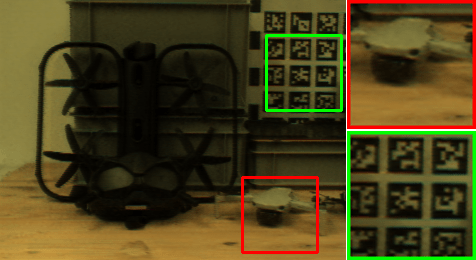}  &
  \includegraphics[width=\linewidth]{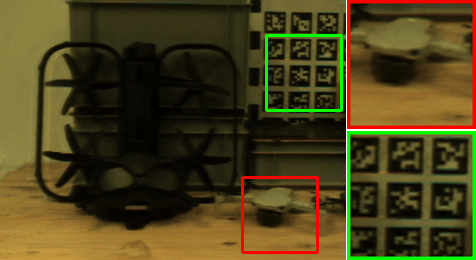} &
  \includegraphics[width=\linewidth]{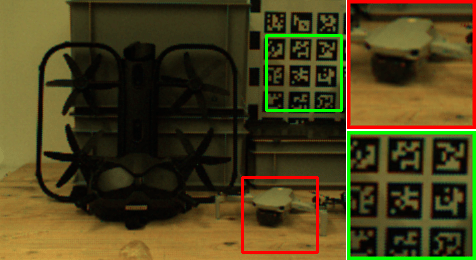}\\

  \raisebox{0.3cm}{\rotatebox[origin=t]{90}{\small Power Supplies}} &
  \includegraphics[width=\linewidth]{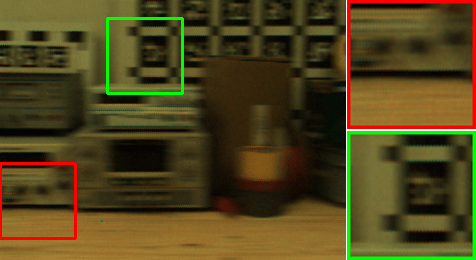} &
  \includegraphics[width=\linewidth]{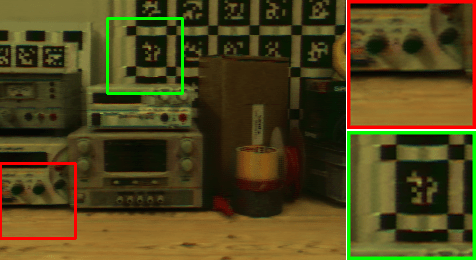}  &
    \includegraphics[width=\linewidth]{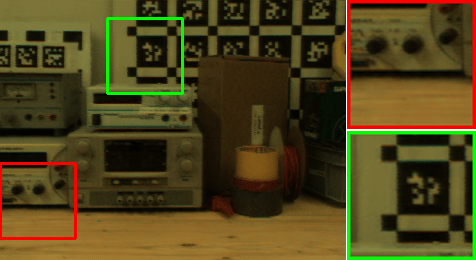}  &
  \includegraphics[width=\linewidth]{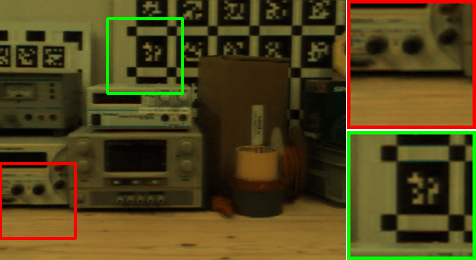} &
  \includegraphics[width=\linewidth]{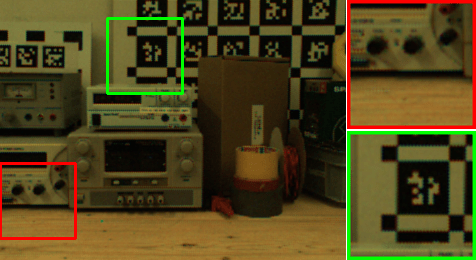}\\

  \raisebox{0.3cm}{\rotatebox[origin=t]{90}{\small Lab Equip}} &
  \includegraphics[width=\linewidth]{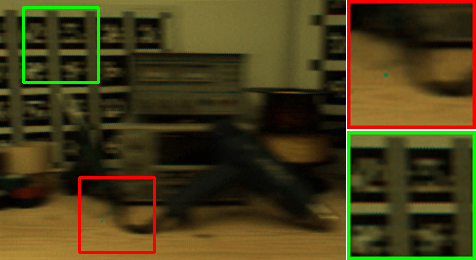} &
  \includegraphics[width=\linewidth]{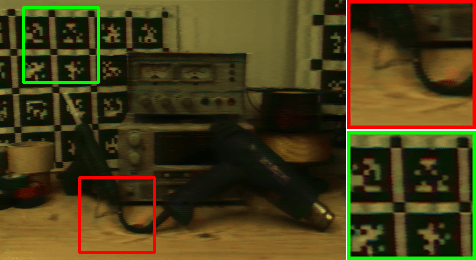}  &
    \includegraphics[width=\linewidth]{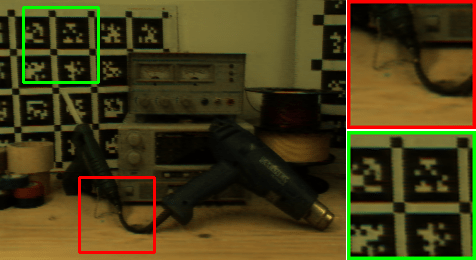}  &
  \includegraphics[width=\linewidth]{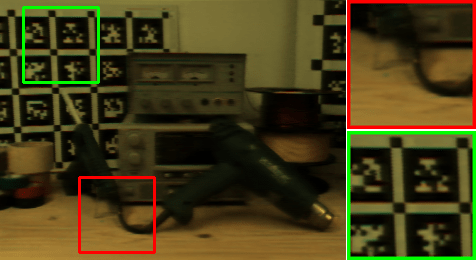} &
  \includegraphics[width=\linewidth]{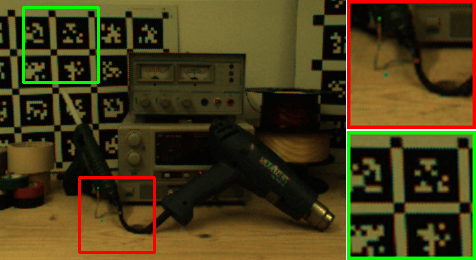}\\

  \end{tabular}

  \caption{Novel view synthesis comparison on the Ev-DeblurCDAVIS dataset.}
  \label{fig:cdavis}
\end{figure*}

%% file: floaters/tables/bslarge_handheld.tex
\begin{table*}[]
\caption{Quantitative comparison on the new Gen3-HandHeld dataset, and ablation of speed vs. exposure
time. USLAM stands for UltimateSLAM~\cite{vidal2018ultimate}.}
\label{tab:bslarge_handheld}
\renewcommand{\arraystretch}{1.15} 
\setlength{\tabcolsep}{4pt}        
\resizebox{\textwidth}{!}{
\begin{tabular}{ccccccccccccccc}
\toprule
\multirow{2}{*}{Exposure} & \multirow{2}{*}{Poses} & \multirow{2}{*}{Method} & \multicolumn{3}{c}{ Max Speed 0.8$\sim$1 m/s} & \multicolumn{3}{c}{ Max Speed 1.2$\sim$1.4 m/s} & \multicolumn{3}{c}{ Max Speed 1.4$\sim$1.8 m/s} & \multicolumn{3}{c}{Average} \\
\cmidrule(lr){4-6} \cmidrule(lr){7-9} \cmidrule(lr){10-12} \cmidrule(l){13-15}
 & & & PSNR$\uparrow$ & LPIPS$\downarrow$ & SSIM$\uparrow$ & PSNR$\uparrow$ & LPIPS$\downarrow$ & SSIM$\uparrow$ & PSNR$\uparrow$ & LPIPS$\downarrow$ & SSIM$\uparrow$ & PSNR$\uparrow$ & LPIPS$\downarrow$ & SSIM$\uparrow$ \\
\midrule
\multirow{4}{*}{10ms}
  & USLAM & Ev-DeblurNeRF~\cite{cannici2024mitigating} 
  & 16.92 & 0.57 & 0.45 & 16.30 & 0.59 & 0.42 & 15.99 & 0.64 & 0.42 & 16.40 & 0.60 & 0.43 \\

  & USLAM & Ours         
    & \textbf{28.14} & \textbf{0.21} & \textbf{0.77} & \textbf{27.34} & \textbf{0.23} & \textbf{0.76} & \textbf{26.69} & \textbf{0.27} & \textbf{0.74} & \textbf{27.39} & \textbf{0.24} & \textbf{0.76} \\

  \cdashline{2-15}\addlinespace[2pt]

  & \textcolor{gray}{our refined} & \textcolor{gray}{Ev-DeblurNeRF~\cite{cannici2024mitigating}} 
& \textcolor{gray}{24.03} & \textcolor{gray}{0.24} & \textcolor{gray}{0.72} & \textcolor{gray}{24.28} & \textcolor{gray}{0.26} & \textcolor{gray}{0.72} & \textcolor{gray}{23.17} & \textcolor{gray}{0.32} & \textcolor{gray}{0.69} & \textcolor{gray}{23.83} & \textcolor{gray}{0.27} & \textcolor{gray}{0.71} \\

  & \textcolor{gray}{our refined} & \textcolor{gray}{E$^2$NeRF~\cite{qi2023e2nerf}} 
& \textcolor{gray}{23.77} 
& \textcolor{gray}{0.24} 
& \textcolor{gray}{0.71}

& \textcolor{gray}{23.47} 
& \textcolor{gray}{0.30} 
& \textcolor{gray}{0.68} 

& \textcolor{gray}{21.46} 
& \textcolor{gray}{0.37} 
& \textcolor{gray}{0.62} 

& \textcolor{gray}{22.90} 
& \textcolor{gray}{0.30} 
& \textcolor{gray}{0.67} \\

\midrule
\multirow{4}{*}{30ms} 
  & USLAM & Ev-DeblurNeRF~\cite{cannici2024mitigating} 
   & 16.49 & 0.46 & 0.43 & 16.33 & 0.51 & 0.45 & 15.69 & 0.55 & 0.43 & 16.17 & 0.51 & 0.44 \\

  & USLAM & Ours         
  & \textbf{24.57} & \textbf{0.14} & \textbf{0.77} & \textbf{24.73} & \textbf{0.18} & \textbf{0.77} & \textbf{23.27} & \textbf{0.21} & \textbf{0.73} & \textbf{24.19} & \textbf{0.18} & \textbf{0.76} \\

  \cdashline{2-15}\addlinespace[2pt]
  
  & \textcolor{gray}{our refined} & \textcolor{gray}{Ev-DeblurNeRF~\cite{cannici2024mitigating}} 
& \textcolor{gray}{24.04} & \textcolor{gray}{0.15} & \textcolor{gray}{0.77} & \textcolor{gray}{23.79} & \textcolor{gray}{0.19} & \textcolor{gray}{0.76} & \textcolor{gray}{22.57} & \textcolor{gray}{0.25} & \textcolor{gray}{0.71} & \textcolor{gray}{23.47} & \textcolor{gray}{0.20} & \textcolor{gray}{0.74} \\

& \textcolor{gray}{our refined} & \textcolor{gray}{E$^2$NeRF~\cite{qi2023e2nerf}} 
& \textcolor{gray}{22.99} 
& \textcolor{gray}{0.19} 
& \textcolor{gray}{0.72} 

& \textcolor{gray}{23.09} 
& \textcolor{gray}{0.33} 
& \textcolor{gray}{0.61} 

& \textcolor{gray}{21.22} 
& \textcolor{gray}{0.35} 
& \textcolor{gray}{0.63} 

& \textcolor{gray}{22.43} 
& \textcolor{gray}{0.29} 
& \textcolor{gray}{0.65} \\

\midrule
\multirow{4}{*}{50ms} 
  & USLAM & Ev-DeblurNeRF~\cite{cannici2024mitigating} 
  & 14.34 & 0.53 & 0.36 & 14.00 & 0.60 & 0.37 & 13.16 & 0.63 & 0.33 & 13.84 & 0.59 & 0.35 \\
  
  & USLAM & Ours  
  & \textbf{23.80} & \textbf{0.13} & \textbf{0.78} & \textbf{22.52} & \textbf{0.18} & \textbf{0.74} & \textbf{21.31} & \textbf{0.26} & \textbf{0.68} & \textbf{22.54} & \textbf{0.19} & \textbf{0.73} \\

  \cdashline{2-15}\addlinespace[2pt]
  
  & \textcolor{gray}{our refined} & \textcolor{gray}{Ev-DeblurNeRF~\cite{cannici2024mitigating}} 
& \textcolor{gray}{22.96} & \textcolor{gray}{0.14} & \textcolor{gray}{0.76} & \textcolor{gray}{21.50} & \textcolor{gray}{0.21} & \textcolor{gray}{0.71} & \textcolor{gray}{18.28} & \textcolor{gray}{0.35} & \textcolor{gray}{0.58} & \textcolor{gray}{20.91} & \textcolor{gray}{0.23} & \textcolor{gray}{0.68} \\

  & \textcolor{gray}{our refined} & \textcolor{gray}{E$^2$NeRF~\cite{qi2023e2nerf}} 
& \textcolor{gray}{21.95} 
& \textcolor{gray}{0.22} 
& \textcolor{gray}{0.69}

& \textcolor{gray}{19.12} 
& \textcolor{gray}{0.32} 
& \textcolor{gray}{0.60} 

& \textcolor{gray}{20.26}
& \textcolor{gray}{0.40}
& \textcolor{gray}{0.62}

& \textcolor{gray}{20.44}
& \textcolor{gray}{0.31}
& \textcolor{gray}{0.64} \\

\bottomrule
\end{tabular}}
\end{table*}

%% file: floaters/figures/qualitative_noisy_blender.tex
\begin{figure*}
  \centering
  \renewcommand{\arraystretch}{0.75}
  \setlength{\tabcolsep}{1pt}

\begin{tabular}[t]{@{}m{0.015\textwidth} m{0.192\textwidth} m{0.192\textwidth} m{0.192\textwidth} m{0.192\textwidth} m{0.192\textwidth} @{}}
  & 
  \makebox[0.192\textwidth]{\small Train View} &
  \makebox[0.192\textwidth]{\small EvDeblurNeRF~\cite{cannici2024mitigating}} &
  \makebox[0.192\textwidth]{\small {\small Ours (F)}} &
  \makebox[0.192\textwidth]{\small {\small \textbf{Ours (E+F)}}}&
  \makebox[0.192\textwidth]{\small Reference View}\\

  \raisebox{0.3cm}{\rotatebox[origin=t]{90}{\small Factory}} &
  \includegraphics[width=\linewidth]{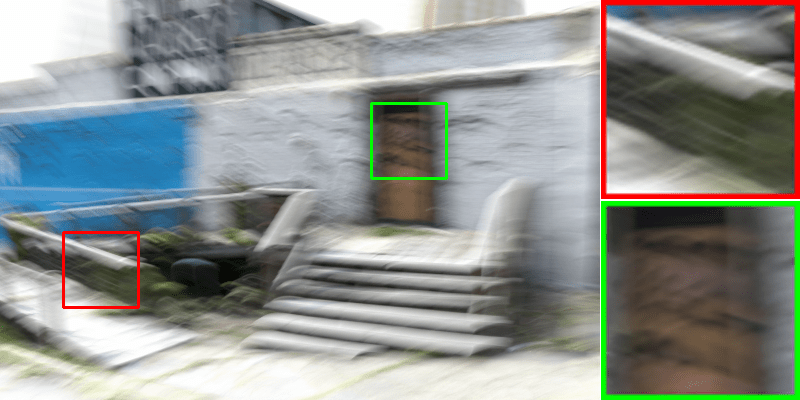} &
  \includegraphics[width=\linewidth]{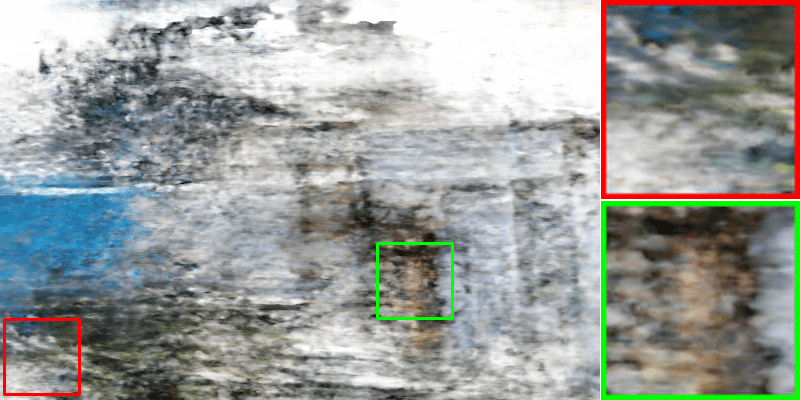}  &
  \includegraphics[width=\linewidth]{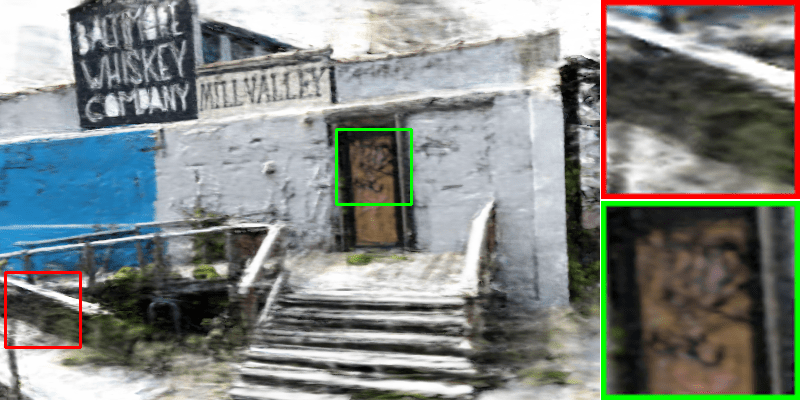}  &
  \includegraphics[width=\linewidth]{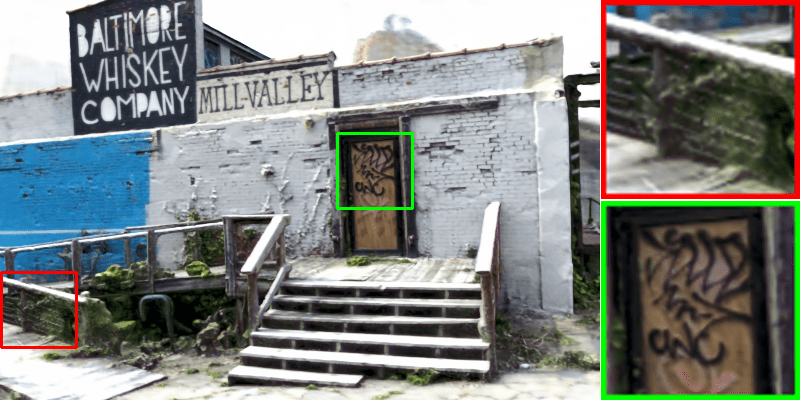} &
  \includegraphics[width=\linewidth]{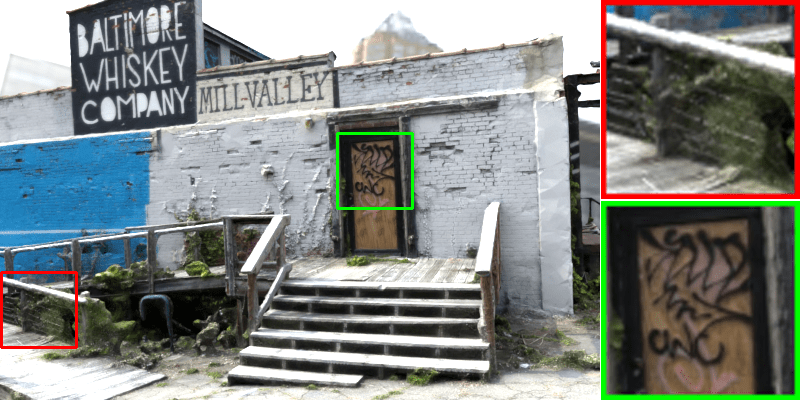}\\

  \raisebox{0.3cm}{\rotatebox[origin=t]{90}{\small Trolley}} &
  \includegraphics[width=\linewidth]{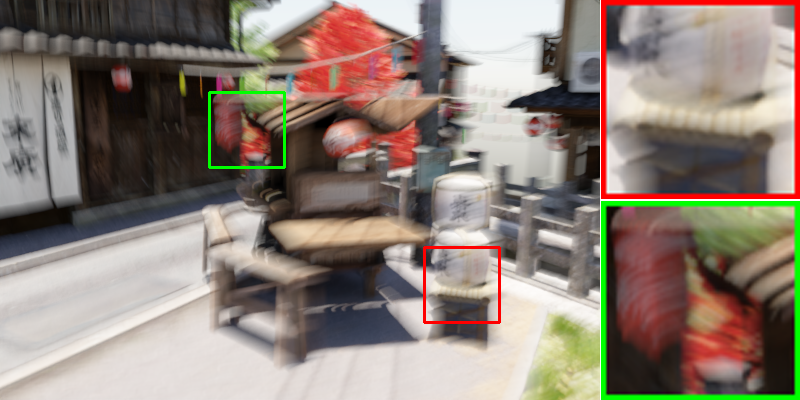} &
  \includegraphics[width=\linewidth]{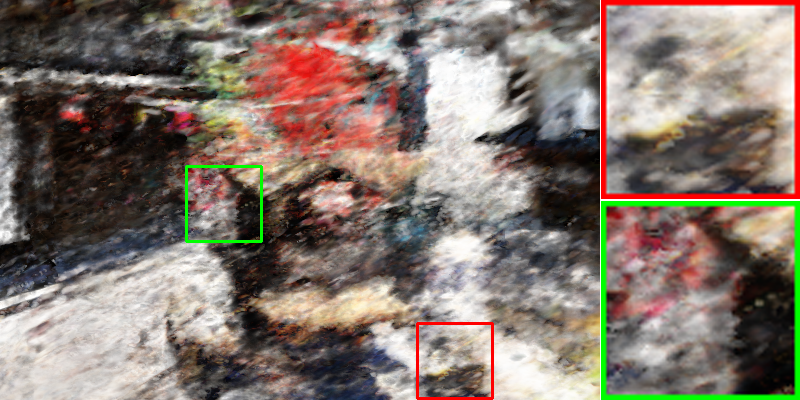}  &
  \includegraphics[width=\linewidth]{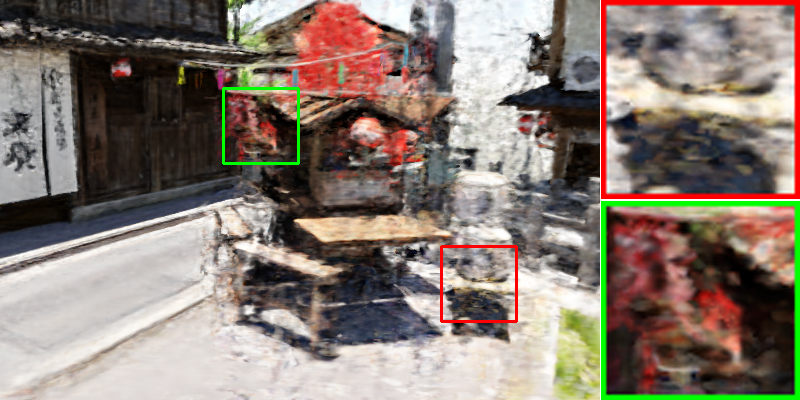}  &
  \includegraphics[width=\linewidth]{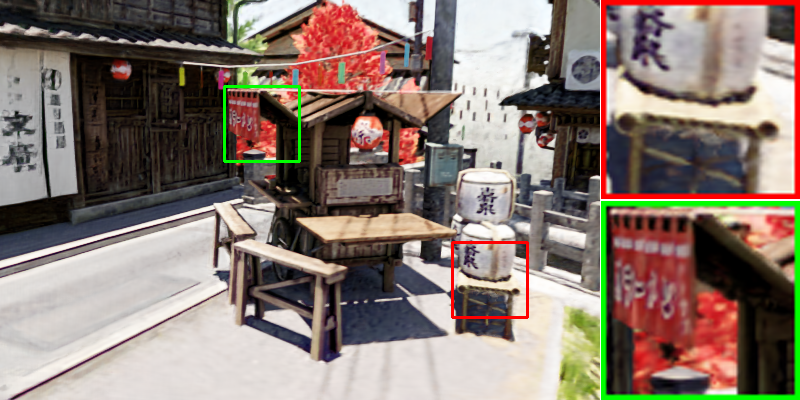}&
  \includegraphics[width=\linewidth]{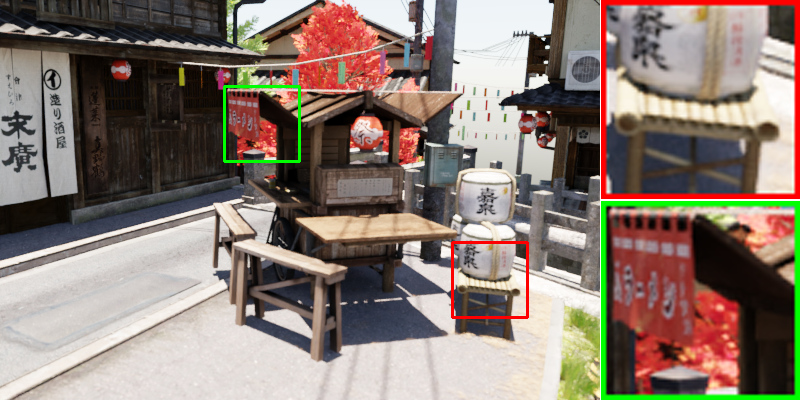}  \\

  \raisebox{0.3cm}{\rotatebox[origin=t]{90}{\small Pool}} &
  \includegraphics[width=\linewidth]{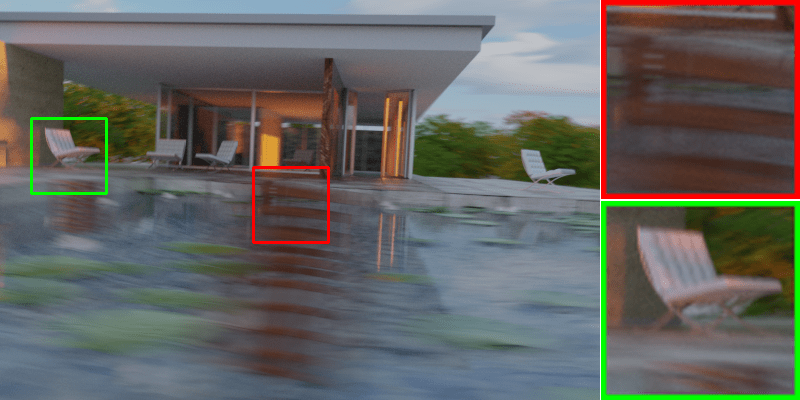} &
  \includegraphics[width=\linewidth]{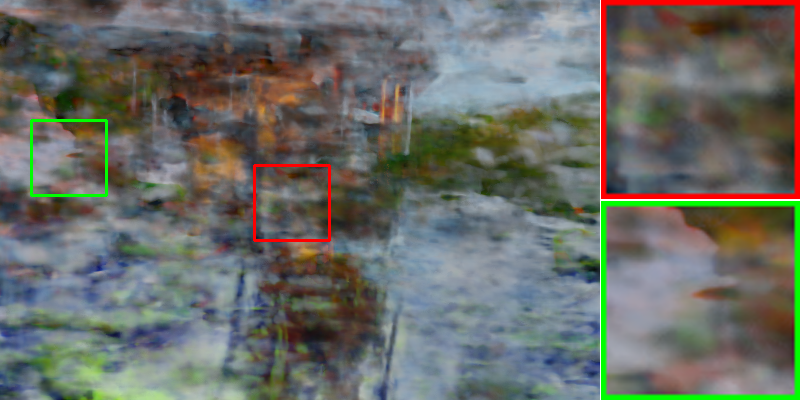}  &
  \includegraphics[width=\linewidth]{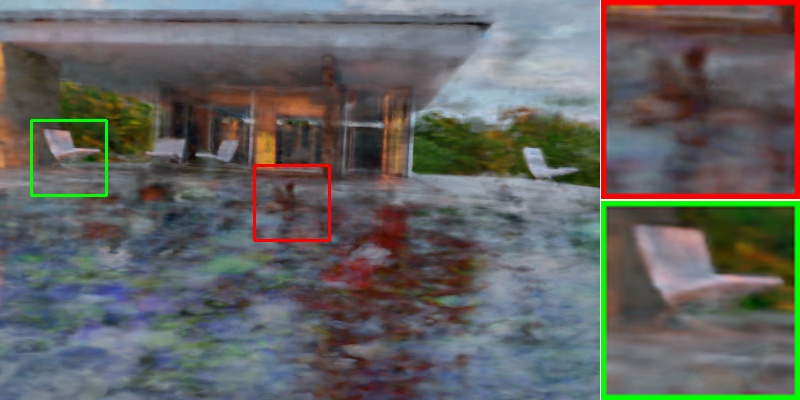}  &
  \includegraphics[width=\linewidth]{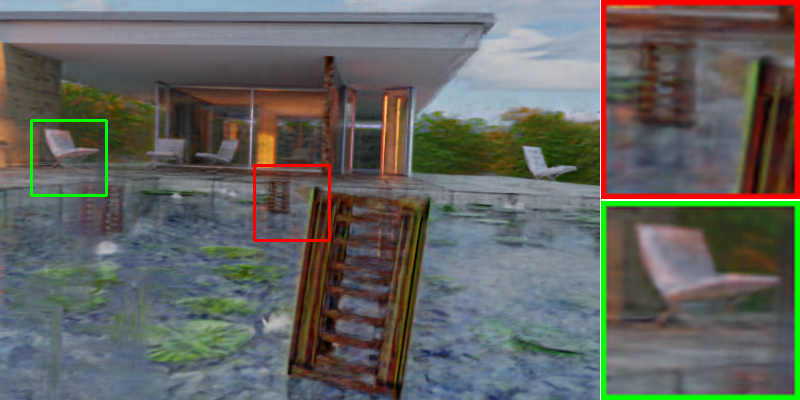} &
  \includegraphics[width=\linewidth]{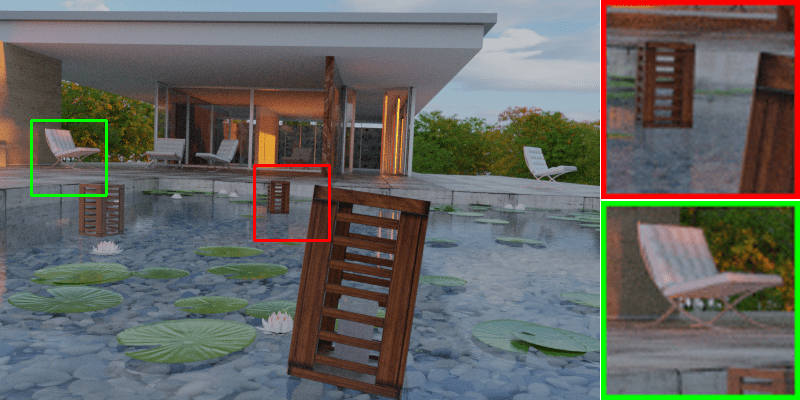} \\

  \raisebox{0.3cm}{\rotatebox[origin=t]{90}{\small Tanabata}} &
  \includegraphics[width=\linewidth]{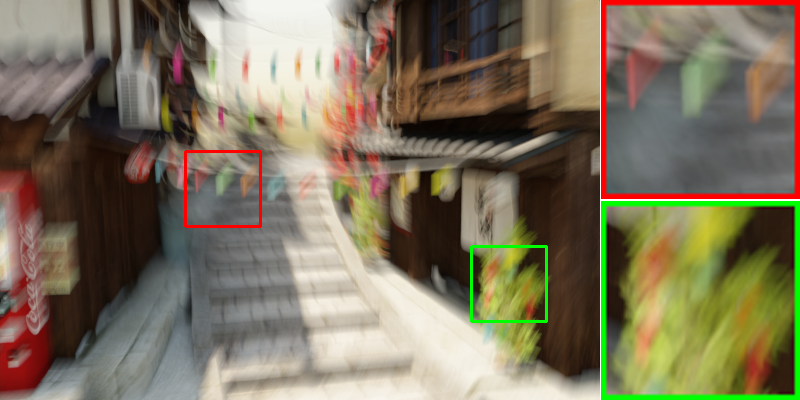} &
  \includegraphics[width=\linewidth]{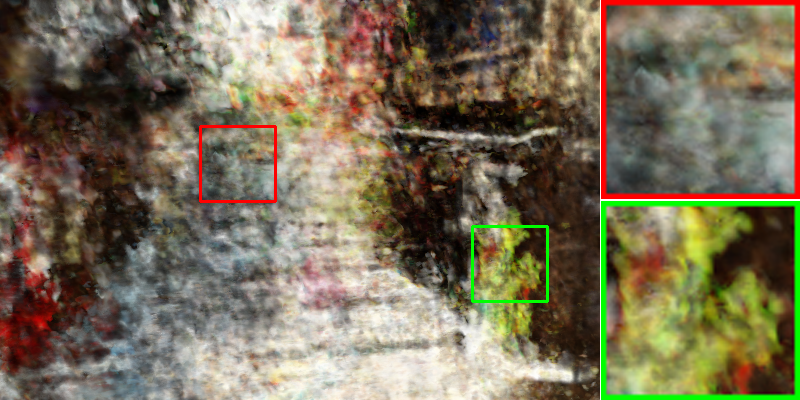}  &
  \includegraphics[width=\linewidth]{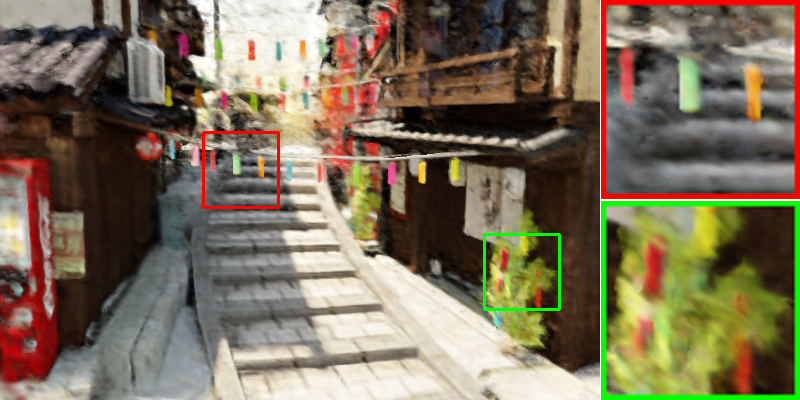}  &
  \includegraphics[width=\linewidth]{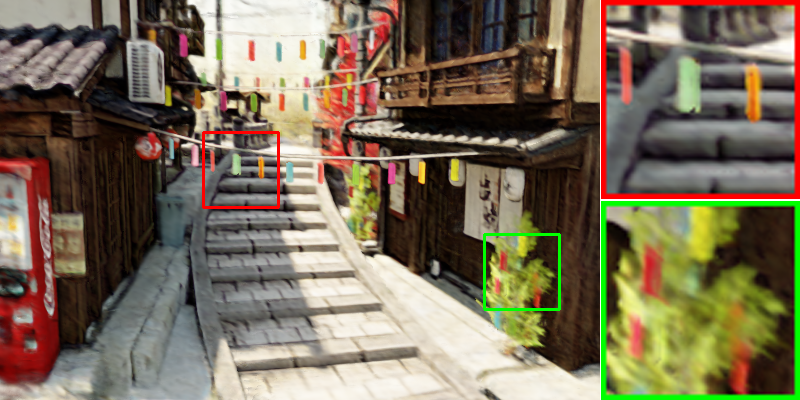} &
  \includegraphics[width=\linewidth]{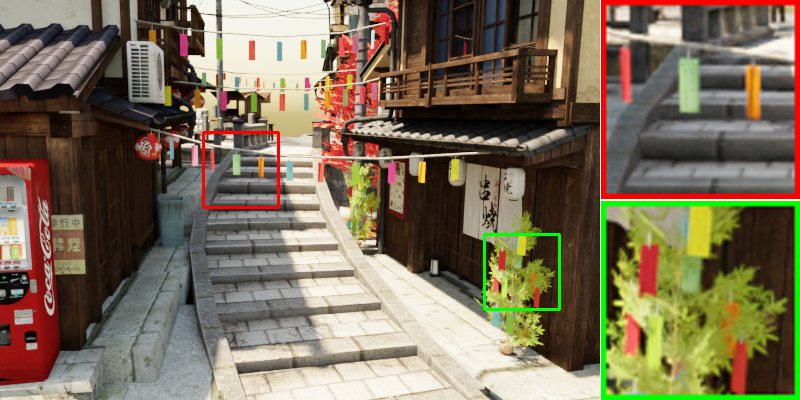}  \\

  \end{tabular}

  \caption{Novel view synthesis comparison on the NoisyPose-EvDeblurBlender dataset under noise level four. We indicate \textit{frame-only} methods with (F) and \textit{frames+events} with (F+E), while \textit{Train view} refers to the closest blurry image in the training set.}
  \label{fig:noisy_poses}
\end{figure*}

%% file: floaters/figures/qualitative_bsgen3_handheld.tex
\begin{figure*}
  \centering
  \renewcommand{\arraystretch}{0.75}
  \setlength{\tabcolsep}{1pt}

\begin{tabular}[t]{@{}m{0.015\textwidth} m{0.192\textwidth} m{0.192\textwidth} m{0.192\textwidth} m{0.192\textwidth} m{0.192\textwidth} @{}}
  & 
  \makebox[0.192\textwidth]{\small Train View} &
  \makebox[0.192\textwidth]{\small E$^2$NeRF~\cite{qi2023e2nerf} ($T_\text{ref}$)} &
  \makebox[0.192\textwidth]{\small EvDeblurNeRF~\cite{cannici2024mitigating} ($T_\text{ref}$)} &
  \makebox[0.192\textwidth]{\small \textbf{Ours} ($T_\text{USLAM}$)} &
  \makebox[0.192\textwidth]{\small Reference View} \\
  
  \raisebox{0.3cm}{\rotatebox[origin=t]{90}{\small 10\,ms}} &
  \includegraphics[width=\linewidth]{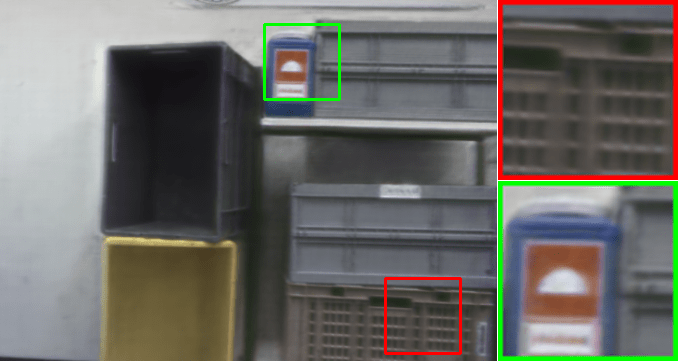} &
  \includegraphics[width=\linewidth]{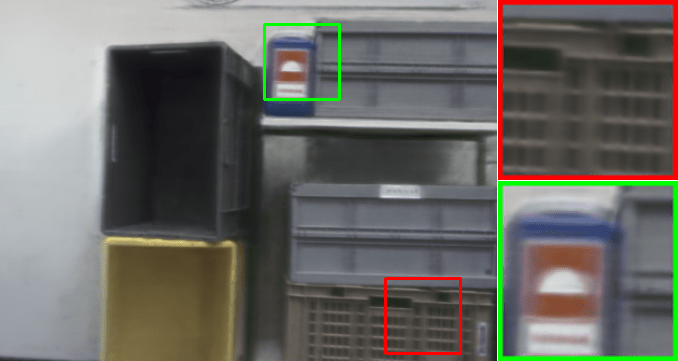} &
  \includegraphics[width=\linewidth]{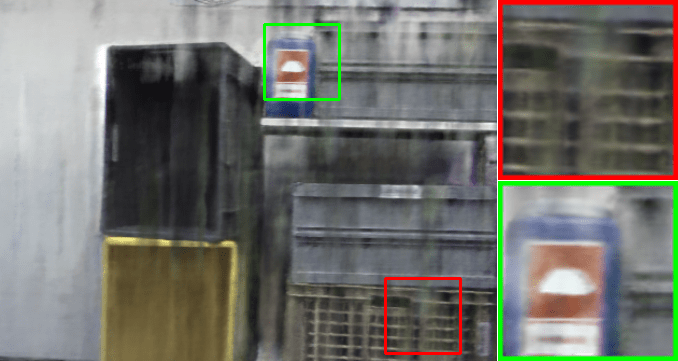}  &
  \includegraphics[width=\linewidth]{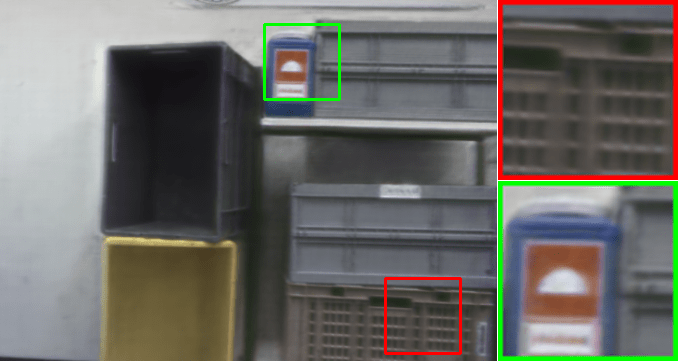} &
  \includegraphics[width=\linewidth]{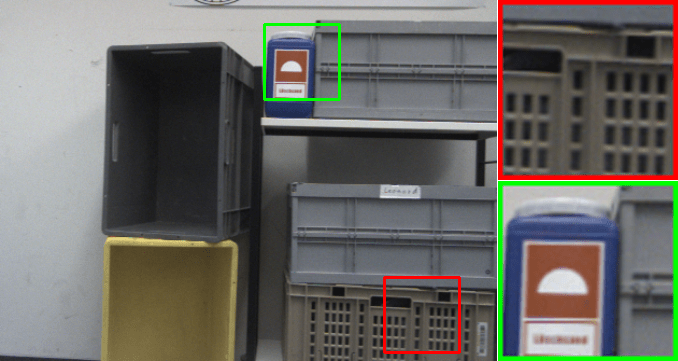} \\
  
  \raisebox{0.3cm}{\rotatebox[origin=t]{90}{\small 30\,ms}} &
  \includegraphics[width=\linewidth]{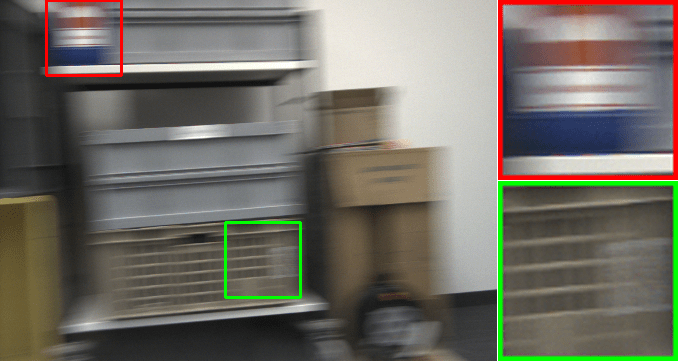} &
  \includegraphics[width=\linewidth]{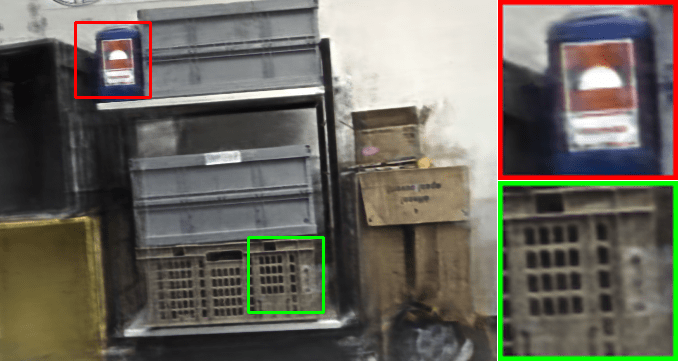} &
  \includegraphics[width=\linewidth]{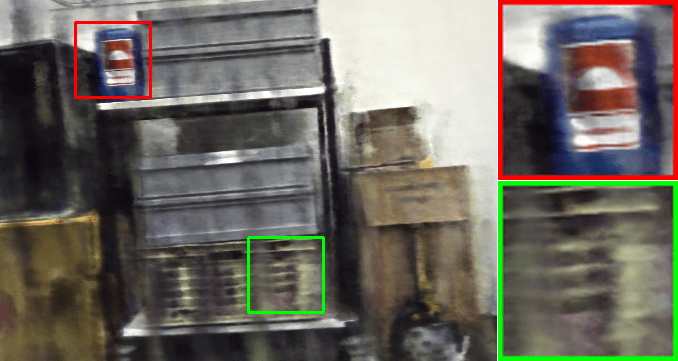}  & 
  \includegraphics[width=\linewidth]{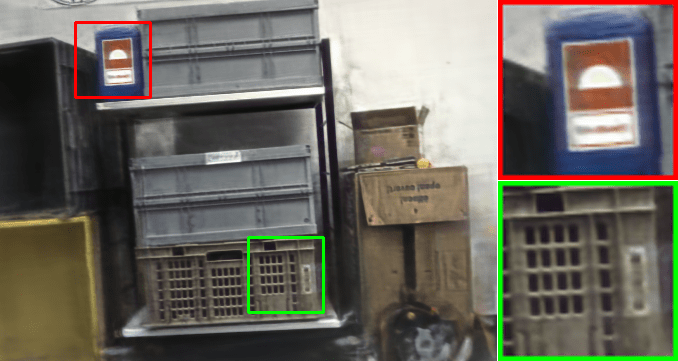} &
  \includegraphics[width=\linewidth]{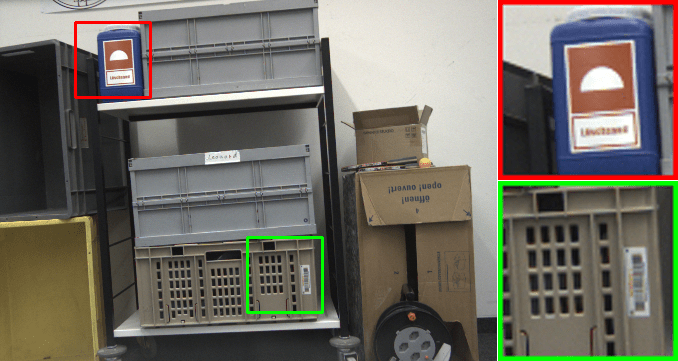} \\

  \raisebox{0.2cm}{\rotatebox[origin=t]{90}{\small 50\,ms}} &
  \includegraphics[width=\linewidth]{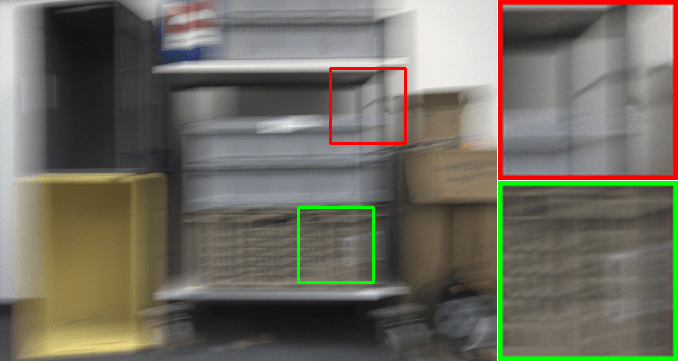} &
  \includegraphics[width=\linewidth]{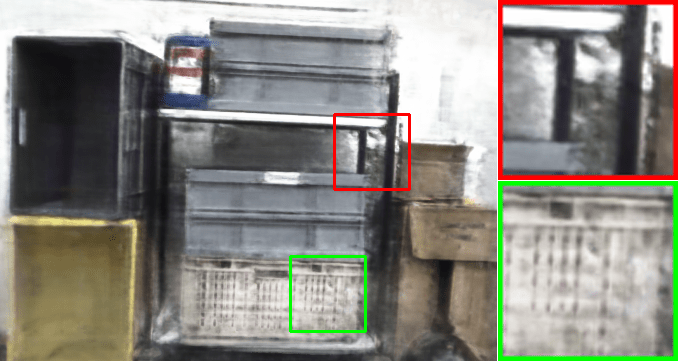} &
  \includegraphics[width=\linewidth]{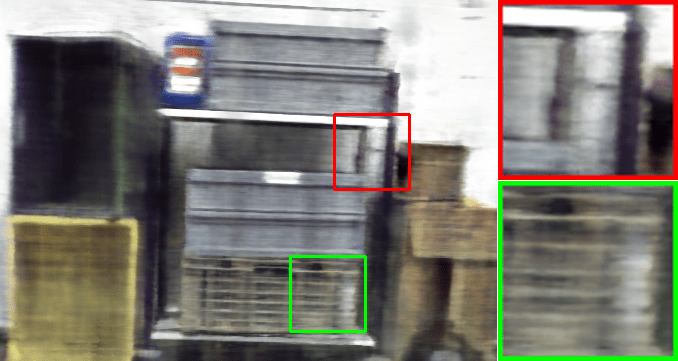} &  
  \includegraphics[width=\linewidth]{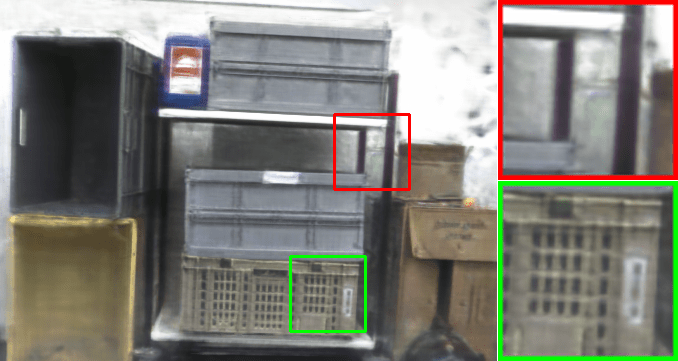} &
  \includegraphics[width=\linewidth]{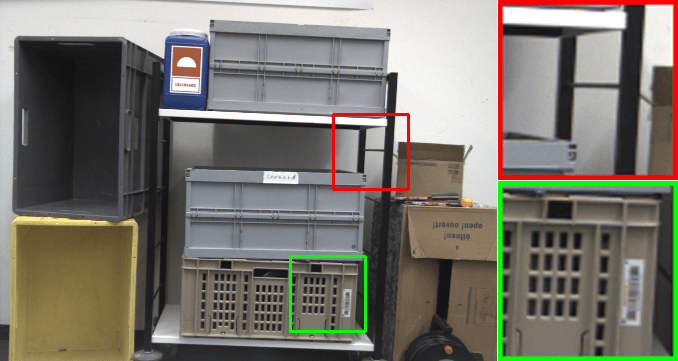} \\

  \end{tabular}

    \caption{Novel view synthesis comparison on the Gen3-HandHeld dataset under the fastest speed profile. We indicate in parentheses the poses each method uses as initialization, $T_\text{ref}$ for our refined poses, $T_\text{USLAM}$ for UltimateSLAM~\cite{vidal2018ultimate} poses.}
  \label{fig:bsgen3_handheld}
\end{figure*}

%% file: floaters/figures/qualitative_bsgen3_drone_flight.tex
\begin{figure*}
  \centering
  \renewcommand{\arraystretch}{0.75}
  \setlength{\tabcolsep}{1pt}

\begin{tabular}[t]{@{}m{0.015\textwidth} m{0.16\textwidth} m{0.16\textwidth} m{0.16\textwidth} m{0.16\textwidth} m{0.16\textwidth} m{0.16\textwidth} @{}}
  
  &
  \makebox[0.16\textwidth]{\small Train View} &
  \makebox[0.16\textwidth]{\small E$^2$NeRF~\cite{qi2023e2nerf} ($T_\text{ref}$)} &
  \makebox[0.16\textwidth]{\small EvDeblurNeRF~\cite{cannici2024mitigating} ($T_\text{ref}$)} &
  \makebox[0.16\textwidth]{\small \textbf{Ours} ($T_\text{USLAM}$)} &
  \makebox[0.16\textwidth]{\small \textbf{Ours} ($T_\text{ref}$)} &
  \makebox[0.16\textwidth]{\small Reference View} \\
  
  \raisebox{0.3cm}{\rotatebox[origin=t]{90}{\small \textsc{Box.}}} &
  \includegraphics[width=\linewidth]{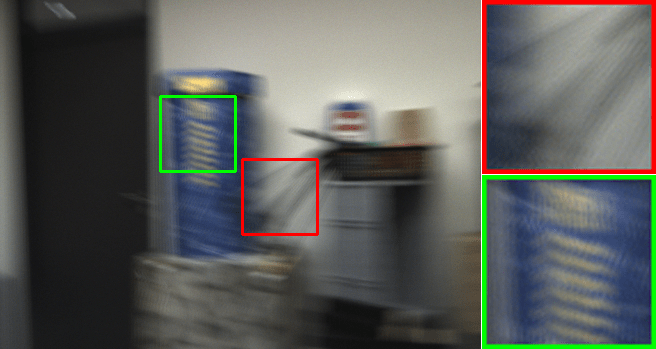} &
  \includegraphics[width=\linewidth]{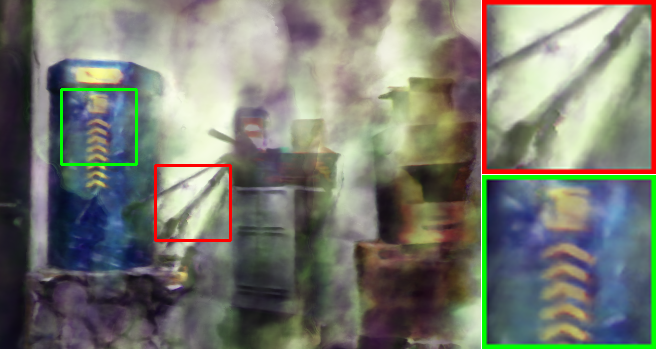} &
  \includegraphics[width=\linewidth]{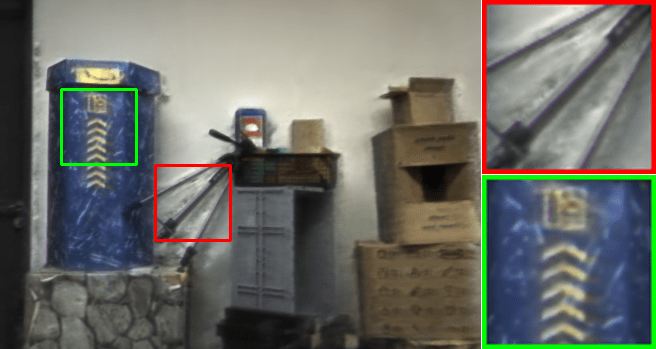} &
  \includegraphics[width=\linewidth]{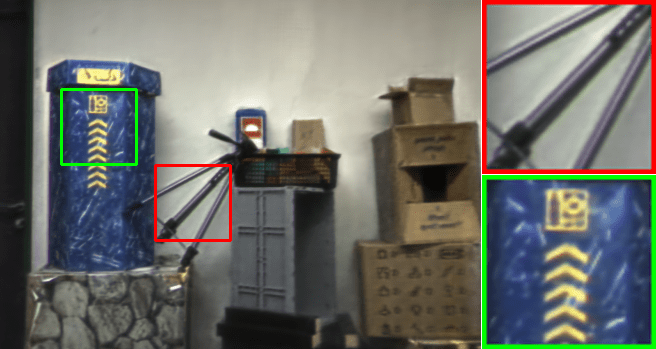} &
  \includegraphics[width=\linewidth]{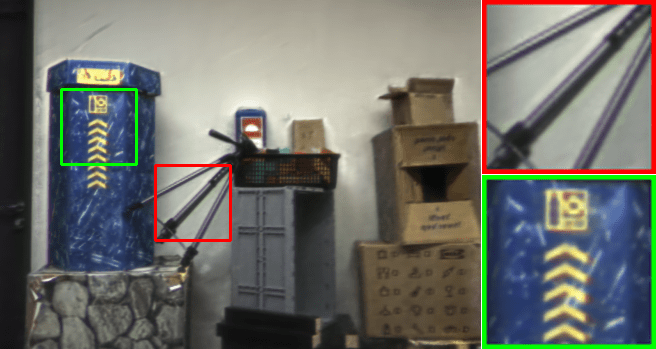} &
  \includegraphics[width=\linewidth]{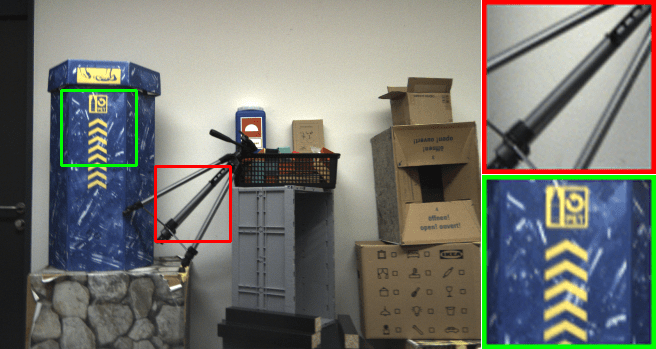} \\

  \raisebox{0.3cm}{\rotatebox[origin=t]{90}{\small \textsc{Equip.}}} &
  \includegraphics[width=\linewidth]{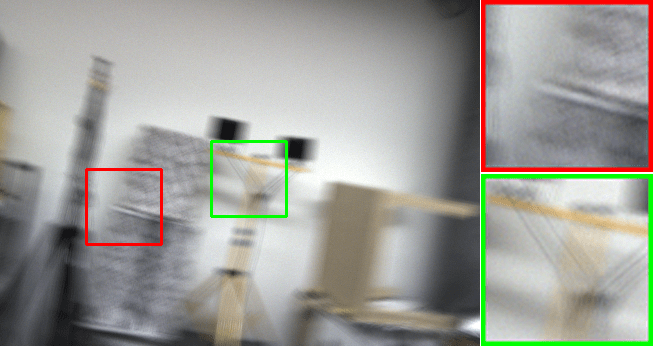} &
  \includegraphics[width=\linewidth]{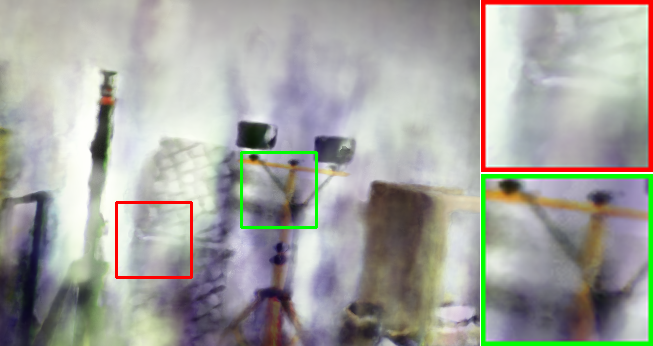} &
  \includegraphics[width=\linewidth]{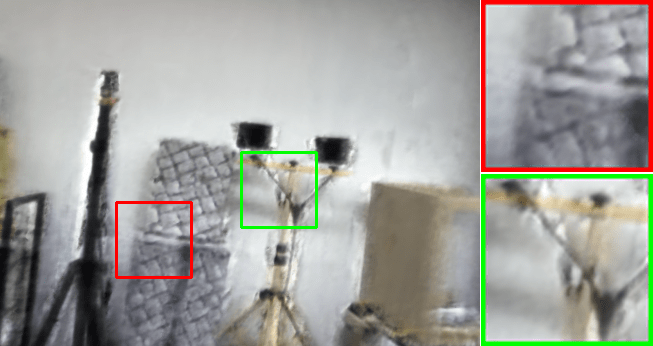} &
  \includegraphics[width=\linewidth]{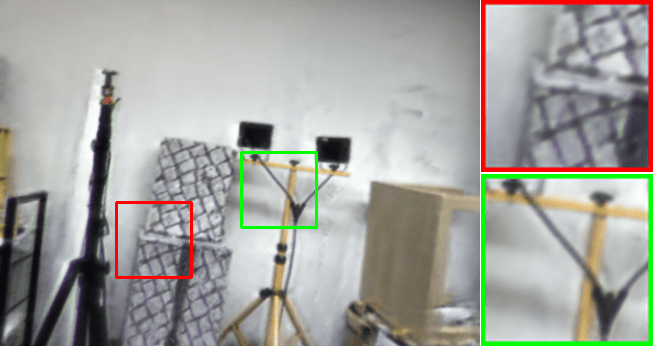} &
  \includegraphics[width=\linewidth]{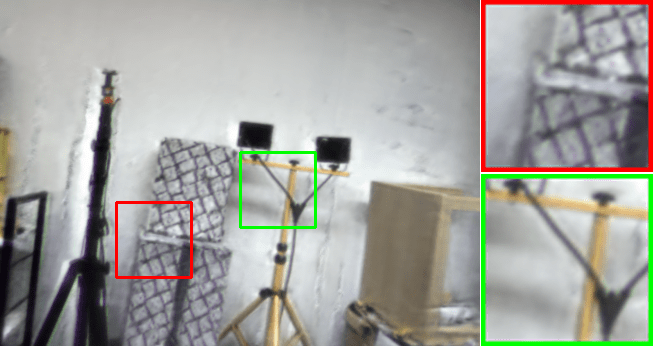} &
  \includegraphics[width=\linewidth]{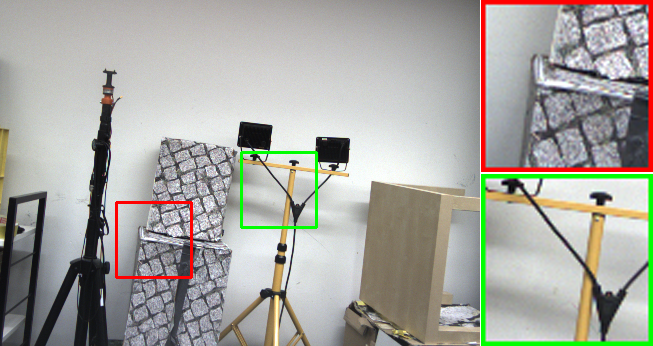} \\

  \raisebox{0.3cm}{\rotatebox[origin=t]{90}{\small \textsc{Play.}}} &

  \includegraphics[width=\linewidth]{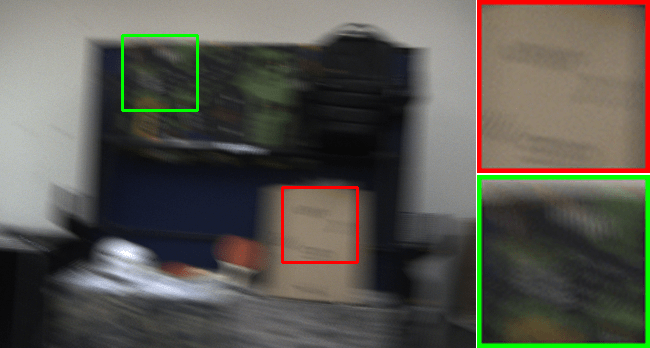} &
  \includegraphics[width=\linewidth]{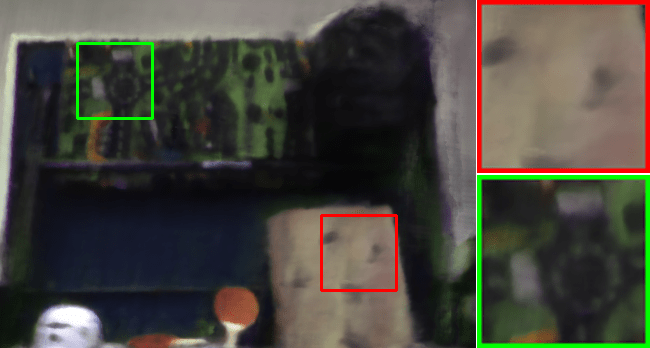} &
  \includegraphics[width=\linewidth]{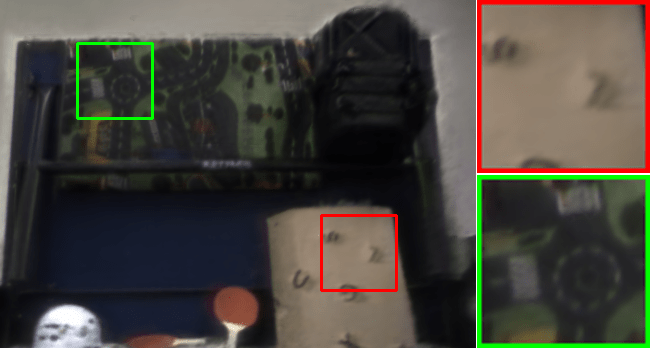} &
  \includegraphics[width=\linewidth]{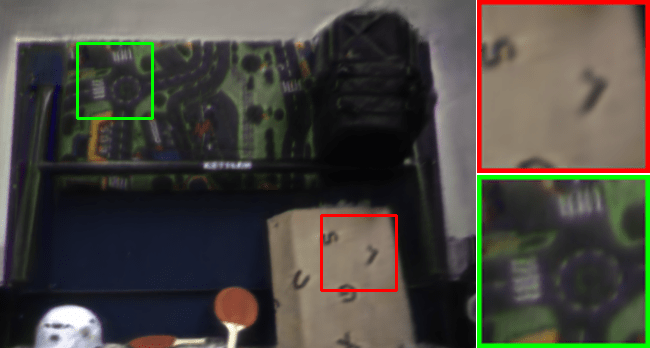} &
    \includegraphics[width=\linewidth]{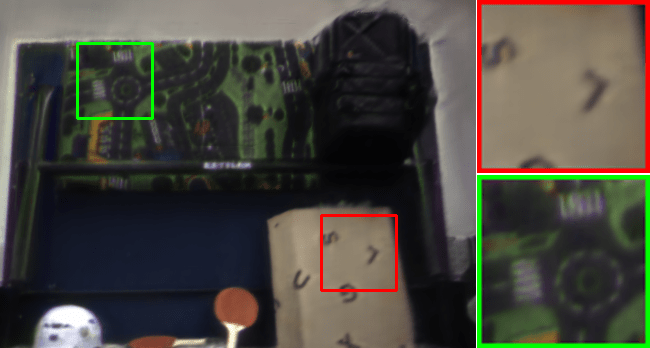} &
  \includegraphics[width=\linewidth]{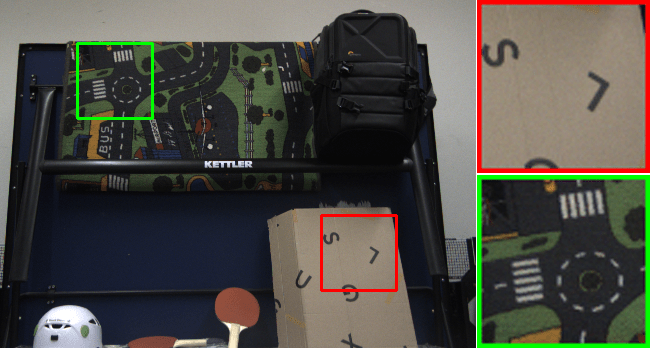} \\

    \raisebox{0.3cm}{\rotatebox[origin=t]{90}{\small \new{\textsc{Read.}}}} &

  \includegraphics[width=\linewidth]{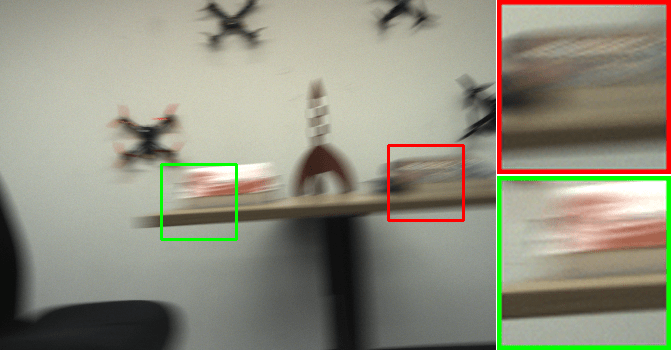} &
  \includegraphics[width=\linewidth]{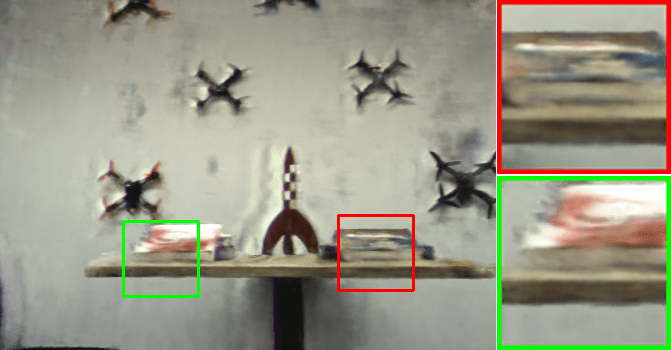} &
  \includegraphics[width=\linewidth]{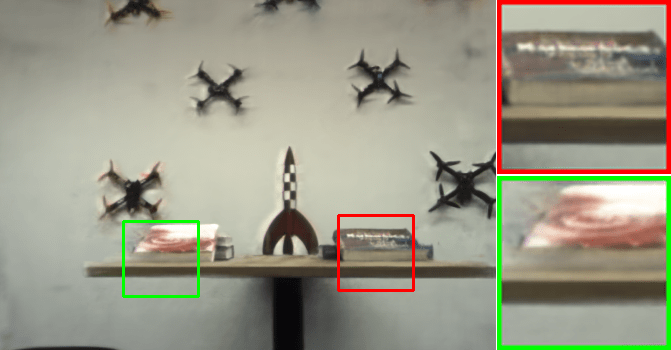} &
  \includegraphics[width=\linewidth]{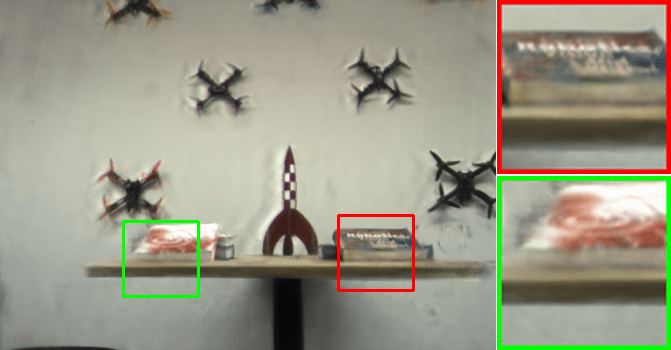} &
    \includegraphics[width=\linewidth]{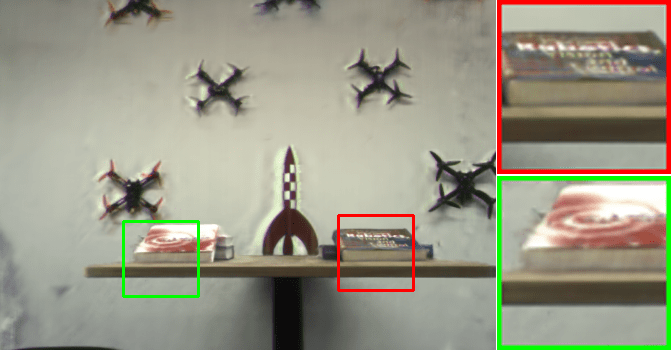} &
  \includegraphics[width=\linewidth]{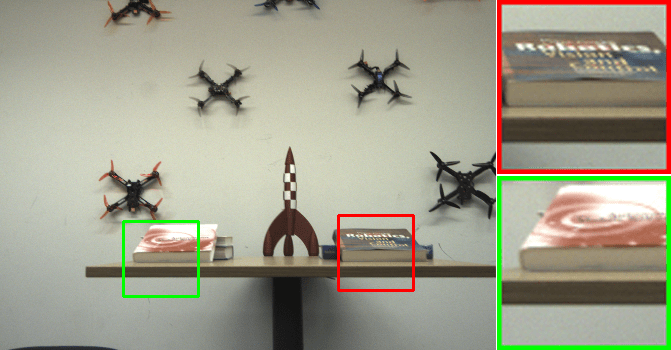} \\

    \raisebox{0.3cm}{\rotatebox[origin=t]{90}{\small \new{\textsc{Models}}}} &
  \includegraphics[width=\linewidth]{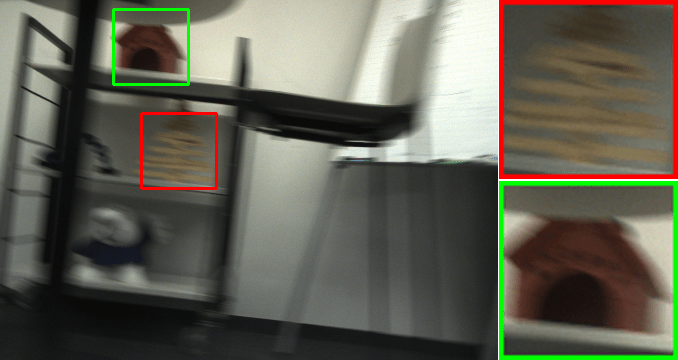} &
  \includegraphics[width=\linewidth]{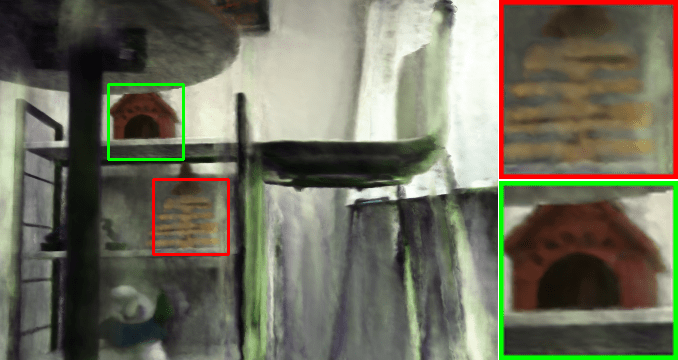} &
  \includegraphics[width=\linewidth]{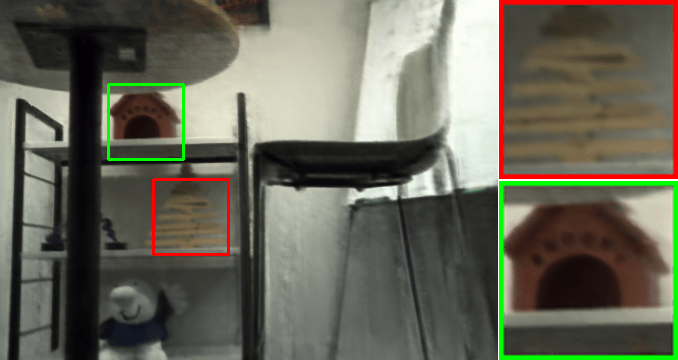} &
  \includegraphics[width=\linewidth]{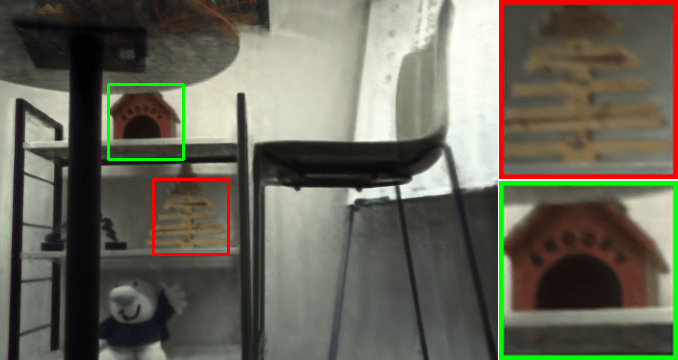} &
    \includegraphics[width=\linewidth]{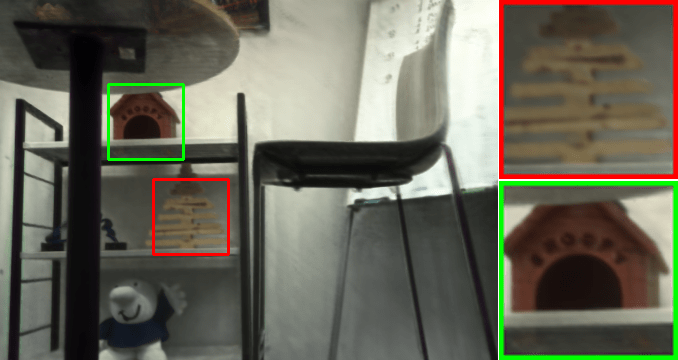} &
  \includegraphics[width=\linewidth]{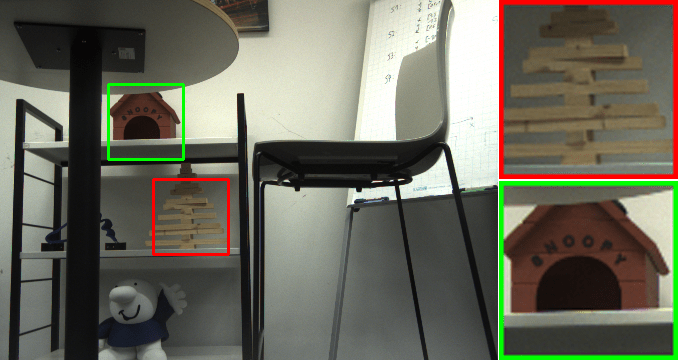} \\

  \raisebox{0.3cm}{\rotatebox[origin=t]{90}{\small \new{\textsc{Lounge}}}} & 
  \includegraphics[width=\linewidth]{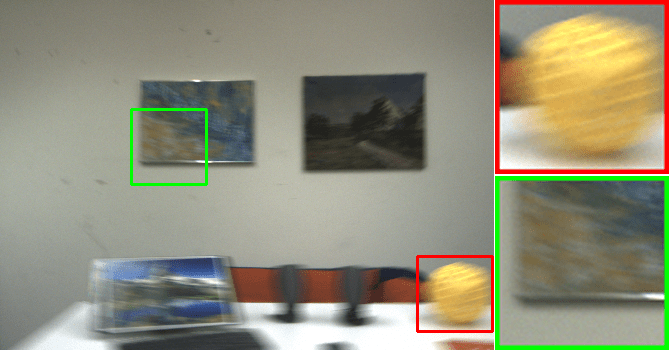} &
  \includegraphics[width=\linewidth]{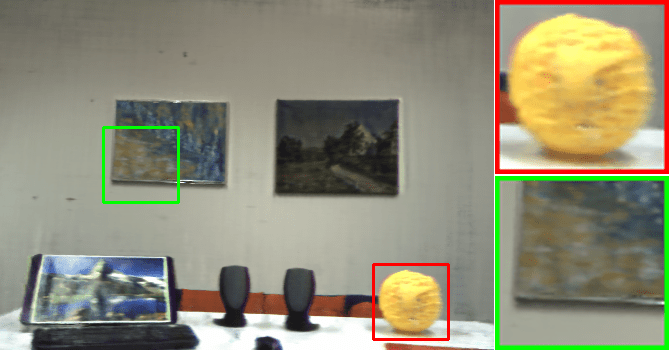} &
  \includegraphics[width=\linewidth]{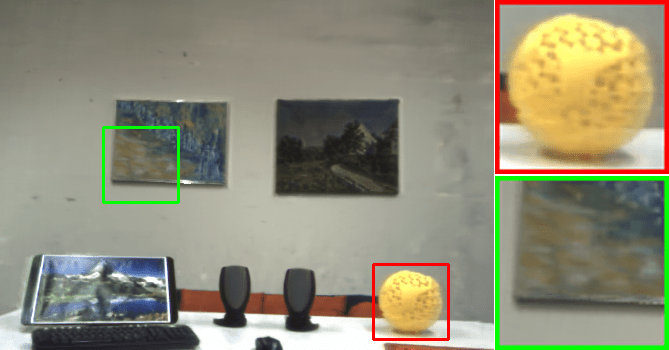} &
  \includegraphics[width=\linewidth]{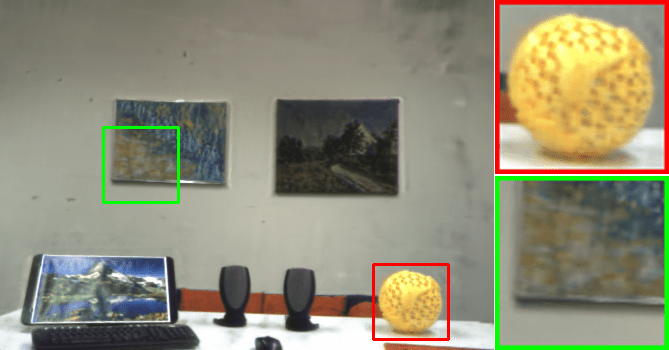} &
      \includegraphics[width=\linewidth]{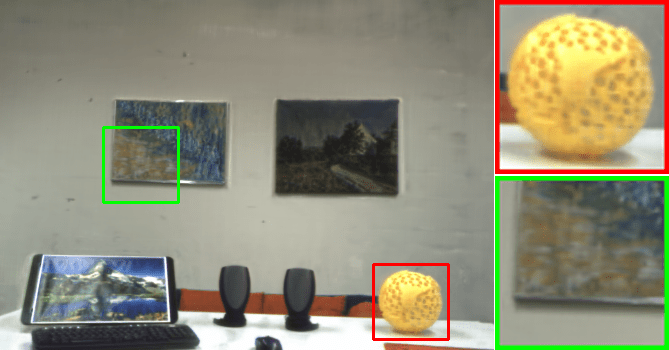} &
  \includegraphics[width=\linewidth]{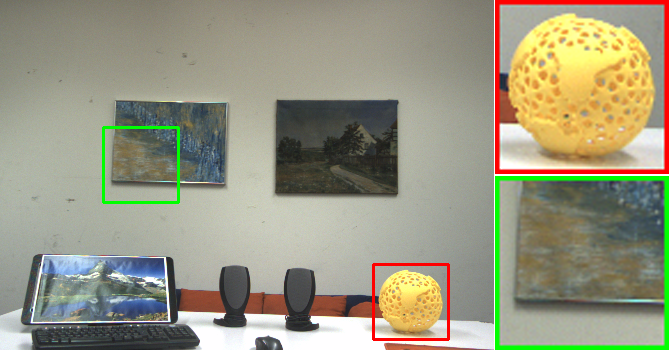} \\
  \end{tabular}

  \caption{\new{Novel view synthesis comparison on the Gen3-DroneFlight dataset under the fastest speed profile. We indicate in parentheses the poses each method uses as initialization, $T_\text{ref}$ for our refined poses, and $T_\text{USLAM}$ for UltimateSLAM~\cite{vidal2018ultimate} poses.}}
  \label{fig:bsgen3_drone}
\end{figure*}

%% file: floaters/tables/bslarge_drone.tex
\begin{table*}
\caption{Quantitative comparison on the new Gen3-DroneFlight dataset. USLAM stands for UltimateSLAM~\cite{vidal2018ultimate}.}
\label{tab:gen3_eval}
\renewcommand{\arraystretch}{1.15} 
\setlength{\tabcolsep}{4pt}        
\resizebox{\textwidth}{!}{
\begin{tabular}{ccccccccccccccc}
\toprule
\multirow{2}{*}{Sequence} & \multirow{2}{*}{Poses} & \multirow{2}{*}{Method} & \multicolumn{3}{c}{ Max Speed 1 m/s} & \multicolumn{3}{c}{Max Speed 1.5 m/s} & \multicolumn{3}{c}{ Max Speed 2 m/s} & \multicolumn{3}{c}{Average} \\
\cmidrule(lr){4-6} \cmidrule(lr){7-9} \cmidrule(lr){10-12} \cmidrule(l){13-15}
 & & & PSNR$\uparrow$ & LPIPS$\downarrow$ & SSIM$\uparrow$ & PSNR$\uparrow$ & LPIPS$\downarrow$ & SSIM$\uparrow$ & PSNR$\uparrow$ & LPIPS$\downarrow$ & SSIM$\uparrow$ & PSNR$\uparrow$ & LPIPS$\downarrow$ & SSIM$\uparrow$ \\
\midrule 
\multirow{5}{*}{\textsc{BoxStack}} 
  & USLAM & Ev-DeblurNeRF~\cite{cannici2024mitigating} & 
  13.34 & 0.58 & 0.30 & 12.08 & 0.64 & 0.25 & 13.58 & 0.66 & 0.30 & 13.00 & 0.63 & 0.28 \\

  & USLAM & Ours 
  & \textbf{23.89} & \textbf{0.21} & \textbf{0.68} & \textbf{23.49} & \textbf{0.26} & \textbf{0.67} & \textbf{23.05} & \textbf{0.33} & \textbf{0.66} & \textbf{23.48} & \textbf{0.26} & \textbf{0.67} \\

  \cdashline{2-15}\addlinespace[2pt]
  
  & \textcolor{gray}{our refined} & \textcolor{gray}{Ev-DeblurNeRF~\cite{cannici2024mitigating}} 
&\textcolor{gray}{21.83} & \textcolor{gray}{0.23} & \textcolor{gray}{0.61} & \textcolor{gray}{22.37} & \textcolor{gray}{0.27} & \textcolor{gray}{0.63} & \textcolor{gray}{21.12} & \textcolor{gray}{0.36} & \textcolor{gray}{0.59} & \textcolor{gray}{21.78} & \textcolor{gray}{0.29} & \textcolor{gray}{0.61} \\

        & \textcolor{gray}{our refined} & \textcolor{gray}{E$^2$NeRF~\cite{qi2023e2nerf}} 
    & \textcolor{gray}{22.23} 
    & \textcolor{gray}{0.28} 
    & \textcolor{gray}{0.61} 
    
    & \textcolor{gray}{18.41} 
    & \textcolor{gray}{0.45} 
    & \textcolor{gray}{0.45} 
    
    & \textcolor{gray}{14.38} 
    & \textcolor{gray}{0.62} 
    & \textcolor{gray}{0.33} 
    
    & \textcolor{gray}{18.34} 
    & \textcolor{gray}{0.45} 
    & \textcolor{gray}{0.46} \\
    
    & \textcolor{gray}{our refined} & \textcolor{gray}{Ours} 
    & \textbf{\textcolor{gray}{24.40}} 
    & \textbf{\textcolor{gray}{0.21}}
    & \textbf{\textcolor{gray}{0.71}}
    
    & \textbf{\textcolor{gray}{23.58}}
    & \textbf{\textcolor{gray}{0.25}}
    & \textbf{\textcolor{gray}{0.67}}
    
    & \textbf{\textcolor{gray}{23.31}}
    & \textbf{\textcolor{gray}{0.30}}
    & \textbf{\textcolor{gray}{0.67}}
    
    & \textbf{\textcolor{gray}{23.76}}
    & \textbf{\textcolor{gray}{0.26}}
    & \textbf{\textcolor{gray}{0.68}}
    \\

\midrule
\multirow{5}{*}{\textsc{Equipment}} 
  & USLAM & Ev-DeblurNeRF~\cite{cannici2024mitigating} & 
    13.86 & 0.53 & 0.30 & 13.49 & 0.63 & 0.27 & 13.08 & 0.73 & 0.24 & 13.48 & 0.63 & 0.27 \\

  & USLAM & Ours 
  & \textbf{23.78} & \textbf{0.23} & \textbf{0.72} & \textbf{22.90} & \textbf{0.26} & \textbf{0.69} & \textbf{22.07} & \textbf{0.36} & \textbf{0.65} & \textbf{22.91} & \textbf{0.28} & \textbf{0.69} \\

  \cdashline{2-15}\addlinespace[2pt]
  
    & \textcolor{gray}{our refined} & \textcolor{gray}{Ev-DeblurNeRF~\cite{cannici2024mitigating}} 
    & \textcolor{gray}{21.54} & \textcolor{gray}{0.28} & \textcolor{gray}{0.66} & \textcolor{gray}{21.02} & \textcolor{gray}{0.31} & \textcolor{gray}{0.62} & \textcolor{gray}{19.21} & \textcolor{gray}{0.44} & \textcolor{gray}{0.53} & \textcolor{gray}{20.59} & \textcolor{gray}{0.34} & \textcolor{gray}{0.60} \\

            & \textcolor{gray}{our refined} & \textcolor{gray}{E$^2$NeRF~\cite{qi2023e2nerf}} 
    & \textcolor{gray}{21.09} 
    & \textcolor{gray}{0.32} 
    & \textcolor{gray}{0.62} 
    
    & \textcolor{gray}{17.14} 
    & \textcolor{gray}{0.46} 
    & \textcolor{gray}{0.43}
    
    & \textcolor{gray}{12.70} 
    & \textcolor{gray}{0.68} 
    & \textcolor{gray}{0.28}
    
    & \textcolor{gray}{16.97} 
    & \textcolor{gray}{0.49} 
    & \textcolor{gray}{0.44} \\

    & \textcolor{gray}{our refined} & \textcolor{gray}{Ours} 
    & \textbf{\textcolor{gray}{23.77}} 
    & \textbf{\textcolor{gray}{0.23}} 
    & \textbf{\textcolor{gray}{0.73}} 
    
    & \textbf{\textcolor{gray}{23.01}} 
    & \textbf{\textcolor{gray}{0.28}} 
    & \textbf{\textcolor{gray}{0.69}} 
    
    & \textbf{\textcolor{gray}{22.07}} 
    & \textbf{\textcolor{gray}{0.35}} 
    & \textbf{\textcolor{gray}{0.65}} 
    
    & \textbf{\textcolor{gray}{22.95}} 
    & \textbf{\textcolor{gray}{0.29}} 
    & \textbf{\textcolor{gray}{0.69}}
    \\

\midrule
\multirow{5}{*}{\textsc{PlayCorner}} 
  & USLAM & Ev-DeblurNeRF~\cite{cannici2024mitigating} & 
    13.30 & 0.70 & 0.31 & 11.66 & 0.78 & 0.20 & 11.92 & 0.76 & 0.26 & 12.29 & 0.75 & 0.26 \\

  & USLAM & Ours
  & \textbf{25.47} & \textbf{0.34} & \textbf{0.75} & \textbf{23.91} & \textbf{0.39} & \textbf{0.69} & \textbf{23.40} & \textbf{0.41} & \textbf{0.66} & \textbf{24.26} & \textbf{0.38} & \textbf{0.70} \\

  \cdashline{2-15}\addlinespace[2pt]
  
& \textcolor{gray}{our refined} & \textcolor{gray}{Ev-DeblurNeRF~\cite{cannici2024mitigating}} 
& \textcolor{gray}{23.68} & \textcolor{gray}{0.36} & \textcolor{gray}{0.66} & \textcolor{gray}{21.16} & \textcolor{gray}{0.44} & \textcolor{gray}{0.60} & \textcolor{gray}{22.51} & \textcolor{gray}{0.43} & \textcolor{gray}{0.63} & \textcolor{gray}{22.45} & \textcolor{gray}{0.41} & \textcolor{gray}{0.63} \\

        & \textcolor{gray}{our refined} & \textcolor{gray}{E$^2$NeRF~\cite{qi2023e2nerf}} 
    & \textcolor{gray}{22.63} 
    & \textcolor{gray}{0.38} 
    & \textcolor{gray}{0.63} 
    
    & \textcolor{gray}{21.35} 
    & \textcolor{gray}{0.47} 
    & \textcolor{gray}{0.58}
    
    & \textcolor{gray}{22.41} 
    & \textcolor{gray}{0.51} 
    & \textcolor{gray}{0.61}
    
    & \textcolor{gray}{22.13} 
    & \textcolor{gray}{0.45} 
    & \textcolor{gray}{0.60} \\

    & \textcolor{gray}{our refined} & \textcolor{gray}{Ours} 
    & \textbf{\textcolor{gray}{25.54}} 
    & \textbf{\textcolor{gray}{0.33}} 
    & \textbf{\textcolor{gray}{0.74}} 
    
    & \textbf{\textcolor{gray}{24.20}} 
    & \textbf{\textcolor{gray}{0.36}} 
    & \textbf{\textcolor{gray}{0.69}} 
    
    & \textbf{\textcolor{gray}{23.55}} 
    & \textbf{\textcolor{gray}{0.40}} 
    & \textbf{\textcolor{gray}{0.67}} 
    
    & \textbf{\textcolor{gray}{24.43}} 
    & \textbf{\textcolor{gray}{0.37}} 
    & \textbf{\textcolor{gray}{0.70}} \\

\midrule

  \new{\multirow{5}{*}{\textsc{Models}}} 
  & \new{USLAM} & \new{Ev-DeblurNeRF~\cite{cannici2024mitigating}} &
  \new{10.50} & \new{0.74} & \new{0.19} & \new{10.71} & \new{0.77} & \new{0.16} & \new{11.48} & \new{0.78} & \new{0.18} & \new{10.90} & \new{0.76} & \new{0.18}\\

  & \new{USLAM} & \new{Ours} &
  \new{\textbf{26.60}} & \new{\textbf{0.16}} & \new{\textbf{0.79}} &
  \new{\textbf{25.04}} & \new{\textbf{0.20}} & \new{\textbf{0.74}} &
  \new{\textbf{24.81}} & \new{\textbf{0.24}} & \new{\textbf{0.70}} &
  \new{\textbf{25.48}} & \new{\textbf{0.20}} & \new{\textbf{0.74}}\\

  \cdashline{2-15}
  \addlinespace[2pt]
  
& \textcolor{gray}{our refined} & \textcolor{gray}{Ev-DeblurNeRF~\cite{cannici2024mitigating}} 

& \textcolor{gray}{26.43} 
& \textcolor{gray}{0.17}
& \textcolor{gray}{0.79}

& \textcolor{gray}{24.95}
& \textcolor{gray}{0.21}
& \textcolor{gray}{0.75} 

& \textcolor{gray}{24.09} 
& \textcolor{gray}{0.28}
& \textcolor{gray}{0.68}

& \textcolor{gray}{25.16}
& \textcolor{gray}{0.22}
& \textcolor{gray}{0.74}\\

& \textcolor{gray}{our refined} & \textcolor{gray}{E$^2$NeRF~\cite{qi2023e2nerf}} 
    & \textcolor{gray}{24.02}
    & \textcolor{gray}{0.18}
    & \textcolor{gray}{0.73}
    
    & \textcolor{gray}{22.67}
    & \textcolor{gray}{0.25}
    & \textcolor{gray}{0.68}
    
    & \textcolor{gray}{21.81}
    & \textcolor{gray}{0.32}
    & \textcolor{gray}{0.62}
    
    & \textcolor{gray}{22.83}
    & \textcolor{gray}{0.25}
    & \textcolor{gray}{0.68}\\

& \textcolor{gray}{our refined} & \textcolor{gray}{Ours} 
& \textbf{\textcolor{gray}{26.90}}
& \textbf{\textcolor{gray}{0.15}}
& \textbf{\textcolor{gray}{0.80}}

& \textbf{\textcolor{gray}{25.76}}
& \textbf{\textcolor{gray}{0.18}}
& \textbf{\textcolor{gray}{0.76}}

& \textbf{\textcolor{gray}{25.11}}
& \textbf{\textcolor{gray}{0.22}}
& \textbf{\textcolor{gray}{0.72}}

& \textbf{\textcolor{gray}{25.93}} 
& \textbf{\textcolor{gray}{0.18}}
& \textbf{\textcolor{gray}{0.76}}\\

\midrule

\multirow{5}{*}{\new{\textsc{ReadingCorner}}} 
  & \new{USLAM} & \new{Ev-DeblurNeRF~\cite{cannici2024mitigating}} & 

    \new{13.23} & \new{0.69} & \new{0.36} & \new{13.76} & \new{0.67} & \new{0.39} & \new{11.77} & \new{0.70} & \new{0.13} & \new{12.92} & \new{0.69} & \new{0.29}\\

  & \new{USLAM} & \new{Ours}

  & \textbf{\new{27.40}} & \textbf{\new{0.20}} & \textbf{\new{0.76}}
  & \textbf{\new{27.22}} & \textbf{\new{0.22}} & \textbf{\new{0.75}}
  & \textbf{\new{25.70}} & \textbf{\new{0.29}} & \textbf{\new{0.72}}
  & \textbf{\new{26.77}} & \textbf{\new{0.24}} & \textbf{\new{0.74}}\\

  \cdashline{2-15}
  \addlinespace[2pt]
  
& \textcolor{gray}{our refined} & \textcolor{gray}{Ev-DeblurNeRF~\cite{cannici2024mitigating}} 

& \textcolor{gray}{27.60} 
& \textcolor{gray}{0.19} 
& \textcolor{gray}{0.76} 

& \textcolor{gray}{26.81} 
& \textcolor{gray}{0.22} 
& \textcolor{gray}{0.74} 

& \textcolor{gray}{26.46}
& \textcolor{gray}{0.25}
& \textbf{\textcolor{gray}{0.76}}

& \textcolor{gray}{26.96} 
& \textcolor{gray}{0.22} 
& \textcolor{gray}{0.75}\\

& \textcolor{gray}{our refined} & \textcolor{gray}{E$^2$NeRF~\cite{qi2023e2nerf}} 
    & \textcolor{gray}{25.66}
    & \textcolor{gray}{0.20}
    & \textcolor{gray}{0.72}
    
    & \textcolor{gray}{26.20}
    & \textcolor{gray}{0.23}
    & \textcolor{gray}{0.72}
    
    & \textcolor{gray}{24.40}
    & \textcolor{gray}{0.31}
    & \textcolor{gray}{0.68}
    
    & \textcolor{gray}{25.42}
    & \textcolor{gray}{0.25}
    & \textcolor{gray}{0.71}\\

& \textcolor{gray}{our refined} & \textcolor{gray}{Ours} 
    & \textbf{\textcolor{gray}{27.63}}
    & \textbf{\textcolor{gray}{0.19}}
    & \textbf{\textcolor{gray}{0.76}}
    
    & \textbf{\textcolor{gray}{27.42}}
    & \textbf{\textcolor{gray}{0.21}}
    & \textbf{\textcolor{gray}{0.76}}
    
    & \textbf{\textcolor{gray}{26.62}}
    & \textbf{\textcolor{gray}{0.24}}
    & \textcolor{gray}{0.75}
    
    & \textbf{\textcolor{gray}{27.22}}
    & \textbf{\textcolor{gray}{0.21}}
    & \textbf{\textcolor{gray}{0.76}}\\

\midrule

\new{\multirow{5}{*}{\textsc{Lounge}}} 
  & \new{USLAM} & \new{Ev-DeblurNeRF~\cite{cannici2024mitigating}} &
  \new{12.26} & \new{0.68} & \new{0.17} & \new{12.65} & \new{0.69} & \new{0.18} & \new{11.74} & \new{0.69} & \new{0.13} & \new{12.22} & \new{0.69} & \new{0.16}\\

  & \new{USLAM} & \new{Ours} &
  \new{\textbf{27.88}} & \new{\textbf{0.23}} & \new{\textbf{0.75}} &
  \new{\textbf{27.33}} & \new{\textbf{0.18}} & \new{\textbf{0.72}} &
  \new{\textbf{25.61}} & \new{\textbf{0.22}} & \new{\textbf{0.67}} &
  \new{\textbf{26.94}} & \new{\textbf{0.21}} & \new{\textbf{0.71}}\\

  \cdashline{2-15}
  \addlinespace[2pt]
  
& \textcolor{gray}{our refined} & \textcolor{gray}{Ev-DeblurNeRF~\cite{cannici2024mitigating}} 

& \textcolor{gray}{27.65} 
& \textcolor{gray}{0.23}
& \textcolor{gray}{0.74}

& \textcolor{gray}{27.32}
& \textcolor{gray}{0.18} 
& \textcolor{gray}{0.72}

& \textcolor{gray}{25.28} 
& \textcolor{gray}{0.22}
& \textcolor{gray}{0.66}

& \textcolor{gray}{26.75} 
& \textcolor{gray}{0.21} 
& \textcolor{gray}{0.71}\\

& \textcolor{gray}{our refined} & \textcolor{gray}{E$^2$NeRF~\cite{qi2023e2nerf}} 
    & \textcolor{gray}{27.13} 
    & \textbf{\textcolor{gray}{0.19}} 
    & \textcolor{gray}{0.74} 
    
    & \textcolor{gray}{26.60} 
    & \textcolor{gray}{0.17} 
    & \textcolor{gray}{0.71} 
    
    & \textcolor{gray}{24.84} 
    & \textcolor{gray}{0.21} 
    & \textcolor{gray}{0.64} 
    
    & \textcolor{gray}{26.19} 
    & \textcolor{gray}{0.19} 
    & \textcolor{gray}{0.70}\\

& \textcolor{gray}{our refined} & \textcolor{gray}{Ours} 
& \textbf{\textcolor{gray}{28.76}} 
& \textcolor{gray}{0.20} 
& \textbf{\textcolor{gray}{0.77}}

& \textbf{\textcolor{gray}{27.57}} 
& \textbf{\textcolor{gray}{0.17}} 
& \textbf{\textcolor{gray}{0.73}} 

& \textbf{\textcolor{gray}{25.68}}
& \textbf{\textcolor{gray}{0.20}} 
& \textbf{\textcolor{gray}{0.67}} 

& \textbf{\textcolor{gray}{27.34}} 
& \textbf{\textcolor{gray}{0.19}} 
& \textbf{\textcolor{gray}{0.72}}\\

\midrule
\midrule
\new{\multirow{5}{*}{Average}}
  & \new{USLAM} & \new{Ev-DeblurNeRF~\cite{cannici2024mitigating}} &
  \new{12.75} & \new{0.65} & \new{0.27} &
  \new{12.39} & \new{0.70} & \new{0.24} &
  \new{12.26} & \new{0.72} & \new{0.21} &
  \new{12.47} & \new{0.69} & \new{0.24}\\

  & \new{USLAM} & \new{Ours} &
  \new{\textbf{25.84}} & \new{\textbf{0.23}} & \new{\textbf{0.74}} &
  \new{\textbf{24.98}} & \new{\textbf{0.25}} & \new{\textbf{0.71}} &
  \new{\textbf{24.11}} & \new{\textbf{0.31}} & \new{\textbf{0.68}} &
  \new{\textbf{24.98}} & \new{\textbf{0.26}} & \new{\textbf{0.71}}\\

  \cdashline{2-15}
  \addlinespace[2pt]
  
& \textcolor{gray}{our refined} & \textcolor{gray}{Ev-DeblurNeRF~\cite{cannici2024mitigating}} 

& \textcolor{gray}{24.79} & \textcolor{gray}{0.24 }& \textcolor{gray}{0.70 }& \textcolor{gray}{23.94} & \textcolor{gray}{0.27 }& \textcolor{gray}{0.68 }& \textcolor{gray}{23.11} & \textcolor{gray}{0.33 }& \textcolor{gray}{0.64 }& \textcolor{gray}{23.95} & \textcolor{gray}{0.28 }& \textcolor{gray}{0.67} \\

& \textcolor{gray}{our refined} & \textcolor{gray}{E$^2$NeRF~\cite{qi2023e2nerf}} 
    & \textcolor{gray}{23.79} & \textcolor{gray}{0.26 }& \textcolor{gray}{0.68 }& \textcolor{gray}{22.06} & \textcolor{gray}{0.34 }& \textcolor{gray}{0.60 }& \textcolor{gray}{20.09} & \textcolor{gray}{0.44}& \textcolor{gray}{0.53}& \textcolor{gray}{21.98} & \textcolor{gray}{0.35 }& \textcolor{gray}{0.60} \\

& \textcolor{gray}{our refined} & \textcolor{gray}{Ours} 
& \textbf{\textcolor{gray}{26.17}} 
& \textbf{\textcolor{gray}{0.22 }}
& \textbf{\textcolor{gray}{0.75 }}

& \textbf{\textcolor{gray}{25.26}} 
& \textbf{\textcolor{gray}{0.24 }}
& \textbf{\textcolor{gray}{0.72 }}

& \textbf{\textcolor{gray}{24.39}} 
& \textbf{\textcolor{gray}{0.29 }}
& \textbf{\textcolor{gray}{0.69 }}

& \textbf{\textcolor{gray}{25.27}} 
& \textbf{\textcolor{gray}{0.25 }}
& \textbf{\textcolor{gray}{0.72}}\\

\bottomrule
\end{tabular}}
\end{table*}

%% file: floaters/tables/traj_length.tex
\begin{table*}
\caption{\new{Quantitative comparison on trajectories of different lengths. All trajectories were flown by a real drone under the fastest speed profile. Scene 1: \textsc{Models}, Scene 2: \textsc{ReadingCorner}.}}
\label{tab:traj_length}

\centering
\resizebox{\textwidth}{!}{
\setlength{\tabcolsep}{1pt}
\begin{tabular}{l c ccc ccc ccc ccc}
\toprule
\multirow{2}{*}{Poses}& \multirow{2}{*}{Method} & \multicolumn{3}{c}{Scene 1 (15 m Traj.)} & \multicolumn{3}{c}{Scene 2 (15 m Traj.)} & \multicolumn{3}{c}{Scene 1+2 (30 m Traj.)} & \multicolumn{3}{c}{\textsc{Average}} \\ 
    \cmidrule(lr){3-5} \cmidrule(lr){6-8} \cmidrule(lr){9-11} \cmidrule(lr){12-14}

&  & PSNR$\uparrow$ & LPIPS$\downarrow$ & SSIM$\uparrow$ & PSNR$\uparrow$ & LPIPS$\downarrow$ & SSIM$\uparrow$ & PSNR$\uparrow$ & LPIPS$\downarrow$ & SSIM$\uparrow$ & PSNR$\uparrow$ & LPIPS$\downarrow$ & SSIM$\uparrow$\\
\midrule
USLAM & Ev-DeblurNeRF~\cite{cannici2024mitigating}  & 11.48 & 0.78 & 0.18 & 11.77 & 0.70 & 0.13 & 10.22 & 0.77 & 0.15 & 11.16 & 0.75 & 0.15 \\
USLAM & Ours & \textbf{24.81} & \textbf{0.24} & \textbf{0.70}  & \textbf{25.70} & \textbf{0.29} & \textbf{0.72} & \textbf{25.59} & \textbf{0.25} & \textbf{0.74} & \textbf{25.37} & \textbf{0.26} & \textbf{0.72} \\

 \cdashline{1-14}\addlinespace[3pt]
\textcolor{gray}{our refined} & \textcolor{gray}{Ev-DeblurNeRF~\cite{cannici2024mitigating}} & 
\textcolor{gray}{24.09} & \textcolor{gray}{0.28} & \textcolor{gray}{0.68} &
\textcolor{gray}{26.46} & \textcolor{gray}{0.25} & \textcolor{gray}{\textbf{0.76}} &
\textcolor{gray}{25.93} & \textcolor{gray}{0.24} & \textcolor{gray}{0.75} &
\textcolor{gray}{25.49} & \textcolor{gray}{0.26} & \textcolor{gray}{0.73} \\

\textcolor{gray}{our refined} & \textcolor{gray}{E$^2$NeRF~\cite{qi2023e2nerf}} &
\textcolor{gray}{21.81} & \textcolor{gray}{0.32} & \textcolor{gray}{0.62} &
\textcolor{gray}{24.40} & \textcolor{gray}{0.31} & \textcolor{gray}{0.68} &
\textcolor{gray}{24.17} & \textcolor{gray}{0.27} & \textcolor{gray}{0.70} &
\textcolor{gray}{23.46} & \textcolor{gray}{0.30} & \textcolor{gray}{0.67} \\

\textcolor{gray}{our refined} & \textcolor{gray}{Ours} &
\textcolor{gray}{\textbf{25.11}} & \textcolor{gray}{\textbf{0.22}} & \textcolor{gray}{\textbf{0.72}} &
\textcolor{gray}{\textbf{26.62}} & \textcolor{gray}{\textbf{0.24}} & \textcolor{gray}{0.75} &
\textcolor{gray}{\textbf{26.13}} & \textcolor{gray}{\textbf{0.23}} & \textcolor{gray}{\textbf{0.75}} &
\textcolor{gray}{\textbf{25.95}} & \textcolor{gray}{\textbf{0.23}} & \textcolor{gray}{\textbf{0.74}} \\

\bottomrule
\end{tabular}
}
\end{table*}

%% file: floaters/tables/ablations.tex
\begin{table*}[]
\caption{\new{Ablation study on the technical components of our method.}}
\label{tab:ablations}
\renewcommand{\arraystretch}{1.15} 
\setlength{\tabcolsep}{4pt}        
\resizebox{\textwidth}{!}{
\begin{tabular}{ccccccccccccccc}
\toprule
\multirow{2}{*}{$\mathcal{L}_\text{blur}$} & \multirow{2}{*}{$\mathcal{L}_\text{prior}$} & \multirow{2}{*}{$\mathcal{L}_\text{event}$} & \multicolumn{3}{c}{ Max Speed 1 m/s} & \multicolumn{3}{c}{ Max Speed 1.5 m/s} & \multicolumn{3}{c}{ Max Speed 2 m/s} & \multicolumn{3}{c}{Average} \\
\cmidrule(lr){4-6} \cmidrule(lr){7-9} \cmidrule(lr){10-12} \cmidrule(l){13-15}
 & & & PSNR$\uparrow$ & LPIPS$\downarrow$ & SSIM$\uparrow$ & PSNR$\uparrow$ & LPIPS$\downarrow$ & SSIM$\uparrow$ & PSNR$\uparrow$ & LPIPS$\downarrow$ & SSIM$\uparrow$ & PSNR$\uparrow$ & LPIPS$\downarrow$ & SSIM$\uparrow$ \\
\midrule
  \checkmark &  & 
  & 23.60 & 0.44 & 0.66 & 18.65 & 0.60 & 0.46 & 15.74 & 0.68 & 0.36 & 19.33 & 0.57 & 0.49\\

 \checkmark & & \checkmark          
  & 25.10 & 0.37 & 0.71 & 21.75 & 0.52 & 0.59 & 23.01 & 0.46 & 0.63 & 23.29 & 0.45 & 0.64\\

  & \checkmark & 
  & 23.81 & 0.43 & 0.67 & 20.53 & 0.57 & 0.55 & 21.25 & 0.55 & 0.56 & 21.86 & 0.52 & 0.59\\

  & \checkmark & \checkmark 
  & 24.73 & 0.35 & 0.74 & 23.53 & 0.41 & 0.71 & 21.75 & 0.47 & 0.62 & 23.34 & 0.41 & 0.69\\

  \checkmark & \checkmark & 
  & 24.11 & 0.42 & 0.67 & 21.48 & 0.52 & 0.57 & 21.86 & 0.51 & 0.58 & 22.49 & 0.48 & 0.61\\

  \checkmark & \checkmark  & \checkmark
  & \textbf{25.47} & \textbf{0.34} & \textbf{0.75 }& \textbf{23.91} & \textbf{0.39} &\textbf{ 0.69} & \textbf{23.40} & \textbf{0.41} & \textbf{0.66} & \textbf{24.26} & \textbf{0.38 }& \textbf{0.70}\\

  \midrule
  \multicolumn{3}{c}{w/o eCRF}
  & 24.96 & 0.36 & 0.72 & 23.37 & 0.46 & 0.66 & 23.24 & 0.45 & 0.65 & 23.85 & 0.42 & 0.68\\

\bottomrule
\end{tabular}}
\end{table*}

%% file: sections/5_discussions.tex
\new{
\section{Discussion and Limitations}
\label{sec:discussions}

\myparagraph{Extension to Gaussian Splatting.}
Our focus in this work is accurate and stable reconstruction under the fast motion of aerial drones, where robustness to sparse input frames and generalization to unseen viewpoints are critical due to high motion speed and limited capture frequency. 
Recent analyses~\cite{he2024nerfgs,fang2025nerfgs} indicate that NeRFs are more stable to train and provide stronger geometric consistency and better generalization than 3D Gaussian Splatting with limited training views. Our choice of a NeRF backbone therefore prioritizes reconstruction fidelity and robustness under sparse training views. However, our framework is agnostic to the underlying representation of the scene. In scenarios where rendering speed is more important, our method can be naturally extended to 3DGS. Compared to NeRF, which represents the scene implicitly through a neural network, 3DGS models it explicitly as a set of Gaussian splats parameterized by position, covariance, and appearance. By rendering these splats differentiably with respect to the continuous-time camera trajectory analogous to our current formulation, the same image- and event-based supervision can be used to jointly optimize both the splat parameters and the camera poses.

\myparagraph{Alternative Hardware Settings.}
In this work, we use a beamsplitter-based sensor rig to obtain synchronized RGB images and events. This hardware setting can be replaced in several ways in real-world deployment. 
A promising hardware alternative is the emerging field of hybrid vision sensors (HVS) that provide co-registered frame and event outputs on the same chip~\cite{alpsentek2023hvs, guo2023hvs, kodama2023hvs}, eliminating synchronization and co-location issues as the image and event outputs share the same pixel array.
When such hybrid sensors are unavailable, non-synchronized cameras (independently mounted RGB and event cameras) can still be used, provided that accurate timestamps and extrinsic calibrations are available. In this case, unsynchronized captures could be handled by introducing time offset parameters between sensors and optimizing them jointly with the continuous-time trajectory and scene parameters, ensuring temporal alignment during training. Similarly, small extrinsic errors can also be refined during optimization.

\myparagraph{Motion Capture Poses for Evaluation.}
Our current quantitative evaluation uses motion capture poses for rendering to compute accurate metrics and ensure fair comparison across methods. However, such infrastructure is often unavailable in practical deployment. We propose several alternatives that do not rely on motion capture for evaluating reconstruction quality: (i) adopting no-reference image quality metrics to assess the visual fidelity of reconstructed views~\cite{liu2013no_reference, wang2011no_reference, mittal2012no_reference}; (ii) measuring multi-view geometric consistency within the reconstructed scene to infer potential inconsistency; and (iii) evaluating reconstruction quality through downstream tasks, where task performance serves as an indirect indicator of reconstruction accuracy. These motion-capture-free evaluation strategies could be used for assessing the quality of our reconstructed radiance field in real-world scenarios.

\myparagraph{Presence of Dynamic Objects.}
Another limitation of the current work is the assumption 
that the observed scene is static, with motion blur arising solely from the rapid movement of the camera. This assumption is consistent with concurrent works~\cite{qi2023e2nerf,deguchi2024e2gs,xiong2024event3dgs} and with traditional Structure-from-Motion pipelines, where scene dynamics are not explicitly modeled. A promising direction for future work would be to relax this assumption by incorporating ideas from recent dynamic NeRF and dynamic 3DGS methods~\cite{wu20244d,lee2024fully,li2022neural}, enabling the system to distinguish between static and dynamic regions of the scene. Leveraging event-based motion cues together with on-board IMU measurements could further help disentangle object motion from camera ego-motion, improving reconstruction fidelity in dynamic environments.
}

%% file: sections/6_conclusions.tex
\section{Conclusion}
\label{sec:conclusion}

We introduced a unified framework for radiance field reconstruction under fast motion, specifically tailored to the challenges of aerial robotics. \new{By leveraging the complementary sensing of motion-blurred RGB images and asynchronous event data, our method enables accurate scene reconstruction without requiring ground-truth poses from e.g., a motion capture system, which is often unavailable in practice.}
Central to our approach is a continuous-time pose refinement module that improves initial VIO estimates, allowing both events and frames to jointly supervise the radiance field and the underlying motion.

We validated our framework on two newly collected real-world datasets, including onboard drone flights at high speed. Our results on synthetic and real-world datasets demonstrate that the proposed system can recover sharp radiance fields even under high-dynamic motion, where RGB frames are heavily degraded by motion blur and pose priors are unreliable.

Our work represents a significant step forward toward reliable, high-fidelity scene reconstruction for aerial robots operating under agile motion, opening the door to practical deployment of NeRF-like representations for robotic tasks such as infrastructure inspection, terrain exploration, and search-and-rescue, where rapid coverage and accurate spatial modeling are essential for operational success.

%% file: sections/7_acknowledgements.tex
\section{Acknowledgements}
We sincerely thank Leonard Bauersfeld for his invaluable support throughout this research.

%% file: sections/n_biography.tex
 
\begin{IEEEbiography}[{\includegraphics[width=1in,height=1.25in,clip,keepaspectratio]{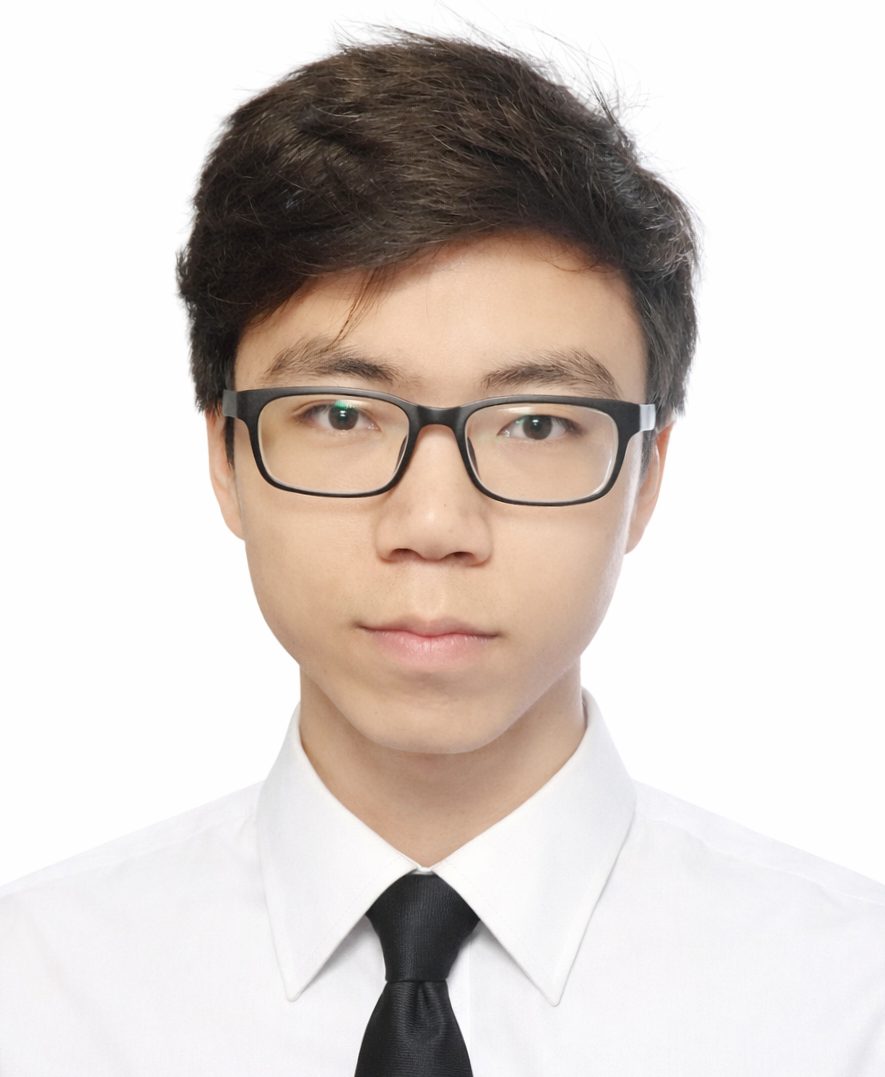}}]{Rong Zou} received the M.Sc. degree in robotics, systems and control from ETH Zurich, Zurich, Switzerland, in 2024. He is currently pursuing the Ph.D. degree in robotic perception with the Robotics and Perception Group led by Prof. Davide Scaramuzza. He is also an Associated Doctoral Researcher with the ETH AI Center, Zurich. His research interests lie at the intersection of vision, learning and robotics, with a focus on event-based visual sensing and transferable representation learning for various robotic tasks such as motion estimation, scene understanding, and robust perception in challenging conditions.
\end{IEEEbiography}

\begin{IEEEbiography}[{\includegraphics[width=1in,height=1.25in,clip,keepaspectratio]{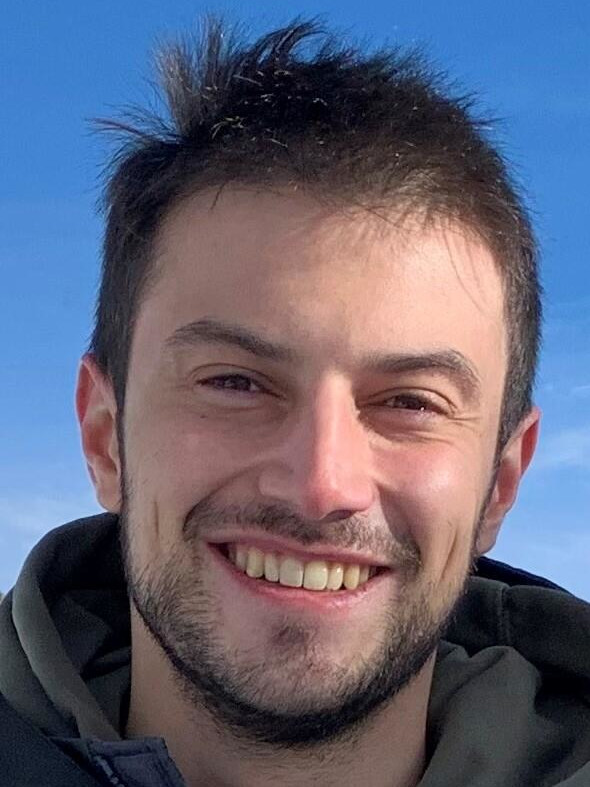}}]{Marco Cannici} received the M.Sc. degree in 2018 and the Ph.D. degree in 2022 from Politecnico di Milano, Italy. From 2022 to 2025, he was a postdoctoral researcher with the Robotics and Perception Group at the University of Zurich, under the supervision of Prof. Davide Scaramuzza. His research focuses on event-based vision and perception for autonomous systems, spanning from visual odometry and monocular obstacle avoidance to high-speed and low-latency perception and 3D reconstruction in challenging conditions. He is currently a researcher at the Smart Eyewear Lab, a joint platform initiative between EssilorLuxottica and the Politecnico di Milano, where he develops event-based perception and sensing systems for smart glasses.
\end{IEEEbiography}

\begin{IEEEbiography}[{\includegraphics[width=1in,height=1.25in,clip,keepaspectratio]{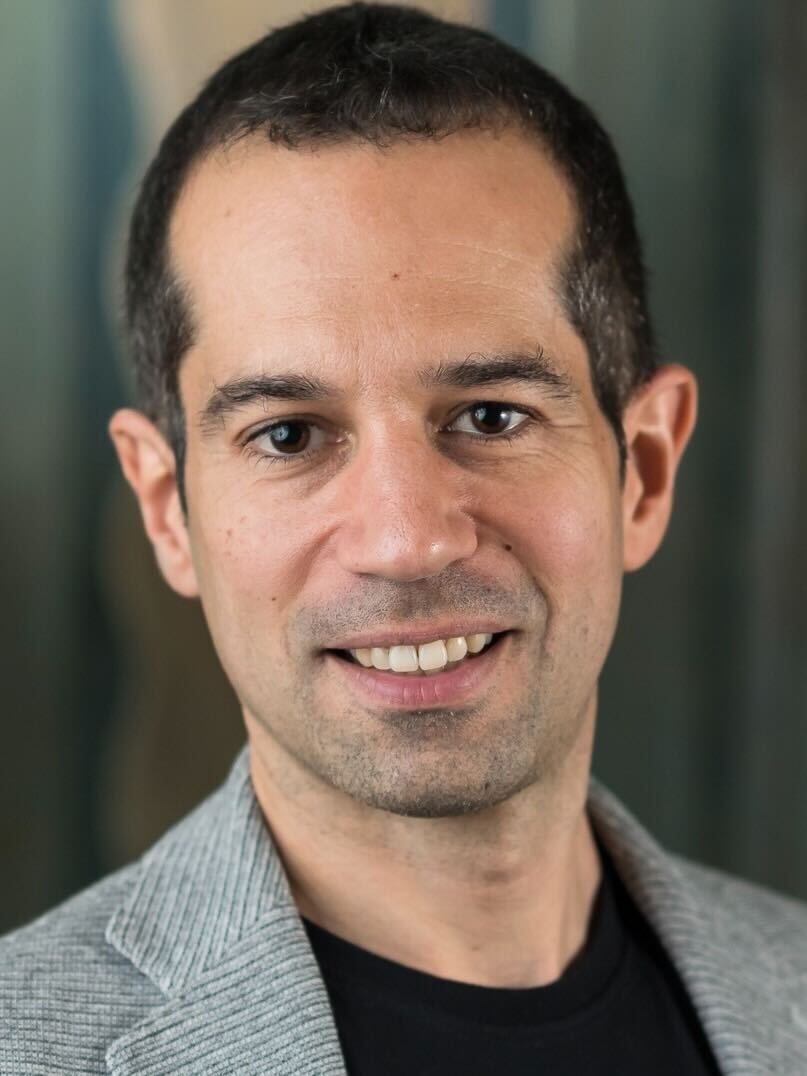}}]{Davide Scaramuzza} is a Professor of Robotics and Perception at the University of Zurich. He did his Ph.D. at ETH Zurich, a postdoc at the University of Pennsylvania, and was a visiting professor at Stanford University and NASA Jet Propulsion Laboratory. His research focuses on autonomous, agile navigation of mobile robots using standard and event-based cameras. He made fundamental contributions to visual-inertial state estimation, autonomous vision-based agile navigation of micro flying robots, and low-latency perception with event cameras, which were transferred to many products, from drones to automobiles, cameras, AR/VR headsets, and mobile devices. He pioneered autonomous, vision-based navigation of drones, which inspired the algorithm of the NASA Mars helicopter. In 2022, his team demonstrated that an AI-powered drone could outperform the world champions of drone racing. He received several awards, including an IEEE Technical Field Award, the IEEE Fellowship, the IEEE Robotics and Automation Society Early Career Award, a European Research Council Consolidator Grant, a Google Research Award, and many paper awards. In 2015, he co-founded Zurich-Eye, today Meta Zurich, which developed the head-tracking software of the Meta Quest. In 2020, he co-founded SUIND, which builds autonomous drones for precision agriculture. Many aspects of his research have been featured in the media, such as The New York Times, The Guardian, The Economist, and Forbes. He co-authored the book "Introduction to Autonomous Mobile Robots," published by MIT Press, which has sold over 10 thousand copies worldwide and is among the most used textbooks for teaching mobile robotics. He has been consulting the United Nations on disaster response, the Fukushima Action Plan, disarmament, and AI for good.
\end{IEEEbiography}


\vfill